\documentclass[10pt,logo,copyright]{nvidiatechreport}
\usepackage[authoryear,sort&compress,round]{natbib}
\usepackage{subcaption}
\usepackage{mathtools}
\usepackage{amsthm}
\usepackage{multirow}
\usepackage{makecell}
\usepackage{adjustbox}
\usepackage{float}
\usepackage[capitalize,noabbrev,nameinlink]{cleveref}

\theoremstyle{plain}

\theoremstyle{definition}

\theoremstyle{remark}

\crefname{equation}{Eq.}{Eqs.}
\crefname{figure}{Fig.}{Figs.}
\crefname{section}{Sec.}{Secs.}
\crefname{table}{Table}{Tables}

\Crefname{equation}{Eq.}{Eqs.}
\Crefname{figure}{Fig.}{Figs.}
\Crefname{table}{Table}{Tables}


\title{Tacinf: A Unified Tactile Infrastructure for Robot Manipulation}

\date{2026-11-22}

\begin{document}

\author{
  Qiwei Wu$^{12\ast\ddagger}$ \;
  Xiangyu Chen$^{2\ast}$ \;
  Xin Xiang$^{2\ast}$ \;
  Rui Zhang$^{1\ast}$ \;
  Yihang Kang$^{2}$ \;
  Weiran Chen$^{2}$ \;
  Zikang Xie$^{2}$ \; \newline
  Yangfan Li$^{2}$ \;
  Tao Wang$^{2}$ \;
  Xiangzheng Wang$^{2}$ \;
  Tianyi Bi$^{2}$ \;
  Zhiyan Zhang$^{2}$ \;
  Xianpeng Wang$^{2}$ \;
  Mingqian Zhou$^{3}$ \;
  Yan Yu$^{2}$ \;
  Chuanyu Li$^{2}$ \;
  Wenjie Zhang$^{2}$ \;
  Tao Li$^{1}$  \;
  Weihua Zhang$^{1\dagger}$ \;
  Renjing Xu$^{2\dagger}$ \;
  Junjie Lai$^{1}$ \;
 \\

  \small $^{1}$Nvidia \quad
         $^{2}$HKUST(GZ) \quad
         $^{3}$University of Cambridge \\
  \small $^{\ast}$Equal Contribution \quad
  \small $^{\ddagger}$ Project Leader \quad
  \small $^\dagger$ Corresponding Authors \\
  {\small \href{https://tacinf.github.io/}{\texttt{tacinf.github.io}}}
}

\maketitle

\begin{abstract}
Force-aware manipulation requires robots to infer appropriate interaction forces from task context and tactile feedback. We present Tacinf, a low-cost, modular infrastructure that enables policies to predict robot actions and fingertip forces from task language, vision, and touch. Task instructions can optionally specify the desired force level. Tacinf integrates simulation data synthesis and tactile benchmarking (Tabero), vision-tactile-language-action (VTLA) policy learning, handheld data collection using the same gripper as robot deployment (Tacumi), and task-aligned sim-to-real evaluation (Taco). Together, these components form an integrated workflow connecting data collection from multiple sources, foundation-model training, and hardware deployment. We validate each component, including the simulation data, benchmark, controller, and Tacumi, and are releasing the full stack as open source. A foundation model fine-tuned solely on tactile simulation data transfers zero-shot to different real-robot tasks. Across eight policy variants, mean task success rates are strongly correlated between simulation and the real robot (Pearson r=0.998, mean maximum rank violation MMRV=0.010). For the four policies with reported force measurements, mean grip force is higher under firm instructions than under gentle instructions on all three tasks in both domains. When both object positions and instances are randomized, a mixture of teleoperation, simulation, and Tacumi data raises mean real-robot task success from 20\% with teleoperation alone to 60\%.
\end{abstract}

\abscontent

\section{Introduction}
\label{sec:intro}

Foundation models for robot manipulation have advanced rapidly. Force-aware tasks require these models to infer and regulate physical interaction as well as select motions from images and task language. Force requirements may be explicit, as in instructions to act ``gently'' or ``firmly'', or implicit in tasks such as opening a container or wiping a surface. In both cases, the policy must infer appropriate contact behavior from the instruction, scene, and tactile feedback. Tactile data are essential for training such policies, but differences in hardware and sensor design make effective datasets costly to collect. During execution, the system must use real-time feedback to adjust its behavior and apply appropriate forces. Poorly performing policies can damage objects or the robot itself, creating a need for low-cost evaluation platforms that support policy testing and refinement.

\begin{figure}[t]
  \centering
  \includegraphics[width=0.9\textwidth]{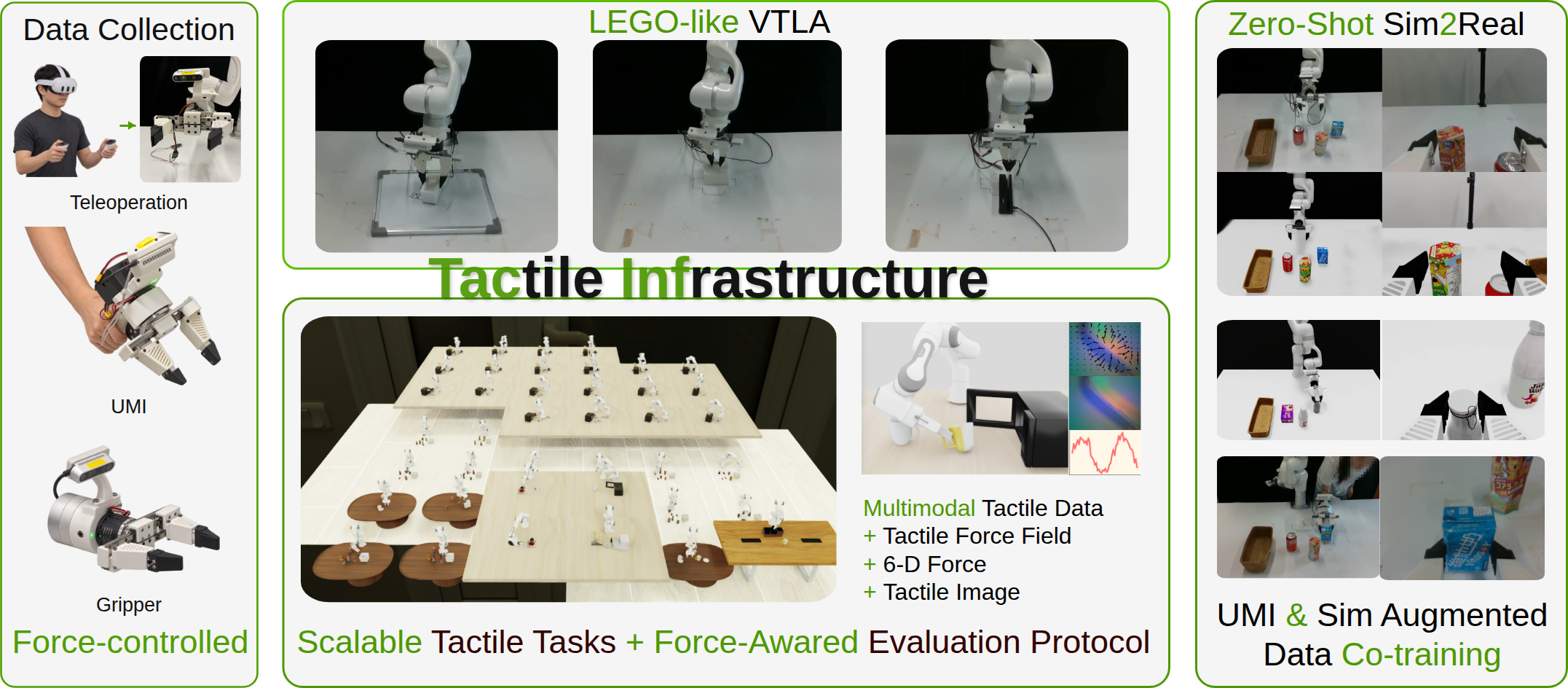}
  \caption{\textbf{Tacinf's tactile data collection, force-aware evaluation, and Sim2Real deployment.} Left: Quest~3 teleoperation, Universal Manipulation Interface (UMI), and the force-controlled gripper; middle: LEGO-like VTLA, scalable tactile tasks, and the force-aware evaluation protocol; right: zero-shot Sim2Real evaluation and joint training on UMI and simulation-augmented data.}
  \label{fig:main_overview}
\end{figure}

\noindent We introduce Tacinf (Fig.~\ref{fig:main_overview}), a full-stack open-source infrastructure for tactile and force-aware manipulation. Tacinf-Tabero generates simulated data with tactile observations and force supervision and provides a benchmark for evaluating vision-tactile-language-action (VTLA) policies. Tacinf-VTLA offers modular tactile interfaces for implementing different policies. Tacinf-Tacumi provides hardware for robot deployment and for collecting tactile observations and force targets through teleoperation and handheld demonstrations. Tacinf-Taco supports task-aligned sim-to-real evaluation and training on data from multiple sources. Together, these components support data collection, simulation-based evaluation, sim-to-real transfer, and policy deployment at low hardware, compute, and data costs. The Tacumi kit costs under \$2,500, and the minimum compute requirement is a single RTX 4090.

\noindent Our experiments validate the system components and yield three main findings. First, a tactile foundation model fine-tuned solely on simulation data can be deployed zero-shot on the real robot, achieving success rates of at least 98\% on LIBERO-like tasks, with predicted forces close to the teleoperation references under our evaluation metrics. Second, across eight Tacinf-VTLA policy variants, simulation closely preserves the real-robot ranking by success rate, Pearson $r=0.998$, mean maximum rank violation (MMRV)$=0.010$. Measured grip force distinguishes firm from gentle instructions for the four policies reported, with absolute forces varying across objects and domains. Third, teleoperation, Tacumi, and simulation data make complementary contributions. We demonstrate the utility of each source, including demonstrations collected without a robot, and increase success from 20\% to 60\% on out-of-distribution objects at randomized positions with a low data-collection cost.

\noindent Our contributions are:
\begin{itemize}
    \item \textbf{Multi-source tactile data collection and synergy:} we develop a low-cost, open-source system for collecting multimodal tactile data through complementary methods. These sources form a data pyramid whose combined benefits are evaluated through ablation studies.
    \item \textbf{Force control and execution:} we use data from multiple sources to build a complete system connecting policy prediction to actuation, enabling scalable force control for contact-rich tasks.
    \item \textbf{Sim-to-real consistency:} we align the simulation and physical systems used by tactile manipulation foundation models, achieving closely matched behavior and reducing evaluation costs.
\end{itemize}

\section{Related Work}
\label{sec:related}

\subsection{Simulation Platforms and Synthetic Data}
Robot simulation provides reproducible environments for policy learning and evaluation. Early infrastructures such as robosuite, LIBERO, and CALVIN established reusable manipulation tasks and language-conditioned evaluation protocols~\cite{RoboSuite,libero,CALVIN}. Recent platforms extend these capabilities through household scene diversity in RoboCasa, GPU-parallel simulation in ManiSkill3 and Isaac Lab, and unified assets and interfaces in RoboVerse~\cite{robocasa,maniskill3,isaaclab,roboverse}. MuBlE couples MuJoCo dynamics with Blender rendering for tasks requiring both visual and physical reasoning~\cite{MuBIE}. At the data-generation level, RoboTwin and RoboTwin 2.0 combine task construction and demonstration generation for bimanual manipulation, while SkillMimicGen reuses demonstrations through skill-level generation~\cite{robotwin1,robotwin2,skillmimicgen}. GraspVLA further demonstrates the use of large-scale synthetic action data for policy pretraining~\cite{graspVLA}. These approaches motivate our reuse of existing trajectories, but trajectory replay alone does not establish that contact forces or tactile observations transfer across simulators.

\noindent A complementary direction reconstructs real environments for simulation-based policy evaluation. SIMPLER studies visual and control gaps and evaluates whether simulated performance predicts real-robot behavior; PolaRiS combines neural scene reconstruction with simulation-data co-training~\cite{simpler,jain2025polarisscalablerealtosimevaluations}. SimFoundry automates scene reconstruction and generates task-relevant scene variations, while SceneRig reconstructs simulation-ready 3D scenes from single images~\cite{simfoundry,scenerig2026}. Seed3D 2.0 generates 3D assets for simulation~\cite{seed3d}. Ctrl-World instead uses an action-conditioned generative world model for policy evaluation and improvement~\cite{ctrlworld}. These works establish precedents for real-to-sim reconstruction, data mixing, and cross-domain policy ranking. Tacinf-Taco adapts this direction to tactile observations and force-conditioned evaluation.

\noindent Tactile simulation datasets and benchmarks are also emerging. VTDexManip supports visuotactile pretraining and dexterous manipulation, SafeDiff evaluates force-aware interaction using SafeDoorManip50k, and UniVTAC unifies multi-sensor tactile synthesis, representation learning, and manipulation benchmarking~\cite{vtdex,forcesafety,univtac}. SoftVTBench pairs visuotactile observations with independent deformation ground truth for evaluating deformable-object interaction quality~\cite{softvtbench}. OpenViTac benchmarks visuo-tactile policies in paired simulation and real-world settings and reports sim-to-real agreement across policy families~\cite{openvitac2026}. Tacinf-Tabero focuses on converting existing manipulation trajectories into synchronized vision-tactile-language supervision and on evaluating instruction-dependent interaction forces alongside task completion.

\subsection{Tactile Simulation Methods}
Vision-based tactile simulation must account for contact geometry, sensor deformation, and image formation. TACTO couples physics-engine contact information with efficient optical rendering~\cite{tacto}. Taxim calibrates an example-based optical mapping and models marker motion using elastic deformation and superposition; FOTS provides fast optical and marker-motion approximations for tactile-motor learning~\cite{taxim,FOTS}. These methods offer efficient sensor-observation models, although their calibration assumptions and approximations must be distinguished from a complete elastomer dynamics model. TacEx integrates soft-body contact simulation with Taxim and FOTS in Isaac Sim. TacSL provides a GPU-parallel visuotactile library with support for policy learning. Its optical rendering and marker-field extraction use the same simulated deformation, keeping the two observation streams consistent~\cite{tacex,tacsl}.

\noindent Physics-based approaches pursue more explicit contact mechanics. DiffTactile combines differentiable sensor and object simulation for system identification and trajectory optimization, while Taccel integrates contact and rigid-body dynamics in a GPU-parallel tactile simulation platform~\cite{difftac,taccel}. HydroShear additionally models path-dependent shear accumulation and stick-slip transitions~\cite{hydroshear}. Learned observation transfer offers a complementary route: SimShear generates shear-aware tactile images from simulated observations and shear conditioning, and Cross-Sensor Touch Generation translates tactile observations between sensor designs~\cite{simshear,crosssensor}. Applications such as AnyRotate, UniTac2Pose, and VT-Refine demonstrate how simulated touch can support in-hand manipulation, pose estimation, and policy refinement~\cite{anyrotate,unitac2pose,vtrefine}. Together, these studies illustrate a trade-off between contact fidelity, calibration effort, and simulation throughput, and they establish tactile sim-to-real transfer as an important evaluation target~\cite{anyrotate,simshear}. Following the modular integration exemplified by TacEx, Tacinf uses Taxim for optical observations and FOTS for marker motion. Our contribution is the calibrated integration of Taxim and FOTS into a scalable data and evaluation pipeline. We evaluate the sim-to-real transfer of language-conditioned force levels across policy variants. Compared with a simulator that generates both streams from one simulated deformation, such as TacSL, this modular approach trades per-sample consistency in deformation for the throughput needed to replay open-source trajectories at scale. We therefore use marker displacement as a cue for relative shear and direction, and derive the wrench for supervision and control directly from physics-engine contact reactions.

\subsection{Tactile Data Collection}
Learning generalizable robot manipulation policies requires diverse, high-quality demonstrations, yet collecting real-world robot interaction data remains costly and difficult to scale. Teleoperation is a widely used approach in which human operators guide physical robots to perform tasks, providing demonstrations that directly reflect the robot's observations and actions. Large-scale efforts such as DROID demonstrate the value of diverse teleoperated data~\cite{droid}. However, each collection setup requires access to a robot, and scaling collection entails additional hardware, operator time, and deployment effort. Universal Manipulation Interface (UMI) addresses these constraints by collecting demonstrations with handheld grippers and transferring the resulting skills to robots, decoupling demonstration collection from physical robot access~\cite{chi2024universal}. FastUMI further develops this direction through a scalable, hardware-independent collection interface~\cite{zhaxizhuoma2025fastumi}. For tactile learning, sensor hardware and observation alignment are equally important: GelSight provides high-resolution contact geometry and deformation measurements, PolyTouch combines complementary sensing modalities, and 3D-ViTac fuses distributed tactile measurements with visual geometry~\cite{gelsight,polytouch,vitac3d}. These approaches provide a practical foundation for expanding manipulation datasets beyond dedicated robot workstations.

\noindent Demonstrations of contact-rich manipulation must record motion, visual observations, and the contact information needed to regulate physical interaction. In teleoperation systems without real-time force feedback, operators rely primarily on visual observations to infer contact states, making it difficult to perceive and regulate interaction forces accurately. Insufficient or excessive force may result in unstable grasps, excessive object loading, or task failures. These issues can compromise demonstration quality and increase repeated attempts and unusable demonstrations, reducing collection efficiency. Moreover, collecting tactile demonstrations through teleoperation still requires sensor-equipped robots, limiting scalability. Integrating tactile sensors directly into handheld manipulation interfaces such as UMI allows operators to manipulate objects directly while synchronously recording motion, vision, and touch. Combined with direct mechanical feedback or active motor feedback, these interfaces allow operators to perceive contact states and adjust interaction forces more intuitively. ViTaMIn combines a compliant handheld gripper with visuotactile sensing and operator-perceptible force feedback, while ViTaMIn-B extends this approach to bimanual collection with tracked controllers~\cite{liu2025vitamin,li2025vitaminb}. FreeTacMan uses a wearable gripper for synchronized visuotactile demonstrations, exUMI integrates modular tactile sensing with calibrated state acquisition, and TacUMI combines tactile observations, force-torque measurements, and pose tracking~\cite{freetacman,xu2025exumi,tacumi}. UMI-FT further demonstrates the use of finger-level wrench measurements for learning compliant manipulation~\cite{choi2026wild}. Building on this direction, Tacinf-Tacumi combines tactile sensing with real-time active force feedback, enabling operators to perceive and regulate grasp effort during demonstration collection. The handheld interface and robot execution system use the same low-cost gripper motor, eliminating the need to select a separate, dedicated gripper motor for deployment. This reduces additional hardware procurement and adaptation effort, lowering deployment costs and integration complexity.

\subsection{Tactile Foundation Models}
Tactile foundation models encompass transferable perception backbones as well as policies conditioned on touch. Learning to Jointly Understand Visual and Tactile Signals and UniTouch align tactile observations with visual or multimodal representations~\cite{visuotactile,unitouch}. Sparsh studies self-supervised tactile pretraining and introduces TacBench; Transferable Tactile Transformers uses a shared backbone with sensor-specific and task-specific modules~\cite{sparsh,t3}. Sensor-Invariant Tactile Representation and AnyTouch address cross-sensor transfer, with AnyTouch explicitly combining static and dynamic tactile information~\cite{sensorinvariant,anytouch}. TacX extends pretraining beyond tactile images to multiple sensing modalities, while AnyTouch 2 emphasizes temporal deformation and force-aware dynamics~\cite{tactile_beyond,anytouch2}. These advances establish that scalable touch learning depends on sensor diversity, temporal information, and appropriate transfer evaluations, not simply on treating a tactile image as another camera view.

\noindent Tactile--language models connect contact observations to semantic reasoning. Octopi and Octopi-1.5 use tactile inputs for object-property inference and language-grounded reasoning~\cite{octopi,octopi15}. On the action side, OpenVLA, OpenVLA-OFT, $\pi_0$, $\pi_{0.5}$, RDT-1B, and GR00T N1 provide generalist vision-language-action (VLA) policy architectures and training recipes that motivate tactile extensions~\cite{openvla,OpenVla-OFT,pi0,pi0.5,RDT,groot}. TLA and VTLA study language-conditioned contact-rich manipulation, including insertion and preference-based learning~\cite{tla,vtla}. TA-VLA examines torque-aware policy design; OmniVTLA aligns heterogeneous tactile signals with semantic representations; VLA-Touch introduces tactile feedback at planning and execution levels~\cite{ta-vla,omni-vtla,vla-touch}. Tactile-VLA already combines tactile-conditioned prediction with hybrid position-force control~\cite{tactile-vla}. Related visuotactile policies such as RoboPack and Reactive Diffusion Policy use touch for dynamics inference or fast reactive control~\cite{robopack,rdp}.

\noindent Tacinf-VTLA builds on these approaches to representation learning and control. It provides a simple, modular interface for incorporating tactile observations and force supervision into foundation-model post-training. The framework supports different tactile inputs, predicts fingertip-force targets, and couples a pretrained policy to separate feedback loops for grip force and applied wrench.

\section{Tacinf}
\label{sec:tacinf}

\subsection{Tacinf-Tabero}
\label{sec:tacinf-tabero}
Tacinf-Tabero is the simulation benchmark component of Tacinf. It offers reproducible language-conditioned tactile manipulation tasks, scalable multimodal data generation, and process-aware metrics for evaluating physical interaction quality.

\subsubsection{Data Reuse}
Open-source manipulation datasets are valuable community resources. We develop a reusable workflow for transferring datasets from MuJoCo and other platforms to the Isaac Sim-based Tacinf-Tabero benchmark. This workflow improves visual fidelity through high-quality rendering and augments existing demonstrations with tactile observations, reducing the need to reconstruct each dataset from scratch.

\noindent We first reconstruct task environments in Isaac Lab~\cite{isaaclab} to match the source domains, including assets, initial object and robot poses, and success logic. This workflow can currently be partially automated with SimFoundry~\cite{simfoundry} or large language models. We then replay the source trajectories in the reconstructed scenes to obtain multimodal data that follow the demonstrated human trajectories. Transferring these trajectories to tactile manipulation settings presents two challenges. First, the end-effector differs: our robot uses a gripper integrated with visuo-tactile sensors whose geometry differs from the original. Second, the underlying physical behavior differs: some source datasets rely on Operational Space Control (OSC)~\cite{osc}, which is highly sensitive to physical parameters. Datasets collected by real-robot teleoperation or UMI devices also introduce simulation-to-real gaps. We therefore make two adjustments: (1) align the end-effector tool center point (TCP) by adjusting the robot-arm base pose; and (2) use a high-gain proportional-derivative (PD) joint controller during trajectory replay to minimize cumulative tracking error.

\noindent We evaluate the success rate of trajectory replay with the tactile gripper and obtain an acceptable proportion of successful episodes. We refer to this proportion as the data retention rate.

\subsubsection{Cross-Modal Data Acquisition}
We use GPU-accelerated parallel rendering in Isaac Lab to build a scalable multimodal acquisition system with synchronized sensory streams. Visual information is obtained from two camera configurations: a wrist-mounted camera and a third-person view camera, both providing synchronized RGB-D data. Tactile observations are produced by a simulated GelSight~\cite{gelsight} sensor with a resolution of $320 \times 240$, providing both RGB tactile images and an $11 \times 9$ marker grid. We also support Xense sensors with a marker grid of 20 rows and 11 columns; image resolution is specified by the sensor configuration. For efficient tactile simulation, we use the computationally lightweight methods of Taxim and FOTS~\cite{taxim,FOTS}.

\noindent\textbf{Tactile Image.}
Tactile images are generated using the Taxim framework~\cite{taxim}. GelSight builds on retrographic sensing~\cite{ps} and reconstructs contact geometry from calibrated illumination-dependent images~\cite{gelsight}; simulation evaluates the forward mapping from geometry to image. Taxim shifts the depth map according to the indentation depth, approximates static gel deformation using multiple layers of Gaussian filtering, and computes surface normals for optical rendering.
\noindent\textbf{Marker Displacement Field.} Following the FOTS approach~\cite{FOTS}, we model the displacement field of surface markers on an elastic tactile sensor. Given simulated contact deformation and relative-motion cues, Gaussian spatial decay functions approximate marker displacement caused by normal expansion, shear, and planar torsion. The resulting field is expressed in sensor image coordinates and provides a raw signal that contains information about shear forces and torques. This marker-field interface can be extended from optical tactile sensors to capacitive, piezoelectric, or magnetic sensors.

\noindent\textbf{Contact Force.} The physics engine provides the left and right sensor reactions, $\mathbf{f}_L^{S_L}$ and $\mathbf{f}_R^{S_R}$, in their local frames. After rotation into the TCP frame, the negative sum gives the force exerted on the object. The two contact forces and their moment arms provide a point-force approximation of the TCP wrench (Section~\ref{sec:tacinf-vtla}); the scalar grip proxy is computed separately from the contact force magnitudes.

\noindent All cameras are rendered in parallel using tiled rendering, and all modalities, including visual, tactile, force, language instructions, and executed actions, are sampled synchronously at 20 Hz to produce temporally aligned multimodal observations at each time step.

\subsubsection{Language-Conditioned Force-Level Data Generation}

We focus on tasks in which the desired interaction cannot be specified adequately by task completion alone, including explicit force instructions and contact-sensitive object handling. We distinguish two categories: (1) language-conditioned force-level tasks; and (2) manipulation of deformable objects or objects with articulated parts. Tabero is primarily used to validate the first category, while we validate the corresponding behavior for the second category directly in real-world scenarios.

\noindent For the first category of tasks, most open-source datasets use continuous actions for arm control but binary or discrete gripper commands, limiting fine-grained force regulation and producing narrow within-task force distributions. We vary the low-level gripper controller parameters to execute the same task at different force magnitudes.

\noindent In our simulation, the gripper uses a spring-damper position servo, consistent with the impedance-control formulation~\cite{impedance}. For a single-DOF gripper with width $p$, the commanded force is:
\begin{equation}
    F^{\text{cmd}} = K_p (p^{\text{target}} - p) + K_d (\dot{p}^{\text{target}} - \dot{p}).
\end{equation}
At steady state after contact, velocities vanish and the static grip force approximates $F_{\text{static}} \approx K_p \cdot \delta$, where $\delta = p - p^{\text{target}} > 0$ is the penetration depth. Thus, varying $K_p$ scales the steady-state force under identical high-level commands. During initial contact, however, the impact force is dominated by damping: $F_{\text{impact}} \propto K_d \cdot |\dot{p}|$, so adjusting $K_d$ diversifies transient forces. This variation is difficult to realize directly with many real robot grippers, motivating the tactile UMI device described below for data collection.

\noindent We replay trajectories under different $(K_p, K_d)$ settings to instantiate an ordered force-level condition $\ell$. In the reported benchmark, $\ell\in\{10\%,25\%,100\%\}$, with the same task and trajectory semantics retained across levels. These interactions produce distinct sensor-side force patterns $\mathbf{f}_{L}^{S_L}$ and $\mathbf{f}_{R}^{S_R}$; their TCP-frame transforms and object-side signs follow Eq.~\eqref{eq:force_frames}. Following Tactile-VLA~\cite{tactile-vla}, we attach language such as ``gently'' or ``softly'' to low-force conditions and ``firmly'' or ``tightly'' to high-force conditions, aligning the language condition with the measured force distribution. The objective is to preserve the ordering and separation of force levels induced by $\ell$, with ``gentle'' corresponding to one end of this range. For special tasks whose nature already demands substantial force, such as wiping a blackboard, we do not add an adverb. We record gripper aperture $p$ as a continuous variable, keeping the action space compatible with the force dynamics. This pipeline provides a reusable way to generate language-conditioned force profiles for arbitrary tasks.

\begin{figure}[t]
  \centering
  \includegraphics[width=0.95\textwidth]{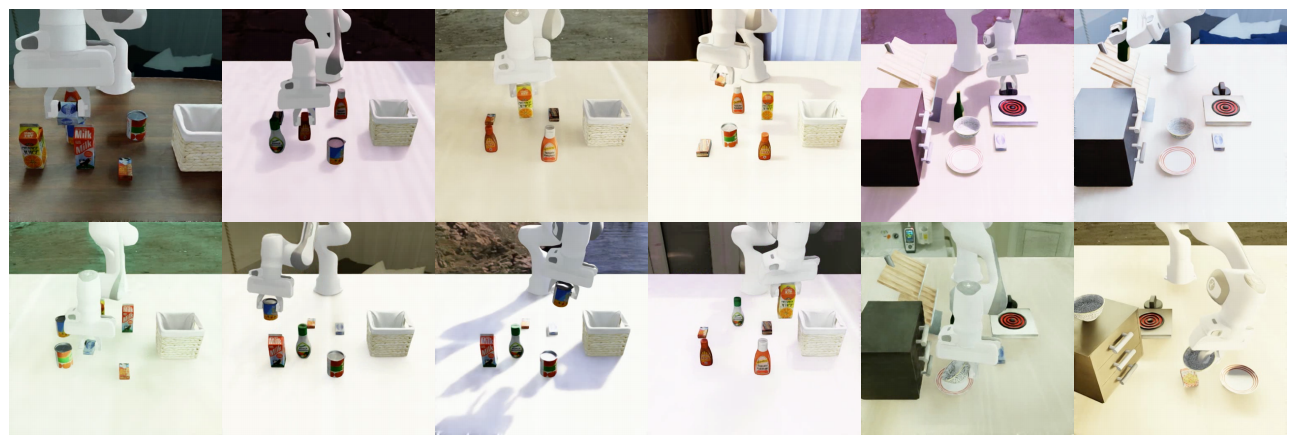}
  \caption{\textbf{Tacinf-Tabero simulation benchmark.} Tacinf-Tabero replicates the LIBERO task environments, enables data reuse, enhances the visual fidelity of simulated data, and provides high-quality tactile observations.}
  \label{fig:sim_platform}
\end{figure}

\subsubsection{Metrics Beyond Success Rate}
Task success alone overlooks critical aspects of physical interaction; force-aware and deformation-aware benchmarks address complementary aspects of this issue~\cite{forcesafety,softvtbench}. To reduce computational cost, we approximate contact using rigid-body simulation with force sensing. We evaluate the resulting fingertip forces to quantify physical interaction quality during task execution. The two grip-force metrics use $F_{\text{grip}}$, derived from the squeeze proxy in Eq.~\eqref{eq:squeeze_force} using the relation in Eq.~\eqref{eq:grip_force}:
\textbf{Maximum Transient Grip Force (MG).}
The average of the top 5\% values of $F_{\text{grip}}$ over an episode, capturing peak grasping effort and transient spikes that may indicate aggressive or damaging behavior.
\textbf{Average Grip Force (AG).}
The mean stable grip force $F_{\text{grip}}$ across both fingers during intervals without relative slip between the object and fingertips. It provides a reference estimate of tangential squeezing force; it does not represent absolute ground truth.
\textbf{Maximum Transient Applied Force (MA).}
The average of the top 5\% magnitudes of the applied force, characterizing extreme interaction events such as impacts.
\textbf{Average Applied Force (AA).}
The mean applied force magnitude during contact, measuring overall interaction intensity.

\noindent With wrench-aware control, the three translational forces and the two modeled torque components about TCP $x$ and $z$ are additionally tracked in terms of their means, peaks, and tracking errors. Under the sufficient-grip, no-slip assumption below, object-dependent torque about the grasp axis is excluded from this sensor-centered evaluation. Together, these metrics enable a more nuanced and physically grounded evaluation of robot policies, moving beyond binary success to assess safety and interaction quality.

\subsection{Tacinf-VTLA}
\label{sec:tacinf-vtla}
Tacinf-VTLA provides shared interfaces for adapting tactile observations, supervising fingertip-force targets, and executing policies with feedback. These interfaces allow different tactile policies to be trained, compared, and deployed under common data and task protocols. We instantiate the framework using the OpenPi infrastructure for $\pi_0$ and $\pi_{0.5}$~\cite{pi0,pi0.5}, focusing on system reuse and the empirical contribution of each component. In this implementation, the policy predicts fingertip-force targets from vision, language, and touch, and a closed-loop force-position controller converts its outputs into compliant robot motion.

\subsubsection{Fingertip Force Decomposition and Two-Contact Wrench Mapping}
The separation of internal grasp forces from the resultant object wrench is standard in grasp mechanics; recent compliant imitation-learning systems also distinguish grasp and external interaction forces~\cite{choi2026wild}. Our force decomposition also draws on the force-constrained formulation used in gentle-manipulation policy learning~\cite{zhou2026gentle}. In our implementation, a two-finger parallel gripper has a single opening and closing degree of freedom, so the planar contact at its fingertips decomposes naturally into two parts: the components squeezing each other along the closing axis form the \emph{grip force}, while the remaining components constitute the \emph{applied interaction} on the object, represented as an estimated six-dimensional wrench at the TCP under a two-contact approximation (3D force plus a 3D moment vector whose grasp-axis component is neglected). During grasping, the normal force that compresses the sensor pads against the object surface dominates; when the gripper transports, wipes, or twists the object, shear forces and moments also influence the object motion.

\subsubsection{Hybrid Controller}
Building on hybrid position-force control and impedance-based interaction control, as well as their use in tactile-conditioned policies, we implement force-error-driven pose and gripper-width corrections. We distinguish a scalar squeeze channel from the applied wrench, which comprises a 3D force and a 3D torque. Let $i\in\{L,R\}$ index the left and right fingertips, $S_i$ denote each sensor frame, and $T$ denote the TCP frame. A superscript $S_i$ or $T$ specifies the coordinate frame, whereas $\top$ denotes transpose. Lowercase $\mathbf{f}_i^{S_i}\in\mathbb{R}^3$ is the contact reaction acting on sensor $i$; its local $z$-axis is normal to the pad. The squeeze force is a one-dimensional scalar computed from the full contact reactions:
\begin{equation}
    F_{\text{squeeze}} = \min\big( \|\mathbf{f}_{L}^{S_L}\|,\; \|\mathbf{f}_{R}^{S_R}\| \big)\in\mathbb{R}_{\geq 0}.
    \label{eq:squeeze_force}
\end{equation}
We assume the object is at the threshold of relative sliding while remaining stationary. Because the pressure directions at the two fingertips are not necessarily collinear, we define the force required to keep the object stationary, $F_{\text{static}}$, in terms of the grip force $F_{\text{grip}}$, the friction coefficient $\mu$, and the squeeze proxy as
\begin{equation}
    F_{\text{static}} = F_{\text{grip}}\mu = 2F_{\text{squeeze}}\mu
    \approx \big(\|\mathbf{f}_{L}^{S_L}\|+\|\mathbf{f}_{R}^{S_R}\|\big)\mu.
    \label{eq:grip_force}
\end{equation}

\begin{figure}[t]
  \centering
  \includegraphics[width=\textwidth]{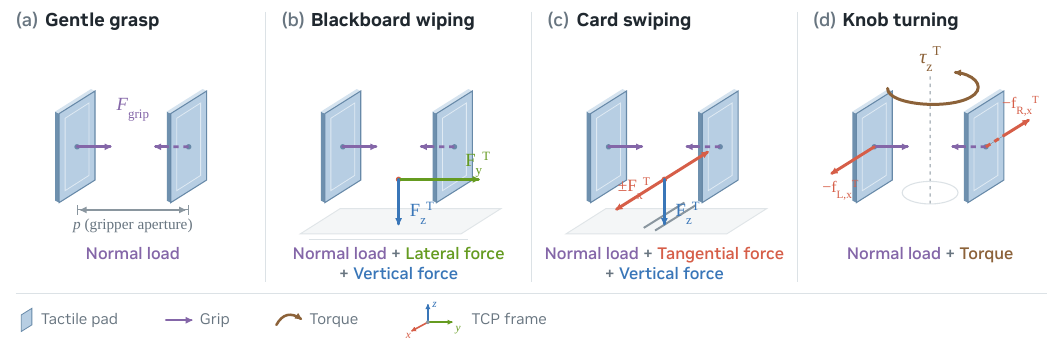}
  \caption{\textbf{Fingertip-force decomposition across tactile skills.} Purple arrows denote internal grip forces, and colored arrows denote TCP forces and torques. Arrows indicate loads acting on the object; dashed segments mark portions occluded by a tactile pad. (a) Gentle grasping, (b) blackboard wiping, (c) card swiping, and (d) knob turning.}
  \label{fig:force_decomposition}
\end{figure}

\noindent The TCP frame is denoted by $T$. Lowercase $\mathbf{f}_L^T,\mathbf{f}_R^T$ are the left and right sensor reactions expressed in this common frame, and capital $\mathbf{F}^T$ denotes the single resultant force exerted on the object:
\begin{equation}
    \mathbf{F}^T = -(\mathbf{f}_L^T+\mathbf{f}_R^T).
    \label{eq:force_frames}
\end{equation}
The sensor-local normal axes map to opposite directions along TCP $y$. The scalar aperture $p$ is the distance between the two tactile contact planes along $y$. Approximating the contacts by the pad centers, at $(0,-p/2,0)^\top$ and $(0,p/2,0)^\top$ relative to the TCP, gives the moment directly from the two forces:
\begin{equation}
    \boldsymbol{\tau}^{T}\approx
    \frac{p}{2}\big(f_{L,z}^{T}-f_{R,z}^{T},\;0,\;f_{R,x}^{T}-f_{L,x}^{T}\big)^{\top}.
    \label{eq:applied_torque}
\end{equation}
We assume the grip force is sufficient to maintain stable contact without relative slip between the object and tactile pads. Under this assumption, torque about the grasp axis $y$ still depends on the object and its contact conditions. To keep the calculation centered on the sensors and known gripper geometry, we leave this object-dependent component outside the modeled wrench and its feedback correction. The zero entry for the $y$-moment in Eq.~\eqref{eq:applied_torque} indicates that this component is omitted from the model; the physical torque may be nonzero. Fig.~\ref{fig:force_decomposition} illustrates the decomposition across gentle grasping, blackboard wiping, card swiping, and knob turning.

\noindent\textbf{Wrench-Aware Control}
The policy predicts an end-effector pose $\mathbf{P}^{\text{pred}}\in SE(3)$ and target sensor reactions in the two local frames. The controller rotates each target and measurement into the TCP frame using rotation matrices before subtraction:
\begin{equation}
\begin{aligned}
    \mathbf{e}_L^T &= \mathbf{R}_{S_L}^{T}
    (\mathbf{f}_L^{S_L,\text{target}}-\mathbf{f}_L^{S_L}),\\
    \mathbf{e}_R^T &= \mathbf{R}_{S_R}^{T}
    (\mathbf{f}_R^{S_R,\text{target}}-\mathbf{f}_R^{S_R}).
\end{aligned}
\end{equation}
The sum and difference of the two fingertip errors give the force-driven pose corrections directly:
\begin{equation}
\begin{aligned}
\Delta\mathbf{p} &= -\mathbf{K}_{p}\mathbf{R}_{T}^{B}
    (\mathbf{e}_L^T+\mathbf{e}_R^T),\\[2pt]
\Delta\boldsymbol{\phi} &= \mathbf{K}_{\phi}\mathbf{R}_{T}^{B}
    \begin{bmatrix}
      \tfrac{p}{2}(e_{L,z}^{T}-e_{R,z}^{T})\\
      0\\
      -\tfrac{p}{2}(e_{L,x}^{T}-e_{R,x}^{T})
    \end{bmatrix},\\[2pt]
\mathbf{p}^{\text{cmd}} &= \mathbf{p}^{\text{pred}}+\Delta\mathbf{p},\qquad
\mathbf{R}^{\text{cmd}} = \exp\!\big([\Delta\boldsymbol{\phi}]_{\times}\big)\mathbf{R}^{\text{pred}}.
\end{aligned}
    \label{eq:wrench_pose_correction}
\end{equation}
Here $\mathbf{R}_{T}^{B}$ rotates from TCP to base coordinates while moments remain referenced to the TCP. The gains $\mathbf{K}_{p}$ and $\mathbf{K}_{\phi}$ have units $\mathrm{m/N}$ and $\mathrm{rad/(N\!\cdot\!m)}$, respectively, and $[\,\cdot\,]_\times$ denotes the cross-product matrix. The controller corrects the pose predicted by the policy using the force-feedback corrections in Eq.~\eqref{eq:wrench_pose_correction}. The rotation vector $\Delta\boldsymbol{\phi}$ updates orientation through the exponential map. Differential inverse kinematics converts the corrected pose into joint velocity commands, while the independent grip loop adjusts aperture.

\noindent\textbf{Grip Force Control}
To overcome the limitation of position control in precise force regulation, we implement a grip force feedback loop.
The predicted sensor-frame reactions $\mathbf{f}_i^{S_i,\text{target}}$ give the contact force magnitudes, yielding the squeeze proxy and then $F^{\text{pred}}_{\text{grip}}$ through Eqs.~\eqref{eq:squeeze_force} and~\eqref{eq:grip_force}; the grip-loop target is
\begin{equation}
    F^{\text{target}}_{\text{grip}} = (1 + k_{\text{ff}}) \, F^{\text{pred}}_{\text{grip}}, \quad k_{\text{ff}} > 0,
\end{equation}
where $k_{\text{ff}}$ is a feedforward gain.
The measured grip force $\tilde{F}_{\text{grip}}$ is exponentially smoothed.
Let $p$ denote the gripper width.
We apply a correction $\Delta p$ only when $|\Delta p| \geq \text{dz}$, where $\text{dz} > 0$ is a deadzone threshold accounting for gripper imprecision:
\begin{equation}
    \Delta p = k^{\text{adm}}_{p} \cdot (F^{\text{target}}_{\text{grip}} - \tilde{F}_{\text{grip}}),
\end{equation}
with $k^{\text{adm}}_{p}$ an admittance gain.
The final gripper command is
\begin{equation}
    p^{\text{cmd}} = \operatorname{clip}\bigl( p^{\text{pred}} + \Delta p,\; 0,\; p_{\max} \bigr),
\end{equation}
where $p^{\text{pred}}$ is the policy-predicted gripper width.
This allows the policy to specify a desired grip force while the controller achieves accurate tracking via feedforward and force feedback. The low-level controller thus uses the two fingertip force vectors and the gripper aperture. Contact-force magnitudes determine the grip-force proxy, while the sum and the difference scaled by half the aperture determine the wrench-based pose corrections. The policy coordinates these commands at the task level.

\subsubsection{Tactile Policy}
\begin{figure}[htbp]
  \centering
  \includegraphics[width=0.45\textwidth]{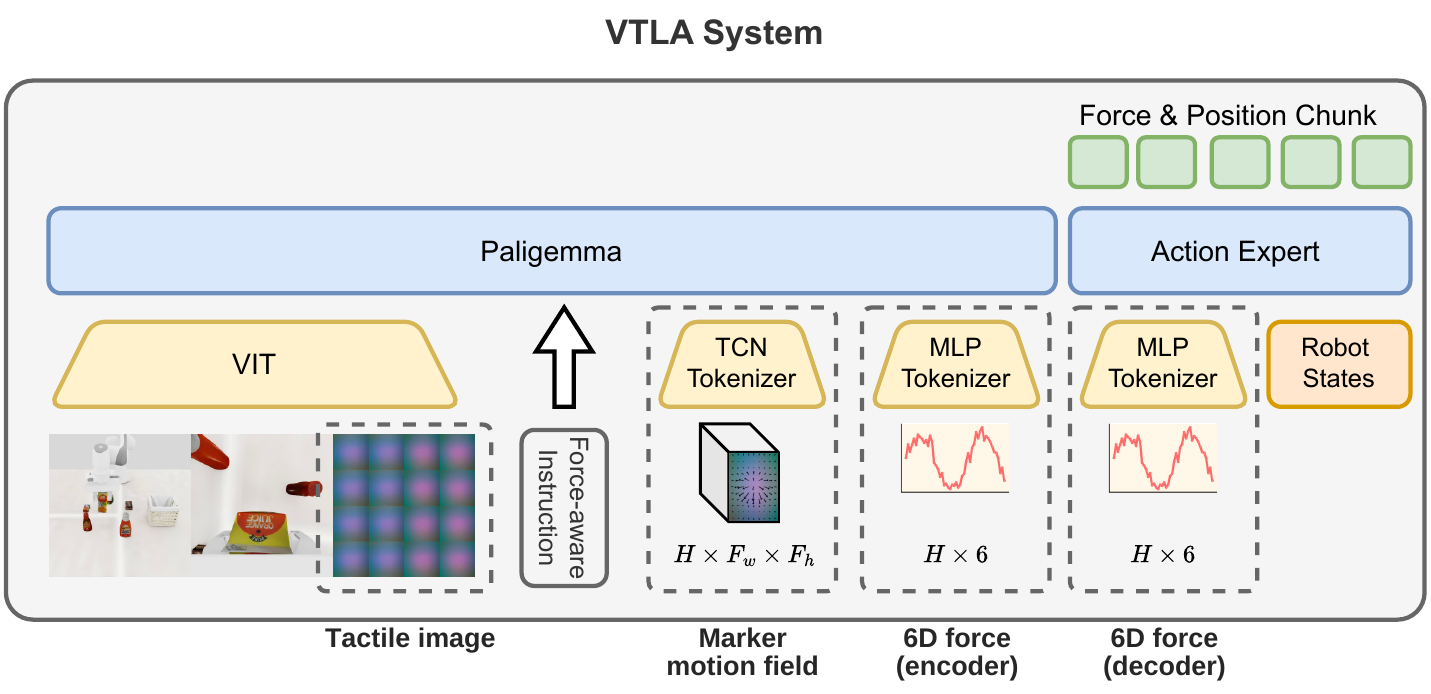}\hfill
  \includegraphics[width=0.45\textwidth]{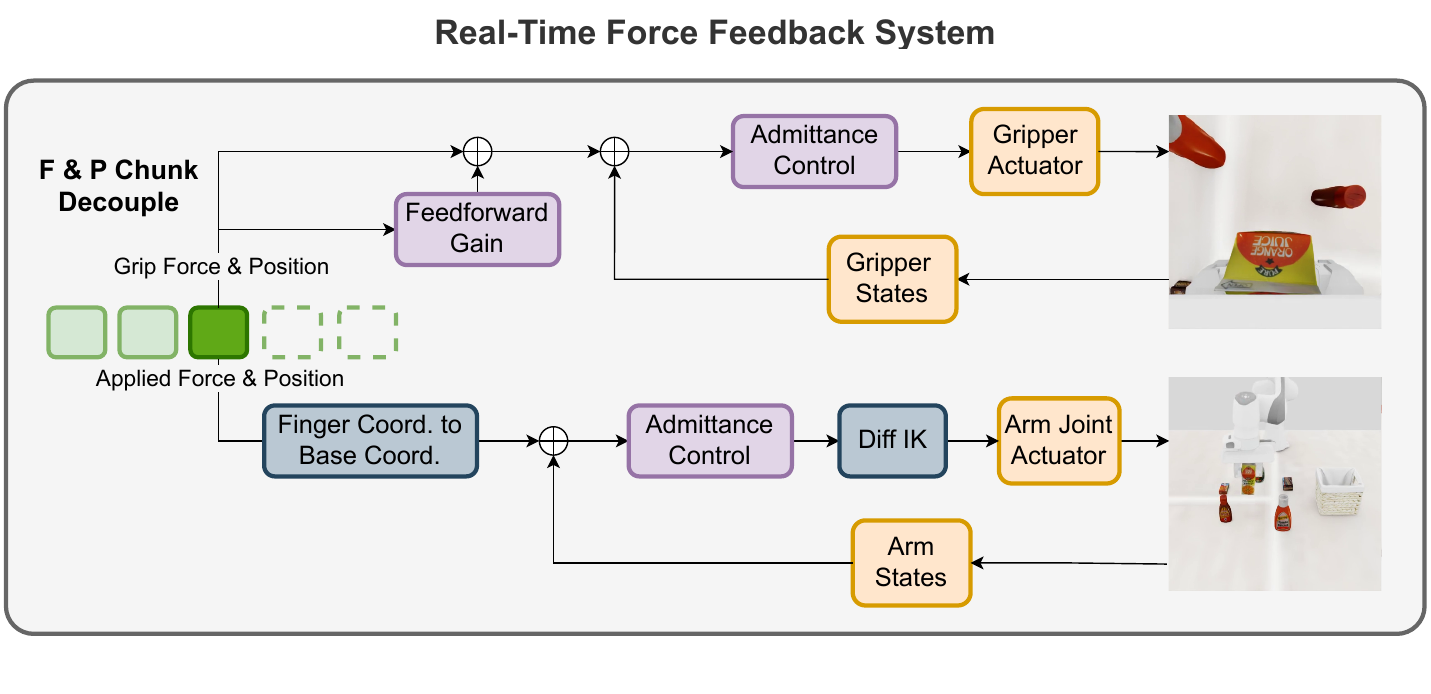}
  \caption{\textbf{Tacinf-VTLA system overview.} Tactile inputs are encoded by input adapters and fused with vision and language. The policy predicts force-position commands, which are tracked by separate grip-force and wrench feedback loops to achieve compliant interaction. IK = inverse kinematics.}
  \label{fig:model_control}
\end{figure}

Tacinf-VTLA is trained on the tactile representations and force-aware instructions generated by Tacinf-Tabero. To integrate tactile signals into the VLA foundation model, we evaluate three tactile input adapters that map different tactile modalities into conditional tokens. The high-level policy then jointly predicts future gripper poses and six-dimensional fingertip-force setpoints based on vision, language, and touch. The fixed admittance-based low-level controller described above tracks these commands and provides physical compliance (Fig.~\ref{fig:model_control}). The high-level policy learns to infer appropriate interaction forces from the task context. Using flow matching within OpenPi, the policy predicts continuous pose commands and six fingertip-force components. The three adapters share the same temporal window, preprocessing, backbone, and training protocol; only the tactile input interface is varied in the ablation. Below, we detail the adapters and loss function, and compare alternative tactile injection strategies inspired by prior work.

\noindent\textbf{Marker-Field Adapter.}
The marker-field adapter processes marker displacement relative to the sensor's initial rest state. The input consists of $H+1$ frames: the first frame captures the undeformed marker layout, and the subsequent $H$ frames record 2D marker positions in the sensor's local plane, resulting in an array $M \in \mathbb{R}^{(H+1) \times N \times 2}$. Here, $N$ is the number of markers, and each 2D entry contains a marker's planar coordinates. A lightweight temporal convolutional network (TCN) encodes this spatiotemporal sequence into tokens for integration into the transformer backbone. Appendix~\ref{App:hyper} summarizes the evaluated variants.

\noindent\textbf{Tactile-Image Adapter.}
Inspired by visuotactile policy conditioning~\cite{vtla,rdp}, we implement an image adapter for our backbone. Specifically, we construct a single composite tactile image by arranging the $H$ historical frames from the left finger and the $H$ frames from the right finger into a unified spatial layout. The value of $H$ is determined by the shared multi-view image resolution and the prescribed tactile-image resolution; when $H=1$, the composite contains only the current contact state. This image, encoding the full tactile history of both fingertips, is processed by the same visual encoder used for RGB-D observations. Its features then interact with visual features via cross-attention in the transformer, so the model can reason jointly about contact and scene geometry. Appendix~\ref{App:hyper} summarizes the evaluated variants.

\noindent\textbf{Force Adapter.}
We use the 3D sensor-frame reaction vectors $\mathbf{f}_{L}^{S_L}$ and $\mathbf{f}_{R}^{S_R}$ as raw inputs, resulting in a 6-dimensional force signal. Although these reactions can be transformed and combined with the known fingertip geometry to compose the applied wrench at the TCP, we find it more effective to feed their concatenation into a multilayer perceptron (MLP) to obtain a compact latent representation of contact dynamics. This representation is injected into the model alongside other modalities to enhance physical awareness. Appendix~\ref{App:hyper} summarizes the evaluated variants.

\noindent\textbf{Force Supervision.}
Flow matching~\cite{pi0} supports continuous commands, including gripper position and fingertip forces. Each normalized target contains 13 effective components: seven motion and gripper-control components and six sensor-frame fingertip-force components, zero-padded to the output width $D=32$. Let $B$ be the batch size, $K$ the future prediction horizon, and $e_{b,k,d}=v_{b,k,d}-u_{b,k,d}$ the difference between predicted and target flow velocities for sample $b$, prediction step $k$, and output dimension $d$. The weighted loss is
\begin{align}
    \mathcal{L} = \frac{1}{BKD}\sum_{b=1}^{B}\sum_{k=1}^{K}\sum_{d=1}^{D}w_d\,e_{b,k,d}^{2},
\end{align}
where $w_d=1$ for $1\leq d\leq7$, $w_d=\lambda_{\mathrm{force}}$ for $8\leq d\leq13$, and $w_d=\lambda_{\mathrm{pad}}$ for the remaining padding dimensions, with nonnegative weights. The loss averages over batch samples, prediction steps, and the full output width, retaining $D$ in the denominator even when padding has zero weight. The batch average estimates the expectation over sampled flow times and Gaussian noise.

\subsection{Tacinf-Tacumi}
\label{sec:tacinf-tacumi}
We develop Tacinf-Tacumi, a handheld multimodal interface for collecting demonstrations of contact-rich manipulation. Following the robot-free teaching paradigm of UMI~\cite{chi2024universal}, the system combines tactile sensing with real-time active force feedback, so operators can perceive and regulate grasping load while directly manipulating objects. It synchronously captures end-effector motion, gripper aperture, and contact observations to provide multimodal demonstrations for tasks requiring fine force regulation.

\begin{figure}[t]
  \centering
  \includegraphics[width=\textwidth]{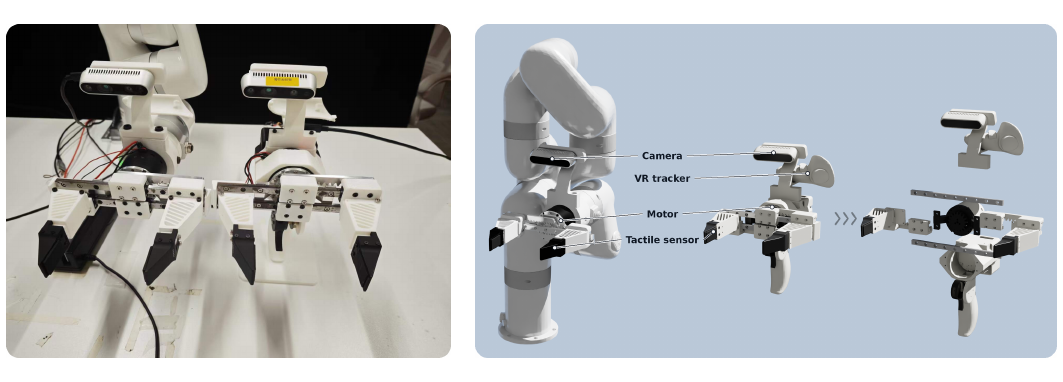}
  \caption{\textbf{Tacinf-Tacumi hardware.} Left: the real handheld collection device, with its quasi-direct-drive (QDD) actuator, wrist RGB camera, and tactile fingertips. Right: structural renderings of the arm-mounted gripper, the handheld device, and an exploded view of the handheld device, shown at the same scale. Labels identify the shared camera, tactile sensors, and motor, as well as the virtual reality (VR) controller used only on the handheld device for UMI pose tracking.}
  \label{fig:tacumi_interface}
\end{figure}

\noindent As shown in Fig.~\ref{fig:tacumi_interface}, Tacinf-Tacumi comprises a motorized trigger, a parallel gripper, and multimodal sensing components. The trigger operates under field-oriented control (FOC) and serves as both the opening and closing command interface and the active feedback actuator. The parallel gripper is driven by a DAMIAO motor that provides position and load feedback, while fingertip-mounted Xense vision-based tactile sensors provide tactile images and contact-force estimates. A VR controller supplies six-degree-of-freedom pose tracking, and a camera records visual observations. The acquired signals and task instructions are organized under a common data schema, allowing handheld demonstrations to be combined with robot teleoperation and simulation trajectories for the mixed-data training evaluated in Section~\ref{sec:eval-cotrain}.

\subsubsection{Fine Force Regulation}
The handheld interface uses position commands and torque feedback to establish a bidirectional interaction loop, following the position-force architecture of bilateral interfaces~\cite{lawrence1993bilateral}. In the command channel, trigger rotation is mapped to a target gripper aperture and converted into a motion command for the DAMIAO motor. The actual aperture is obtained from motor position feedback through the transmission relationship. The target and actual apertures represent operator intent and gripper execution state, respectively.

\noindent In the feedback channel, the torque estimate reported by the gripper motor is mapped to a feedback command for the trigger:
\begin{equation}
    \tau_{\mathrm{trigger}}^{\mathrm{cmd}} = k_{\mathrm{fb}}\hat{\tau}_{\mathrm{gripper}},
    \label{eq:torque_feedback}
\end{equation}
where $\hat{\tau}_{\mathrm{gripper}}$ denotes the estimated gripper motor torque and $k_{\mathrm{fb}}$ is a signed feedback gain. Its direction follows the coordinate conventions of the two motors so that the grasping load produces a reaction torque opposing further trigger closure. The trigger motor executes the command in Eq.~\eqref{eq:torque_feedback} through FOC~\cite{bhardwaj2013foc}, rendering changes in gripper load as resistance perceptible to the operator and supporting continuous force regulation through real-time interaction feedback.

\noindent Motor torque feedback and tactile force estimation serve complementary roles. Motor torque includes contributions from transmission friction and actuator dynamics~\cite{letien2008friction}; it conveys load variations but cannot directly measure fingertip contact force. The Xense sensors provide contact-force estimates using the manufacturer's calibration, supporting contact observations, real-robot force statistics, and the grip-force feedback controller in Section~\ref{sec:tacinf-vtla}. This design combines active feedback for the operator with contact sensing for policy learning and robot control while preserving the distinct physical meanings of the two signals.

\subsubsection{Data Collection}
The system separately acquires the gripper's six-degree-of-freedom rigid-body pose and single-degree-of-freedom aperture to represent end-effector motion and grasp state. The tracking design considers trajectory reliability, system integration complexity, and deployment overhead across collection environments.

\noindent Existing handheld collection systems employ different pose-acquisition strategies. UMI estimates trajectories from wrist-camera observations and inertial measurements, but its original paper acknowledges a dependence on sufficient environmental texture, with limitations in texture-deficient settings such as rooms with plain white walls~\cite{chi2024universal}. TacUMI integrates a VIVE Tracker into the handheld device, providing an example of acquiring manipulation trajectories with a dedicated tracking device~\cite{tacumi}. For Lighthouse-based implementations such as VIVE Tracker~3.0, external base-station installation and spatial configuration introduce additional preparation when moving between collection sites~\cite{htcvive_tracker3}. ViTaMIn-B instead adopts Meta Quest controllers and reports a higher proportion of valid demonstrations than the simultaneous localization and mapping (SLAM) approach used in UMI on its Weight Placement task, providing empirical support for consumer VR tracking in handheld collection~\cite{li2025vitaminb}.

\noindent Building on these precedents, we adopt the commercial headset-and-controller tracking system of Meta Quest~3, separating end-effector pose acquisition from wrist-camera trajectory reconstruction. This solution provides real-time poses without external tracking base stations, reducing site configuration requirements and facilitating portable demonstration collection across everyday environments. The controller is rigidly attached to the handheld device, and its tracked pose is transformed into the gripper tool center point (TCP) pose through a fixed extrinsic transformation, consistent with the controller-to-end-effector mapping used in prior handheld interfaces~\cite{li2025vitaminb}:
\[
\mathbf{T}_{T}^{W}(t)
=
\mathbf{T}_{C}^{W}(t)\,
\mathbf{T}_{T}^{C},
\]
where $W$, $C$, and $T$ denote the tracking reference frame, Quest controller frame, and TCP frame defined in Section~\ref{sec:tacinf-vtla}, respectively. The fixed extrinsic $\mathbf{T}_{T}^{C}$ maps coordinates from $T$ to $C$ and is estimated from 20 paired robot and controller poses using hand-eye calibration, as detailed in Appendix~\ref{app:tacumi_handeye}.

\noindent Gripper aperture is computed from motor position feedback through the transmission relationship, allowing actuation and state acquisition to share the same hardware. End-effector pose, gripper aperture, control commands, load feedback, and visual and tactile observations are aligned on a common timeline to form multimodal trajectories that preserve the correspondence between actions and contact information. The native acquisition rates and common host timestamp convention are described in Appendix~\ref{app:tacumi_sync}.

\subsubsection{Low Cost and Actuator Consistency}
Prior interfaces address collection-to-deployment differences through hardware design, calibration, or adaptation~\cite{mimictouch,dexumi,kinedex,tactalign}. Tacinf-Tacumi uses the same low-cost gripper motor for handheld collection and robot execution, reusing the gripper actuation design and its command and state interfaces. During deployment, the robot operates the corresponding motorized gripper, while the trigger serves only as the human interface during demonstration collection. This design avoids selecting a different, dedicated gripper motor for deployment, reducing hardware selection, control-interface adaptation, and system integration effort.

\noindent Reusing the actuation design helps reduce control and calibration differences introduced by actuator replacement. Its cost advantage derives from the low-cost motor selection and design reuse across collection and execution. With a shared actuation design, contact responses can still differ between collection and deployment. Transmission geometry, fingertip geometry, sensor mounting, and control parameters require matching or calibration. Within this scope, the design reduces additional hardware and engineering requirements for deployment while providing a basis for consistency between the demonstration data and the robot execution interface.

\subsection{Tacinf-Taco}
\label{sec:tacinf-taco}
\begin{figure}[t]
  \centering
  \includegraphics[width=\textwidth]{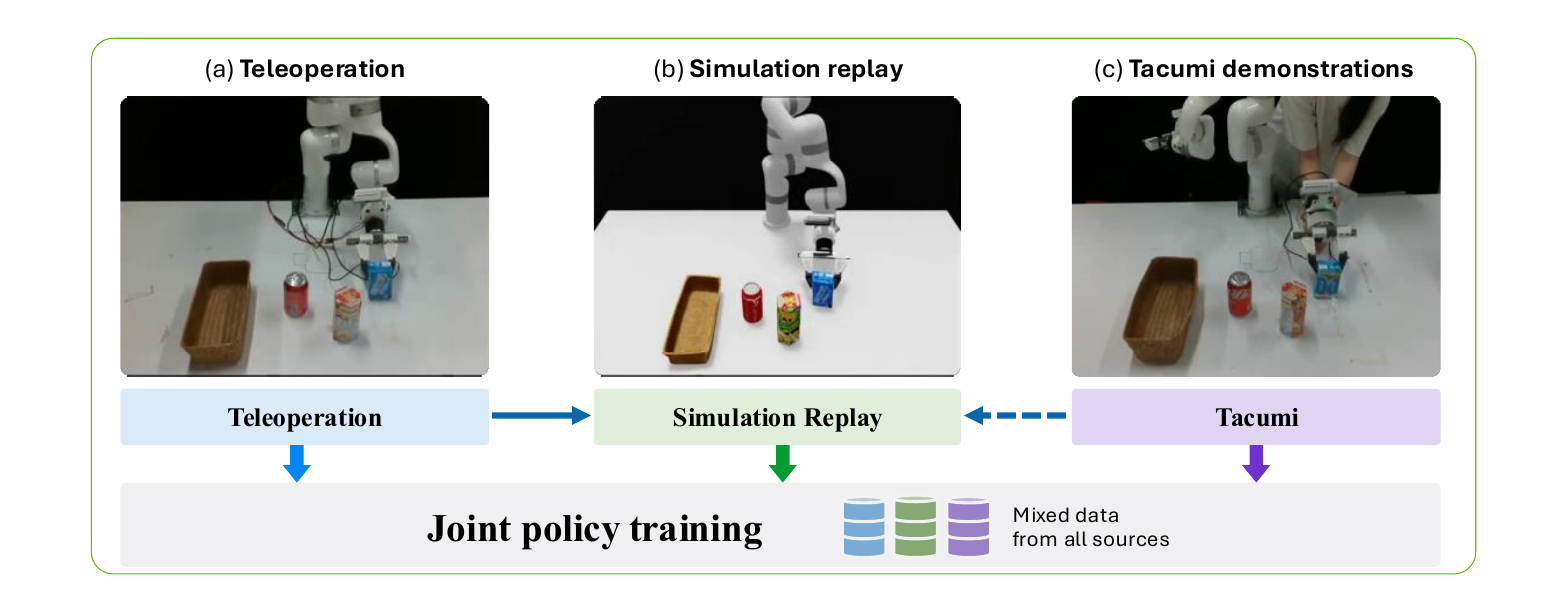}
  \caption{\textbf{Tacinf-Taco data sources and co-training.} (a) Third-person camera view of real-robot teleoperation. (b) Simulation replay of the same trajectory in the configured Tacinf-Taco scene, with Vitasoy, Coca-Cola, cookie, and basket. (c) Tacumi demonstration: the UMI camera view at the moment of grasping the Vitasoy box. Simulation replays teleoperation trajectories; Tacumi demonstrations are collected independently in parallel. Training uses all three sources.}
  \label{fig:taco_data_loop}
\end{figure}

Tacinf-Taco (Tactile policy Co-training) combines teleoperation, simulation, and Tacumi (UMI) demonstrations for VTLA post-training. Simulation data are generated by replaying teleoperation trajectories in task-aligned Tacinf-Tabero scenes. Tacumi demonstrations provide a parallel real-data source with varied object positions, orientations, and approach motions. All three sources are mixed to train one policy. Collection effort decreases from teleoperation to Tacumi to simulation. Fig.~\ref{fig:taco_data_loop} summarizes these sources and their relation.

\noindent The framework consists of three dependent parts:
\begin{enumerate}
    \item \textbf{Task-aligned scene reconstruction:} transfer workspace appearance and interactive-object geometry into a common task frame, providing the assets for controlled visual variation.
    \item \textbf{Manipulator-system Real2Sim:} align the robot, cameras, tactile sensing, and action interfaces so that replay produces synchronized observations and force targets compatible with real data.
    \item \textbf{Augmentation and post-training:} combine simulation-based visual and force-level variation with UMI-based spatial variation, then return the resulting VTLA policy to the shared physical interface.
\end{enumerate}

\noindent The aligned simulator also supports policy evaluation. The simulation-only adaptation protocol and paired Sim/Sim2Real evaluation are specified in Sections~\ref{sec:eval-sim2real} and~\ref{sec:eval-consistency}; they are separate from the mixed-data post-training design described here.

\noindent The shared interface specification allows the framework to accommodate new assets and backbones. New datasets are mapped to a common schema for tasks, states, and success criteria; new tactile sensors supply calibrated observation models; and new pretrained VLAs provide input adapters and action heads. Each extension retains the common time step, normalization, force supervision, and evaluation interfaces.

\noindent Following real-to-sim evaluation approaches that align policy-relevant observations and control~\cite{simpler,jain2025polarisscalablerealtosimevaluations}, and tactile sim-to-real studies such as AnyRotate and SimShear~\cite{anyrotate,simshear}, Taco uses two practical criteria. First, \textbf{visual-physical separation}: Gaussian or neural appearance models are used only for rendering and never for contact; explicit collision geometry handles contact, support, and shadow reception. Second, \textbf{policy-relevant consistency}: parameters are adjusted to match the data structures, distributions, and visual observations used by the policy. Exact matching of physical parameters is often impractical because it requires extensive measurement and calibration, and some parameters are commercially restricted. We assess policy-relevant consistency through checks of contact onset, contact-patch persistence, force scale, marker response, and action replay.

\subsubsection{Task-Relevant Scene Reconstruction}
This module transfers the real workspace and interactive objects into Isaac Sim, so that rendering and interaction use the same camera extrinsics and operation region. The workspace is represented visually by 3D Gaussians, while objects enter the physical layer as independent textured meshes. Both are aligned to the task coordinate system and obey the visual-physical separation principle.

\noindent\textbf{Task-Aligned Workspace Reconstruction.}
We represent workspace appearance with 3D Gaussian Splatting (3DGS)~\cite{gsplat} and convert it into an Isaac Sim-loadable NuRec Volume. The task-side robot, white collision table, cameras, and operation region remain unchanged; the Gaussian asset is aligned to the task frame using a single registration matrix $\mathbf{T}_{\mathrm{reg}}$. We use the Odin1 spatial camera from Liuxing Technology for reconstruction data capture, because in addition to a directly usable video stream, its built-in LiDAR odometry provides a more accurate reference for camera extrinsics. A spatial camera continuously records trajectories around the operation region that cover the reachable space and fields of view of the third-person and wrist cameras, while logging calibrated images, poses, and COLMAP intrinsics/extrinsics~\cite{colmap}. The reconstruction is exported as a standard 3DGS point set in a Z-up, meter-based convention. Capture is concentrated in the views used by the policy: the third-person camera frustum and the frusta swept by the wrist camera throughout the arm workspace. Uniform coverage of the room is unnecessary.

\noindent\textbf{3DGS Post-Processing.}
During scanning, the robot arm is present in the scene, so retaining its Gaussians would create double geometry with the simulated robot defined in the Unified Robot Description Format (URDF). Under a tabletop-plane constraint, we perform random sample consensus (RANSAC) plane estimation~\cite{ransac}, candidate-point selection with tabletop protection, connected-component clustering and morphological filtering, and multiple cleanup passes based on bounding-box neighborhoods, nearest-neighbor growth, and base-column constraints, while protecting fine structures such as cables wrapped around the arm. For rendering compatibility, we remove far-field outliers, contract higher-order spherical harmonics to DC-only, and ensure that the declared SH order matches the specular tensor dimensions, zero-padding missing dimensions when necessary. Finally, the scene in Universal Scene Description (USD) format uses an outer transform to cancel the task's fixed external frame and retains only $\mathbf{T}_{\mathrm{reg}}$ in the inner transform relative to the robot base.

\noindent\textbf{Object Reconstruction.}
Interactive objects are reconstructed separately and imported as independent USD assets; they are not extracted from the workspace Gaussians. For each real object, we capture one image and use Seed3D 2.0~\cite{seed3d} to generate a textured 3D model exported as USD. The explicit mesh follows Isaac's rasterization path; when Gaussian appearance and an object mesh overlap, the object asset handles physical interaction while the Gaussian layer supplies only unmodeled background. The exported USD has no collision geometry. We therefore add rigid-body properties through a USD overlay and generate collision meshes with convex decomposition without modifying the generated asset itself. The visual mesh is used only for rendering and the collision mesh only for contact; object physics is supplied by the overlay. Because Seed3D scale and initial pose can differ from the real object, we import the asset into Isaac Sim, replay the recorded real demonstrations, and iteratively adjust object scale and initial position along each axis until grasp and placement trajectories align. Calibration aims to improve agreement in trajectories and success rates between simulated replay and the real robot, beyond visual similarity alone. Task demonstrations and robot trajectories remain fixed during calibration. Before evaluation, we check the reconstructed scene with task initialization and camera observations.

\subsubsection{Embodiment Real2Sim}
Scene alignment alone is insufficient for policy transfer: the simulated embodiment must expose the same action and observation semantics as the real system. To ensure the generality of our framework, we do not reconstruct the robot arm's complete Gaussian appearance; instead, we reproduce only its dynamics and kinematics system. In addition, we explicitly simulate the shared Tacumi gripper, including its gripper control and tactile sensing.

\noindent\textbf{Manipulator System.}
We build the XArm dynamics in Isaac Lab and tune its PD parameters to match the real robot's control behavior. Hand-eye calibration provides the third-person camera pose relative to the robot base. The shared Tacumi gripper and sensor assembly determines the wrist camera's relative pose through its mechanical design, so this pose does not require estimation from an external sensor model. Finally, we tune the shared tactile gripper's passive-joint mode, PD parameters, and the force-position hybrid controller introduced above.

\noindent\textbf{Tactile Observation Modeling.}
Our tactile Real2Sim path combines the Taxim optical model~\cite{taxim} and the FOTS marker-motion model~\cite{FOTS} in a shared Isaac Lab sensor interface. The simulator first derives a contact representation from the depth image beneath the pad. Taxim maps local surface geometry to a tactile RGB observation, while FOTS combines contact deformation with contact-centroid changes and relative sensor-object yaw to produce a marker displacement field in sensor image coordinates. The two models are complementary forward models: they reproduce the signals exposed to the policy without requiring an exact replica of the commercial sensor's internal optics or elastomer mechanics. Following the calibration methods of Taxim and FOTS, we calibrate the above models using a small set of real tactile observations and, through parameter settings in Isaac Sim, align the simulated and real data streams in resolution, channel semantics, preprocessing, and temporal-update conventions. We jointly adjust the observation mapping and the contact geometry. In particular, the normal thickness of the tactile-pad collision volume is treated as an asset-level variable and tuned to reduce discrete-time interpenetration while preserving a stable contact patch and consistent activation of both tactile models. This calibration targets contact responses relevant to the policy, extending beyond visual mesh alignment. Appendix~\ref{app:tactile_real2sim} details the depth-to-indentation mapping, optical and marker interfaces, and calibration scope; the complete simulator configuration is released with the calibrated asset and pinned by the run manifest. Due to commercial restrictions, we cannot obtain the complete optical and mechanical specifications of Xense. We therefore only tune the cameras' rendered view and operating depth range at the observation level, and do not claim strict structural extrinsic registration.

\subsubsection{Data Augmentation and Joint Post-Training}
\label{sec:taco-cotrain-loop}
Taco combines three data sources for co-training: teleoperation data, Tacumi data, and simulation data. The goal is to improve the policy's generalization on real-robot tasks while preserving consistent language, contact, and action semantics across sources.

\noindent\textbf{Real seeds and simulated visual augmentation.}
Real-robot trajectories supply the original task instructions, motions, and available observations, and anchor the scene and sensor calibration. Relatively fixed object positions facilitate simulation alignment, while UMI data supply complementary spatial variation. Starting from the aligned assets, the simulation branch varies object appearance or instances and background visual elements. Building on Tacinf-Tabero, we replay the task under these variations and controlled force regimes, producing synchronized visual observations, tactile images, marker fields, contact forces, and action targets. Background appearance changes leave the collision model unchanged; replacing an interactive object's geometry requires a corresponding collision asset and a new contact rollout. Replaying the varied scene generates new tactile signals and force labels that reflect the updated physical interactions. After replay, a contact-signal check and a task-completion check filter out data that produced no contact or failed the task.

\noindent\textbf{Spatial augmentation with UMI.}
Tacumi extends spatial coverage with human demonstrations. Objects are placed at different positions and orientations and approached along different paths. The shared gripper records the resulting changes in contact and motion together, adding physical spatial variation to the simulated visual and force-level variations. Each variation in object pose is recorded as a new demonstration with corresponding actions and contact observations. Tracked UMI poses and sensor observations are expressed through the common task and observation-action interfaces before mixing.

\noindent\textbf{Joint VTLA post-training.}
The real, simulated, and UMI trajectories are synchronized, normalized, and sampled into mixed batches to jointly post-train a single pretrained Tacinf-VTLA policy with the action and fingertip-force supervision in Section~\ref{sec:tacinf-vtla}. The source sampling ratio and task/force-level coverage are explicit training settings. The post-trained policy is deployed with the same tactile gripper and hybrid controller.

\noindent\textbf{Data synergy.}
Among the three data sources, simulation data are the cheapest, UMI data come next, and teleoperation data are the most expensive. Simulation data provide visual-element and object generalization, UMI data provide grasp-approach and object-pose variation, and teleoperation data provide precise examples. The co-training study in Section~\ref{sec:eval-cotrain} evaluates these complementary contributions. The three-source mixture substantially outperforms teleoperation-only training on out-of-distribution (OOD) tasks. Position generalization is limited without UMI data, and object generalization is limited without simulation data. Teleoperation provides precise reference demonstrations for the other sources, while UMI-only training performs poorly in the evaluated setting. Together, these results illustrate the benefits of combining the three sources.

\section{Evaluation}
\label{sec:evaluation}

\noindent Sections~\ref{sec:eval-cross-platform}--\ref{sec:eval-vtla-ablation} evaluate nine tactile-input variants of the foundation model on Tabero; Sections~\ref{sec:eval-taco-align}--\ref{sec:eval-consistency} evaluate eight models retrained with Taco data; Section~\ref{sec:eval-controller} evaluates one model on separate controller data; and Section~\ref{sec:eval-cotrain} evaluates six models retrained with different data mixtures. The hardware consists of one XArm7 arm equipped with an Xense-based tactile gripper and its UMI counterpart for data collection, together with a Meta Quest 3 for UMI pose tracking. Compute devices include four H20 and two A6000 GPUs; a single RTX 4090 workstation is the minimum compute requirement for reproducing the complete system.

\noindent\textbf{Evaluation protocol.} Counting the two force levels as distinct task requirements, Tabero contains 18 base tasks and Taco contains six base tasks, six position-randomized tasks, and six tasks with both position and object randomization; three additional manipulation tasks are used to validate the controller. The task categories are language-conditioned force following and special-object manipulation. Each base Tabero and Taco task has 30 demonstrations; data quantities for mixed-data and standalone task experiments are specified separately. In simulation, each model is evaluated for 50 trials per task. On the real robot, each model is evaluated for 20 trials on force-following tasks and 10 trials on special tasks, using the criteria in Appendix~\ref{app:task-success}. Sections~\ref{sec:eval-cross-platform}--\ref{sec:eval-vtla-ablation} use the per-finger minimum in Eq.~\eqref{eq:squeeze_force} as grip force. Sections~\ref{sec:eval-taco-align}--\ref{sec:eval-cotrain} use the sum of left and right fingertip force magnitudes as an approximation because individual fingertip readings are inaccurate during contact; see Appendix~\ref{app:single_finger}. A trial is terminated when it exceeds the maximum episode length. Firm and gentle conditions of the same task share identical initial object and robot states. The real-robot policy runs at 20~Hz and executes actions asynchronously through temporal ensembling, with an inference latency of roughly 150--200~ms; the tactile closed-loop controller runs at the same 20~Hz rate.

\subsection{Data Reuse and Annotation}
\label{sec:eval-cross-platform}
We evaluate the Tacinf-Tabero data transfer and reuse workflow using LIBERO, a representative open-source dataset built on MuJoCo. We assess task success and distributional consistency after transfer to Isaac Lab. Specifically, we compare success rates for four LIBERO task subsets between the original MuJoCo dataset and our replayed trajectories in Isaac Lab. Fig.~\ref{fig:sim_platform} illustrates the range of task scenarios our simulation platform can provide. Subsequent experiments use a subset of these configurations without visual randomization.

\noindent When using the same robot kinematics and control policy as in the original dataset, our baseline configuration yields a success rate distribution that closely matches that reported in OpenVLA~\cite{openvla}. However, replacing the standard parallel-jaw end effector with a tactile-sensor-integrated gripper on the same Franka arm introduces mechanical differences that lead to a measurable drop in success rates. This degradation is especially pronounced in tasks requiring delicate manipulation, where lower grip forces strongly correlate with reduced success. The results shown in Table~\ref{tab:cross_platform_validation} highlight the sensitivity of force-aware tasks to end-effector design and force regulation.

\begin{table}[htb]
  \caption{Cross-platform data validation: task success rates across four LIBERO subtasks. We compare the original MuJoCo dataset, our replay in Isaac Lab with identical robot configuration (denoted Isaac), and our modified setup with a tactile-equipped Franka gripper at 100\%, 25\%, and 10\% force scaling (denoted T-100, T-25, and T-10).}
  \label{tab:cross_platform_validation}
  \centering
  \small
  \begin{tabular}{lccccc}
    \toprule
    Task & MuJoCo & Isaac & T-100 & T-25 & T-10 \\
    \midrule
    Spatial & 0.86 & 0.83 & 0.42 & 0.24 & 0.07 \\
    Object & 0.91 & 0.77 & 0.84 & 0.87 & 0.73 \\
    Goal & 0.76 & 0.78 & 0.55 & 0.44 & 0.30 \\
    LIBERO-10 & 0.86 & 0.66 & 0.60 & 0.47 & 0.32 \\
    Average & 0.85 & 0.76 & 0.60 & 0.50 & 0.36 \\
    \bottomrule
  \end{tabular}
\end{table}

\begin{figure}[t]
  \centering
  \begin{minipage}{0.63\textwidth}
    \centering
    \subcaptionbox{LIBERO Spatial}{\includegraphics[width=0.42\linewidth]{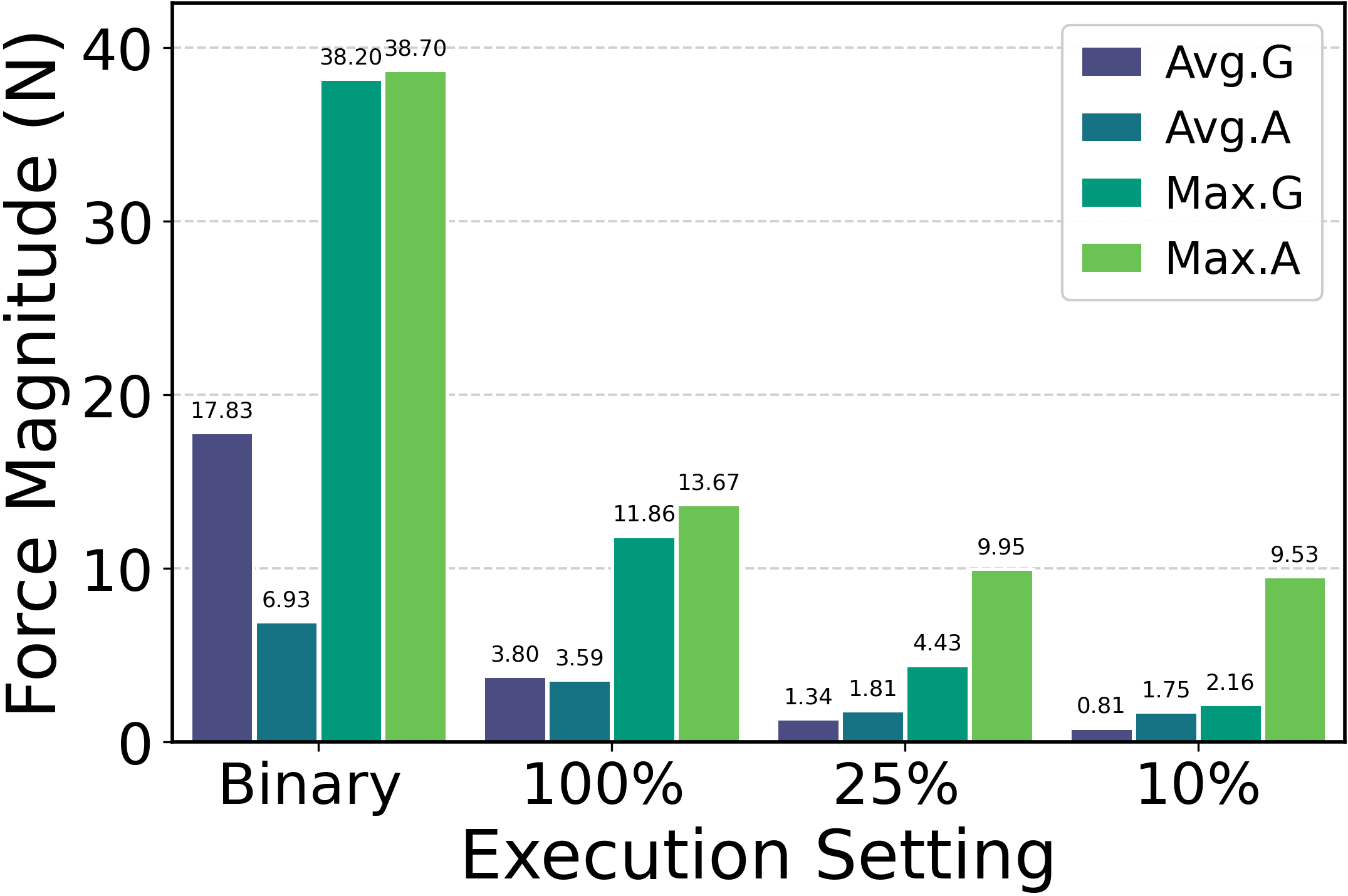}}\hfill
    \subcaptionbox{LIBERO Object}{\includegraphics[width=0.42\linewidth]{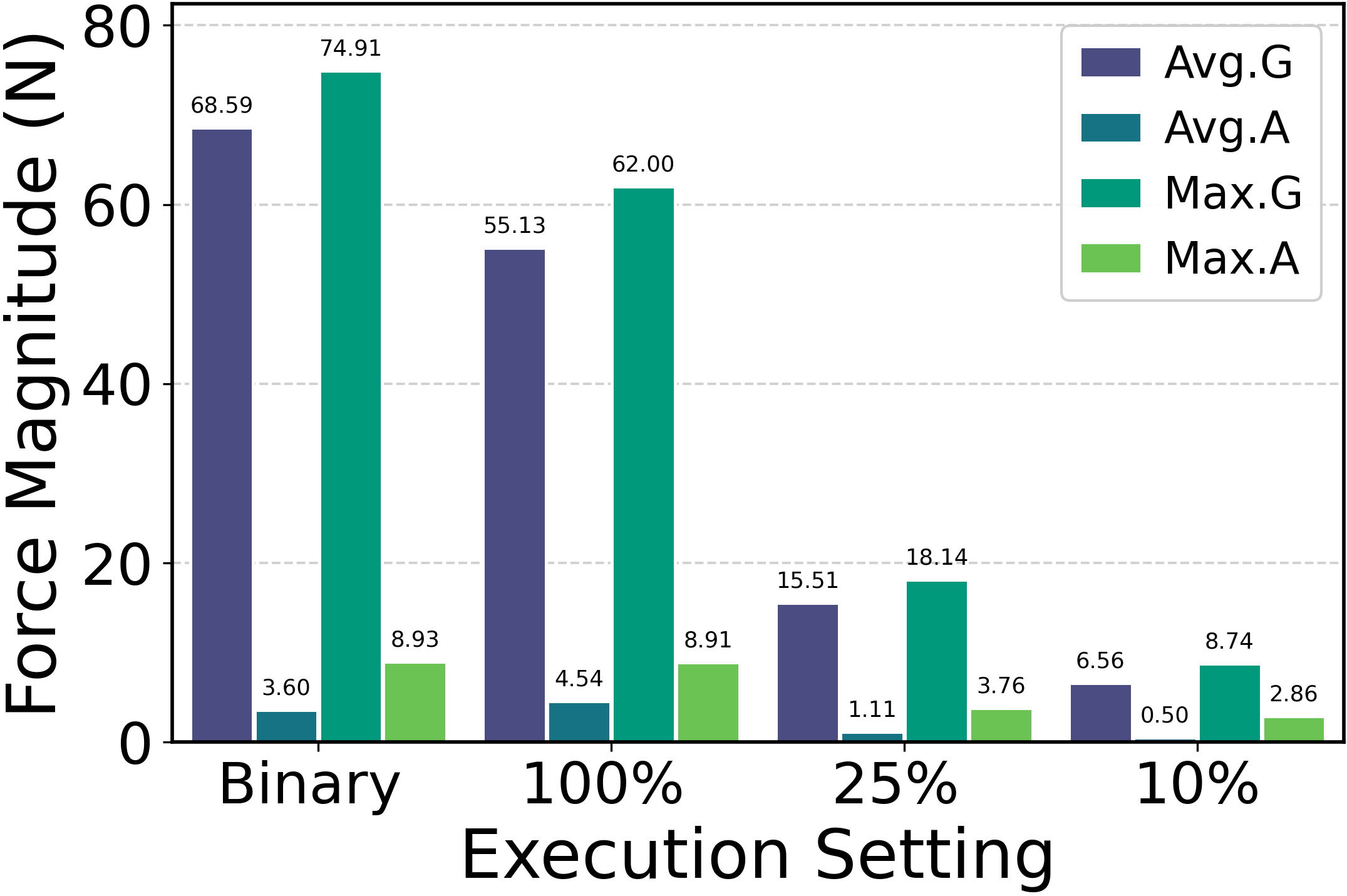}}\\[0.6em]
    \subcaptionbox{LIBERO Goal}{\includegraphics[width=0.42\linewidth]{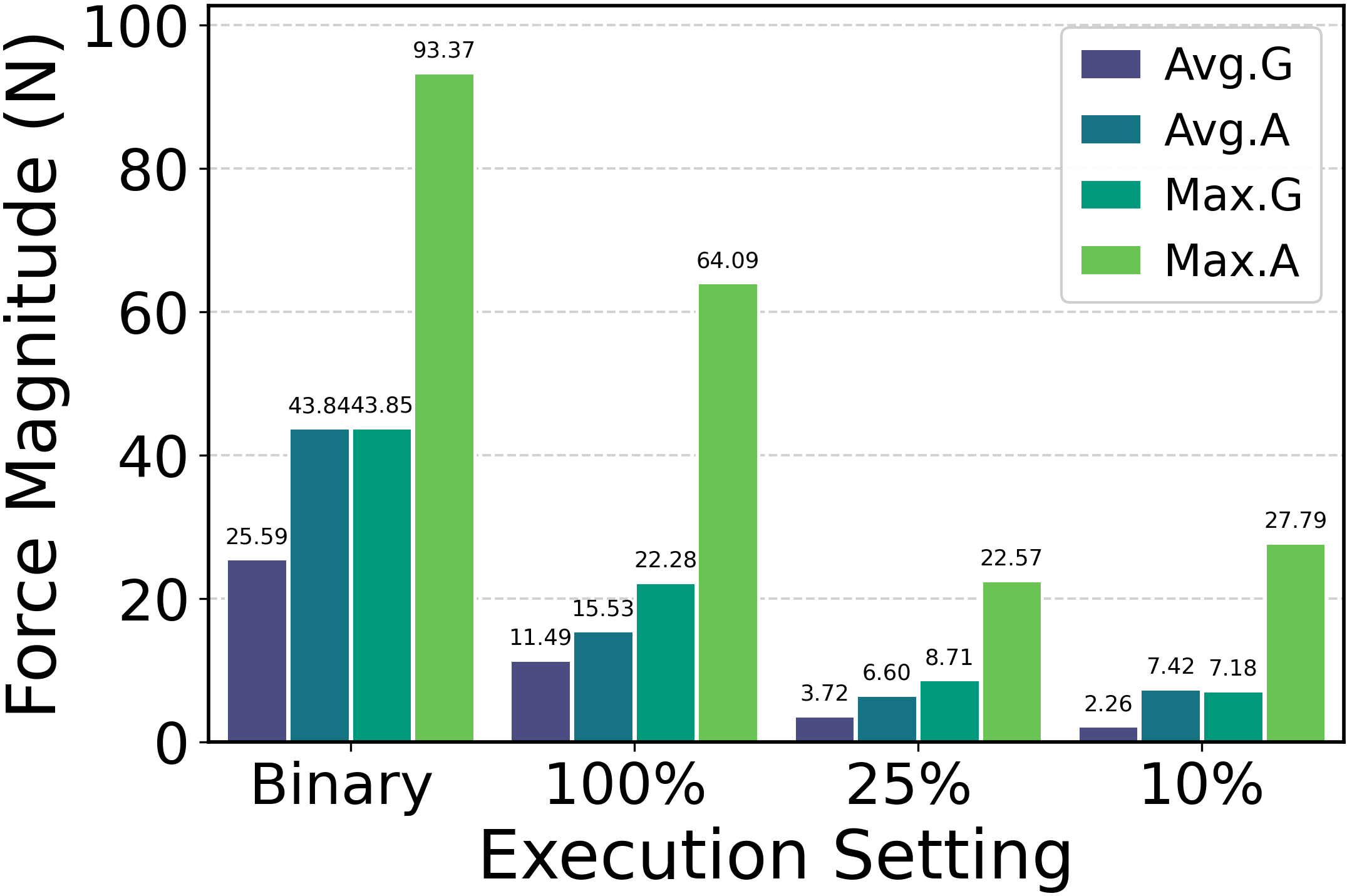}}\hfill
    \subcaptionbox{LIBERO-10}{\includegraphics[width=0.42\linewidth]{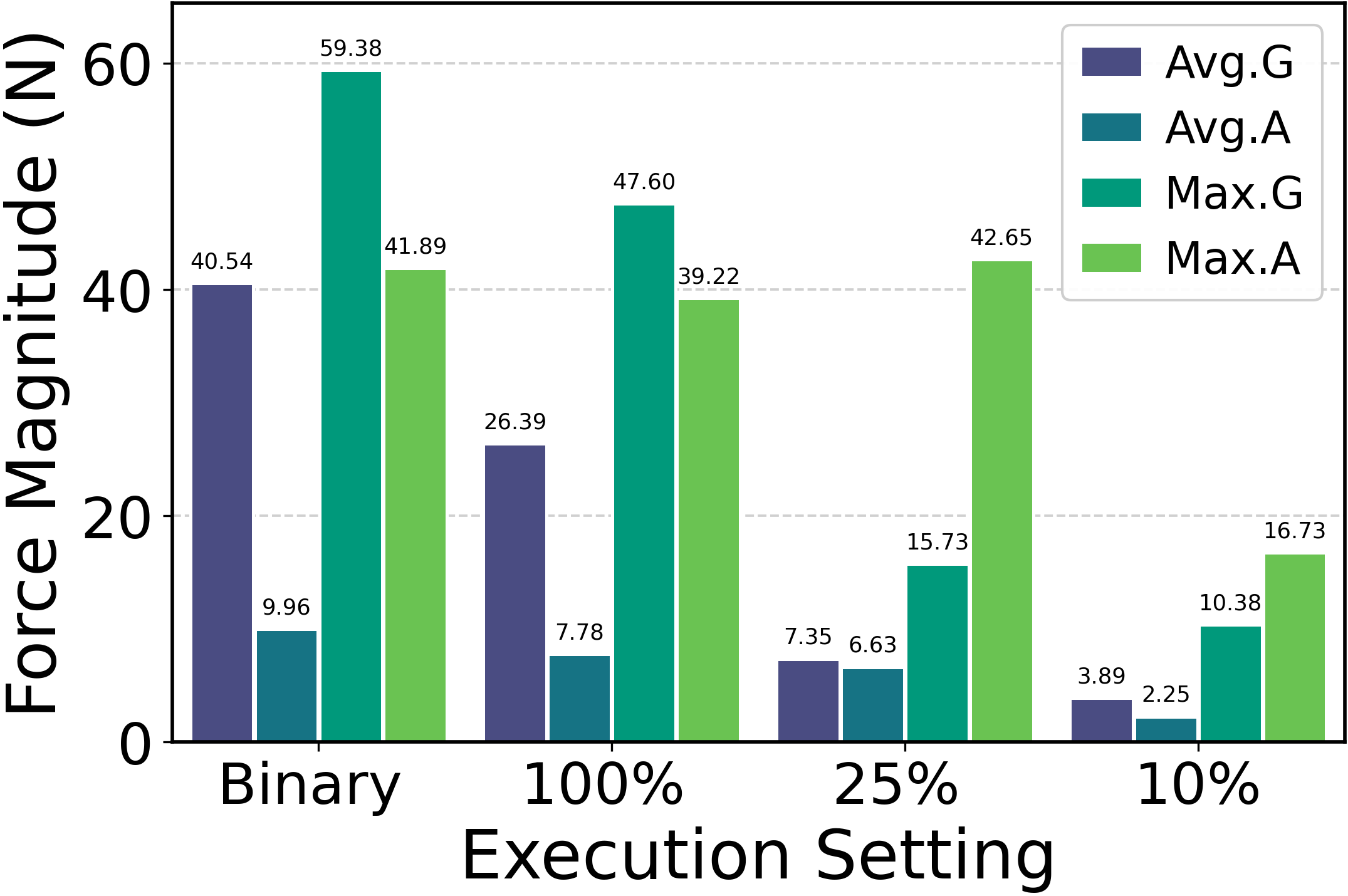}}
  \end{minipage}\hfill
  \begin{minipage}{0.32\textwidth}
    \centering
    \includegraphics[width=0.98\linewidth]{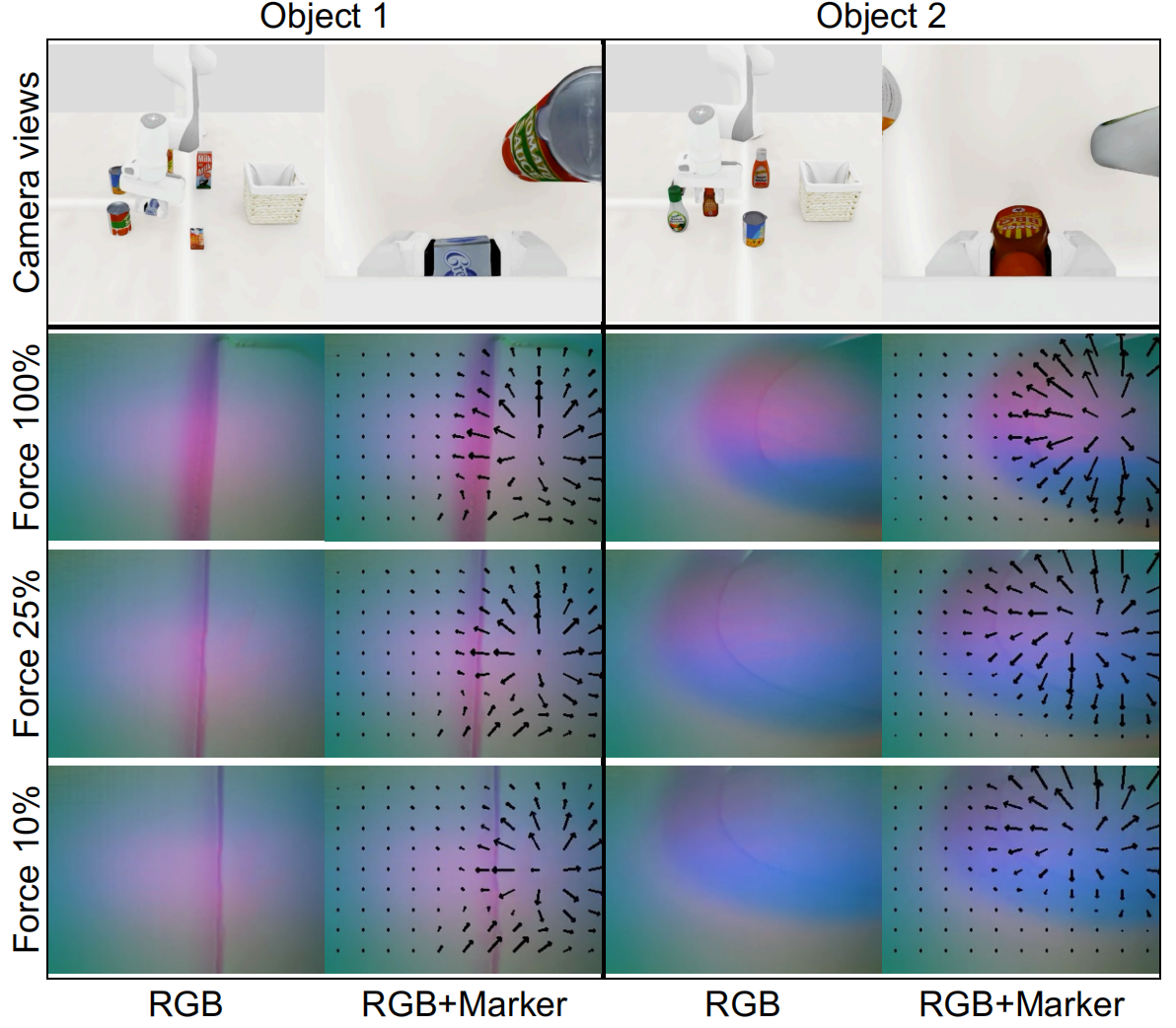}
  \end{minipage}
\caption{\textbf{Force distribution and tactile observations at different force magnitudes.} \textbf{Left:} force distributions under various control modes across different task suites. ``Binary'' represents the binarized control commands applied by a non-tactile gripper; ``100\%'', ``25\%'', and ``10\%'' indicate the force distributions of a tactile gripper under different force settings, where ``100\%'' retains the default gripper PD gains of the original continuous-control setup. The ``Binary'' setting records discrete open/close commands without regulating a force target, resulting in a higher peak applied force than the 100\% setting. \textbf{Right:} tactile images and corresponding camera images at different force magnitudes. The left two columns represent the first category of objects and the right two columns the second; RGB denotes the simulated tactile RGB image, and RGB+Marker overlays the simulated marker motion field on the RGB image. In the camera views, the left view is from the third-person camera and the right view is from the wrist-mounted camera.}
\label{fig:force_distribute}
\end{figure}

\subsection{From Discrete to Continuous Force-Position Control}
\label{sec:eval-task-design}

We verify whether our data collection framework effectively expands the distribution of interaction forces. We compare a baseline using binary gripper control, which logs only discrete open/close commands and regulates no force, against our approach, which explicitly sets different force parameters during execution; results are shown in Fig.~\ref{fig:force_distribute}. Specifically, we define 100\% grip force as the ``firm'' condition, modifying the original instruction with the adverbs ``firmly'' and ``tightly'', and 25\% and 10\% as the ``gentle'' conditions, annotated with ``gently'' and ``softly''. These conditions define an ordered force scale. The adverb set corresponds to increasing grip force as the commanded level increases, and unseen adverbs reserved for testing (Appendix~\ref{app:semantic-force}) produce intermediate values between the two endpoints. Language thus specifies a position along a continuous force scale. Here, ``Binary'' refers specifically to the baseline gripper's discrete open/close command interface. In our experiments, we set $K_p \in \{2000, 500, 200\}$ N/m and $K_d \in \{100, 25, 10\}$ N·s/m to correspond to 100\%, 25\%, and 10\% force levels, respectively, where the 100\% setting follows the default parameters for the Franka robot in Isaac Sim. For the same task, we present tactile images, marker fields, and contact force measurements at comparable contact stages under these distinct force settings in Fig.~\ref{fig:force_distribute}, demonstrating clear multimodal variations aligned with the intended interaction intensity.

\noindent We also observe that under the extreme 10\% force setting, the retention rate of most tasks is relatively low. Appendix~\ref{app:C} gives the formal definition of the retention rate. We therefore construct a Tabero subset to evaluate policy performance under these extreme conditions in the subsequent ablations. This subset includes 9 tasks from the Object dataset, each executed under two force conditions specified by linguistic adverbs. The composition of the dataset is presented in Appendix~\ref{app:force-language}. We collect three datasets over the task group, differing only in force magnitude and gripper actuation strategy.
\textbf{Dataset A} uses continuous control with ``gentle'' forces at 25\% of the corresponding ``firm'' level.
\textbf{Dataset B} also employs continuous control but reduces the force to 10\%, representing an extreme low-force regime where slippage is likely.
\textbf{Dataset C} uses the same 10\% force level but switches to binary open/close commands, with only discrete gripper states logged.

\noindent During evaluation, the arm and gripper operate with impedance parameters corresponding to 100\% force scaling. The Tacinf-VTLA policy predicts target contact forces conditioned on language and visual inputs; these targets are tracked by a closed-loop admittance controller that modulates impedance during execution. We adapt a base VLA model using low-rank adaptation (LoRA) to incorporate tactile marker fields (Dataset A and B), while a vision-language-only variant is trained on Dataset C for ablation, where the binary gripper commands carry no continuous force target to predict. As shown in Table~\ref{tab:dataset_compare}, that variant does not modulate force at all (AG 35.8~N under firm against 34.6~N under gentle), consistent with the ablation below: force modulation appears only when the policy is given a continuous force target to predict. The sharp drop in success rate from 25\% to 10\% force scaling reflects the increased difficulty of maintaining stable contact under ultra-gentle constraints, showing that Dataset B represents a substantially more challenging regime for gentleness-aware policies. Appendix~\ref{app:semantic-force} reports the corresponding language-to-force trend on this dataset, including the out-of-domain adverbs.

\begin{table}[htb]
  \caption{Training results on the three datasets. SR = success rate. Each cell gives the firm / gentle value, where the gentle condition is 25\% for Dataset A, 10\% for Dataset B, and 10\% with binary gripper commands for Dataset C.}
  \label{tab:dataset_compare}
  \centering
  \small
  \setlength{\tabcolsep}{4pt}
  \begin{tabular}{lccc}
    \toprule
    Metric & A 100\% / 25\% & B 100\% / 10\% & C 100\% / 10\% \\
    \midrule
    SR & 0.87 / 0.79 & 0.86 / 0.52 & 0.86 / 0.87 \\
    AG & 31.3 / \textbf{8.5} & 32.4 / \textbf{3.7} & 35.8 / 34.6 \\
    MG & 61.3 / 19.9 & 62.7 / 11.4 & 65.7 / 65.0 \\
    \bottomrule
  \end{tabular}
\end{table}

\subsection{Ablation and Comparison of VTLA}
\label{sec:eval-vtla-ablation}
We use \textbf{Dataset B} to compare tactile input methods and ablate force-control components. FS denotes force supervision. Force E+FS, Img, and Force D+FS adapt the design principles of~\citet{tactile-vla}, \citet{vtla}, and \citet{ta-vla}, respectively, within our shared backbone and training protocol. These experiments assess the design choices under a common backbone, dataset, and protocol. Because the implementations differ from the original systems in these respects, the results are not directly comparable with those reported in the source papers. Except where the ablation architecture requires a different tactile interface, all models were fine-tuned via LoRA~\cite{lora} with an identical set of hyperparameters; Appendix~\ref{App:hyper} summarizes the evaluated variants.

\begin{figure}[t]
  \centering
  \includegraphics[width=0.98\textwidth]{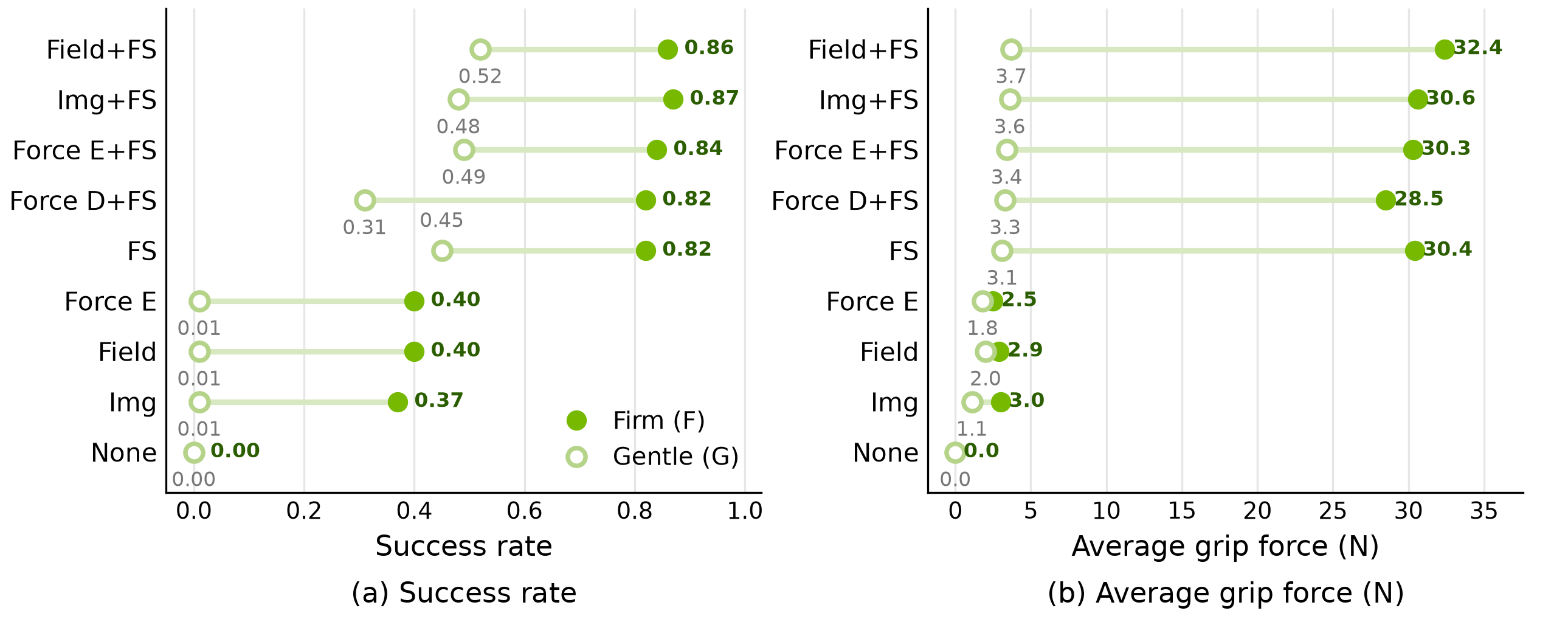}
  \caption{\textbf{Ablation study on tactile modalities.} (a) Task success rates under firm (F) and gentle (G) conditions; line length indicates the difference in success rate between conditions. (b) Average grip force (N) under the two conditions; line length indicates the difference in mean force. Filled and open circles denote firm and gentle conditions, respectively. None = no tactile input; Img = tactile image input; Field = marker displacement field input, i.e.\ the spatial marker field; Force E = six-dimensional force input via MLP encoder; Force D = six-dimensional force input via decoder; FS = force supervision loss enabled.}
  \label{fig:tactile_ablation}
\end{figure}

\noindent Fig.~\ref{fig:tactile_ablation} shows that the baseline without tactile input and without force supervision fails completely. The following ablations distinguish the contributions of these two factors.

\noindent\textbf{Role of force supervision.} Under the training and execution framework used for Dataset B, explicit supervision of fingertip-force targets and an interface for executing those targets account for the main performance gains. Without force supervision, Img, Field, and Force E each achieve a success rate of 0.01 under the gentle condition. FS, which receives no tactile observations at the high-level policy, achieves 0.45 under gentle and 0.82 under firm conditions. These results identify force-target learning and execution as important components of force-level following in this setting.

\noindent\textbf{Impact of tactile input.} The FS high-level policy receives no tactile images, marker fields, or fingertip-force observations. It nevertheless receives fingertip-force supervision and executes actions through a low-level controller that uses measured fingertip forces for feedback. Comparing FS with variants such as Field+FS therefore measures the additional value of high-level tactile observations under the same force supervision and feedback control. With force supervision enabled, the marker field provides the best balance of performance across firm and gentle conditions, and most other variants outperform FS under both conditions. The benefit varies across representations: Force D+FS performs worse than FS under the gentle condition, and the variation across models exceeds the average gain over FS. Tactile observations thus provide contact information that helps the policy infer the supervised force targets and use them to determine force and action outputs.

\subsection{From Single-Source to Multi-Source Data}
\label{sec:eval-taco-align}

\begin{figure}[t]
  \centering
  \includegraphics[width=\textwidth]{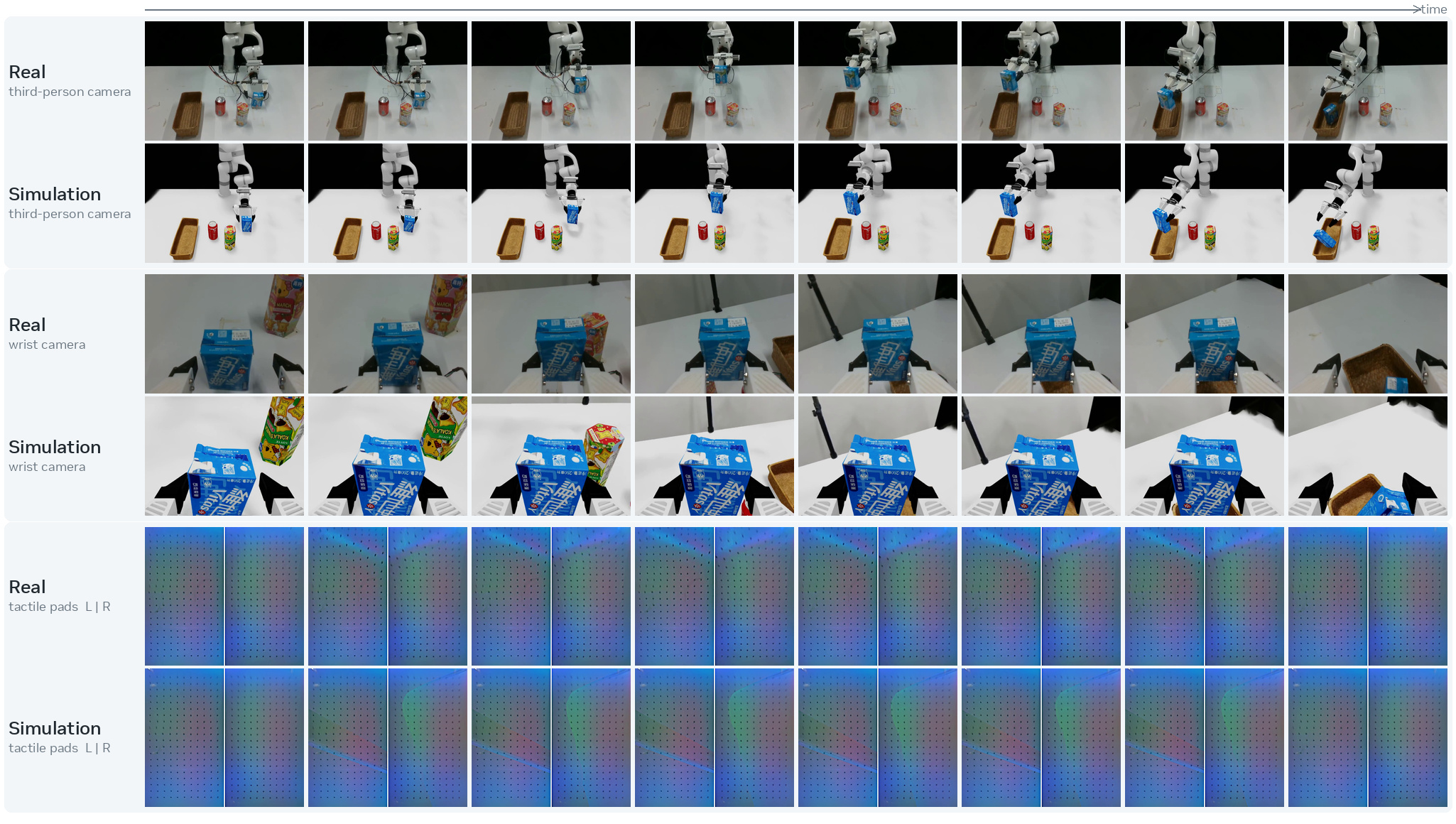}
  \caption{\textbf{Milk-carton grasping and placement in reality and simulation.} Eight time-aligned observations show the real and simulated third-person views, wrist-camera views, and left/right tactile pads as the robot grasps and places a milk carton.}
  \label{fig:sim2real_task_keyframes}
\end{figure}

We use the data-replay pipeline introduced by Tabero to obtain, from single-distribution real-robot data, simulated Taco data with ``firm'' and ``gentle'' force distributions, and evaluate the consistency between the two domains. To complement task-level validation, we measure the consistency along three channels: visual images, tactile images, and action labels. Visual consistency is quantified by the Contrastive Language-Image Pre-training (CLIP) similarity~\cite{radford2021clip} of the third-person and wrist camera views; tactile consistency by the structural similarity (SSIM) and the mean absolute error (MAE) between simulated and real tactile images; and action consistency by the MAE of the action labels. Table~\ref{tab:simreal_consistency} reports these metrics for three firm-force tasks. The results show that the wrist and third-person camera views are well aligned (both CLIP similarities above 0.75); the tactile signals further agree in SSIM and MAE, and the actions are also well aligned, demonstrating the data reliability of the Taco replay pipeline. Fig.~\ref{fig:sim2real_task_keyframes} shows this alignment qualitatively for one task, pairing the real run with the simulated replay across the third-person, wrist, and tactile views. UMI data further extend the data sources: because UMI data cost less than teleoperation, we can collect data over a larger spatial randomization range. Appendix~\ref{app:umi-validation} shows how the expanded spatial range differs from the original teleoperated data in trajectories and distributions, and Section~\ref{sec:eval-cotrain} further evaluates the synergy of multi-source data.

\begin{table}[htb]
  \caption{Sim-to-real data consistency across visual, tactile, and action channels. All three channels show agreement on every task: CLIP similarity exceeds 0.75, tactile SSIM exceeds 0.93, and action MAE is below 0.074. These results indicate that the replay pipeline closely reproduces the real data distributions on these tasks.}
  \label{tab:simreal_consistency}
  \centering
  \small
  \begin{tabular}{lccccc}
    \toprule
    & Third-person CLIP & Wrist CLIP & Tactile SSIM & Tactile MAE & Action MAE \\
    \midrule
    Firm Task 1 & 0.781 & 0.770 & 0.941 & 0.0160 & 0.0708 \\
    Firm Task 2 & 0.785 & 0.754 & 0.931 & 0.0182 & 0.0685 \\
    Firm Task 3 & 0.794 & 0.766 & 0.932 & 0.0196 & 0.0733 \\
    \bottomrule
  \end{tabular}
\end{table}

\subsection{Policy-Level Zero-Shot Sim2Real Transfer}
\label{sec:eval-sim2real}

We test whether Taco supports transfer of language-conditioned force levels to the physical robot when task-specific policy training uses no real-robot samples. Each policy is initialized from a pretrained VLA and adapted using simulation data only. After Real2Sim calibration, the same adapted policy, language conditions, and controller are used in the aligned simulation (Sim) and on the physical robot (Sim2Real). The force regimes are changed during simulation data generation, while the robot evaluation measures whether the corresponding force level is reproduced. Real tactile observations used for simulator and observation-mapping calibration are separate from policy optimization. Simulation evaluation uses random seeds different from those used for training; objects are placed manually for real-robot evaluation. Following the paired evaluation protocol of PolaRiS~\cite{jain2025polarisscalablerealtosimevaluations}, the two evaluations share the same task set, language instructions, controller settings, and termination criteria. For every policy we report the mean success rate (Mean SR), averaged over the firm (F) and gentle (G) conditions and over the three tasks, and the average grip force the policy predicts. The reference is the human teleoperation label: for task $t$ and force level $x\in\{\mathrm{F},\mathrm{G}\}$, the teleoperation demonstrations define $\mathrm{AG}^{\mathrm{teleop}}_{x,t}$, the average grip force the operator applies under the same instruction. We define the signed prediction deviation
\begin{equation}
\Delta_{x,t} = \mathrm{AG}^{\mathrm{pre}}_{x,t} - \mathrm{AG}^{\mathrm{teleop}}_{x,t},
\label{eq:pred_delta}
\end{equation}
so that $\Delta>0$ means the policy predicts a firmer grasp than the human label and $\Delta<0$ a gentler one. All deviations in this section use the predicted force \emph{pre} to assess agreement with the teleoperation reference. Measured forces (\emph{meas}) and controller tracking are evaluated separately in Section~\ref{sec:eval-consistency}. Each model contributes six deviations defined by Eq.~\eqref{eq:pred_delta} per domain, covering three tasks and two force levels. Table~\ref{tab:taco_sim2real} reports their mean absolute deviation:
\[
\bar{\Delta}=\frac{1}{6}\sum_{x\in\{\mathrm{F},\mathrm{G}\}}\sum_t |\Delta_{x,t}|.
\]

\noindent The two variants without force supervision, None and Img (Img uses a tactile image but has no force supervision), did not solve the evaluated tasks in either domain (Mean SR 0), and neither provided a force reading. Among the tactile policies with force supervision, Force D+FS attains the highest Mean SR, followed by Field+FS ($\pi_{0.5}$, no state input, NS) and Force E+FS; Field+FS ($\pi_0$) is lowest because it fails almost all Vitasoy episodes.

\noindent In simulation, every tactile model has a mean absolute deviation within 1.4~N, indicating predictions close to the teleoperation labels.

\begin{table}[htb]
  \caption{Predicted grip force against the human teleoperation label. Mean SR (\%) averages the firm (F) and gentle (G) success rates over the three tasks. $\bar{\Delta}$ is the mean absolute deviation $|\mathrm{AG}^{\mathrm{pre}}-\mathrm{AG}^{\mathrm{teleop}}|$ over the three tasks and both force levels, in N; larger $\bar{\Delta}$ means the predicted force departs further from the human label. Sim evaluates the adapted policy in the aligned simulation; Sim2Real evaluates it on the physical robot. A dash marks a policy with no force reading; None and Img fail both conditions in both domains. For Field+FS ($\pi_0$) on the robot, the Vitasoy gentle condition has no completed episode, so its $\bar{\Delta}$ averages the five remaining cells. The table uses AG as its force metric and omits MG.}
  \label{tab:taco_sim2real}
  \centering
  \small
  \setlength{\tabcolsep}{6pt}
  \begin{tabular}{lcccc}
    \toprule
    & \multicolumn{2}{c}{Sim2Real} & \multicolumn{2}{c}{Sim} \\
    \cmidrule(lr){2-3}\cmidrule(lr){4-5}
    Model & Mean SR (\%) $\uparrow$ & $\bar{\Delta}$ (N) $\downarrow$ & Mean SR (\%) $\uparrow$ & $\bar{\Delta}$ (N) $\downarrow$ \\
    \midrule
    None ($\pi_{0.5}$) & 0 & -- & 0 & -- \\
    Img ($\pi_0$, $w{=}0$) & 0 & -- & 0 & -- \\
    Force D+FS ($\pi_0$) & 98.3 & 3.63 & 99.0 & 0.62 \\
    Force E+FS ($\pi_0$) & 92.5 & 1.35 & 98.0 & 0.94 \\
    Field+FS ($\pi_0$) & 67.5 & 0.90 & 69.3 & 0.47 \\
    Field+FS ($\pi_{0.5}$, state) & 82.5 & 4.12 & 89.0 & 0.78 \\
    Field+FS ($\pi_{0.5}$, no state) & 96.7 & 1.14 & 96.7 & 1.38 \\
    Img+FS ($\pi_{0.5}$, no state) & 88.3 & 1.12 & 95.7 & 1.11 \\
    \bottomrule
  \end{tabular}

  \vskip 0.08in
  \begin{footnotesize}
    Table Notes: $\bar{\Delta}$ is computed from predicted forces (\emph{pre}). Averaging absolute deviations prevents positive and negative errors from canceling across force conditions. Both simulation and real-robot AG use the sum of the left and right fingertip-force magnitudes described in Section~\ref{sec:evaluation}. The remaining Img+FS ($\pi_{0.5}$) variants are evaluated in simulation only and reported under the force-separation metric in Appendix~\ref{app:imgfs_ablation}.
  \end{footnotesize}
\end{table}

\begin{table}[htb]
  \caption{Measured grip force in simulation and on the robot for the four policies with the highest task success rates. S2R = Sim2Real; NS = no state input. F SR/G SR = firm/gentle success rate (\%); F AG/G AG = measured average grip force (N); Gap $=$ Sim $-$ Sim2Real, so a positive value means the simulation measures a firmer grasp than the robot. The Mean row averages each column over the three tasks. Both simulation and real-robot AG are computed from the sum of the left and right fingertip-force magnitudes. Real-robot values are obtained from the Xense software development kit (SDK) and are subject to sensor estimation error.}
  \label{tab:meas_gap}
  \centering
  \small
  \setlength{\tabcolsep}{4pt}
  \begin{adjustbox}{max width=\textwidth}
  \begin{tabular}{ll cccc cccc cc}
    \toprule
    & & \multicolumn{4}{c}{Sim} & \multicolumn{4}{c}{Sim2Real} & \multicolumn{2}{c}{Gap (Sim$-$S2R)} \\
    \cmidrule(lr){3-6}\cmidrule(lr){7-10}\cmidrule(lr){11-12}
    Model & Task & F SR & G SR & F AG & G AG & F SR & G SR & F AG & G AG & F AG & G AG \\
    \midrule
    \multirow{4}{*}{Force D+FS ($\pi_0$)} & Vitasoy & 98.0 & 96.0 & 25.29 & 9.00 & 100.0 & 100.0 & 19.66 & 13.23 & +5.63 & -4.23 \\
    & Coca-Cola & 100.0 & 100.0 & 7.73 & 4.54 & 100.0 & 100.0 & 7.95 & 6.42 & -0.22 & -1.88 \\
    & Cookie & 100.0 & 100.0 & 24.53 & 7.46 & 90.0 & 100.0 & 6.87 & 6.27 & +17.66 & +1.19 \\
    & Mean & 99.3 & 98.7 & 19.18 & 7.00 & 96.7 & 100.0 & 11.49 & 8.64 & +7.69 & -1.64 \\
    \midrule
    \multirow{4}{*}{Force E+FS ($\pi_0$)} & Vitasoy & 98.0 & 94.0 & 26.42 & 10.27 & 100.0 & 90.0 & 10.81 & 5.72 & +15.61 & +4.55 \\
    & Coca-Cola & 96.0 & 100.0 & 7.80 & 4.70 & 100.0 & 100.0 & 10.03 & 8.25 & -2.23 & -3.55 \\
    & Cookie & 100.0 & 100.0 & 25.13 & 8.22 & 70.0 & 95.0 & 7.55 & 3.08 & +17.58 & +5.14 \\
    & Mean & 98.0 & 98.0 & 19.78 & 7.73 & 90.0 & 95.0 & 9.46 & 5.68 & +10.32 & +2.05 \\
    \midrule
    \multirow{4}{*}{Field+FS ($\pi_{0.5}$, NS)} & Vitasoy & 98.0 & 94.0 & 27.23 & 10.05 & 85.0 & 95.0 & 20.14 & 10.29 & +7.09 & -0.24 \\
    & Coca-Cola & 100.0 & 100.0 & 7.80 & 4.74 & 100.0 & 100.0 & 8.12 & 7.47 & -0.32 & -2.73 \\
    & Cookie & 94.0 & 94.0 & 25.58 & 8.43 & 100.0 & 100.0 & 18.56 & 10.37 & +7.02 & -1.94 \\
    & Mean & 97.3 & 96.0 & 20.20 & 7.74 & 95.0 & 98.3 & 15.60 & 9.38 & \textbf{+4.60} & \textbf{-1.64} \\
    \midrule
    \multirow{4}{*}{Img+FS ($\pi_{0.5}$, NS)} & Vitasoy & 88.0 & 88.0 & 26.47 & 10.69 & 90.0 & 80.0 & 13.05 & 10.60 & +13.42 & +0.09 \\
    & Coca-Cola & 100.0 & 100.0 & 7.78 & 5.10 & 100.0 & 100.0 & 10.46 & 7.01 & -2.68 & -1.91 \\
    & Cookie & 100.0 & 98.0 & 25.36 & 8.48 & 80.0 & 80.0 & 18.90 & 10.59 & +6.46 & -2.11 \\
    & Mean & 96.0 & 95.3 & 19.87 & 8.09 & 90.0 & 86.7 & 14.14 & 9.40 & \textbf{+5.73} & \textbf{-1.31} \\
    \bottomrule
  \end{tabular}
  \end{adjustbox}
\end{table}

\subsection{Crossing the Sim2Real Gap}
\label{sec:eval-consistency}

\begin{figure}[!htbp]
  \centering
  \includegraphics[width=0.92\textwidth]{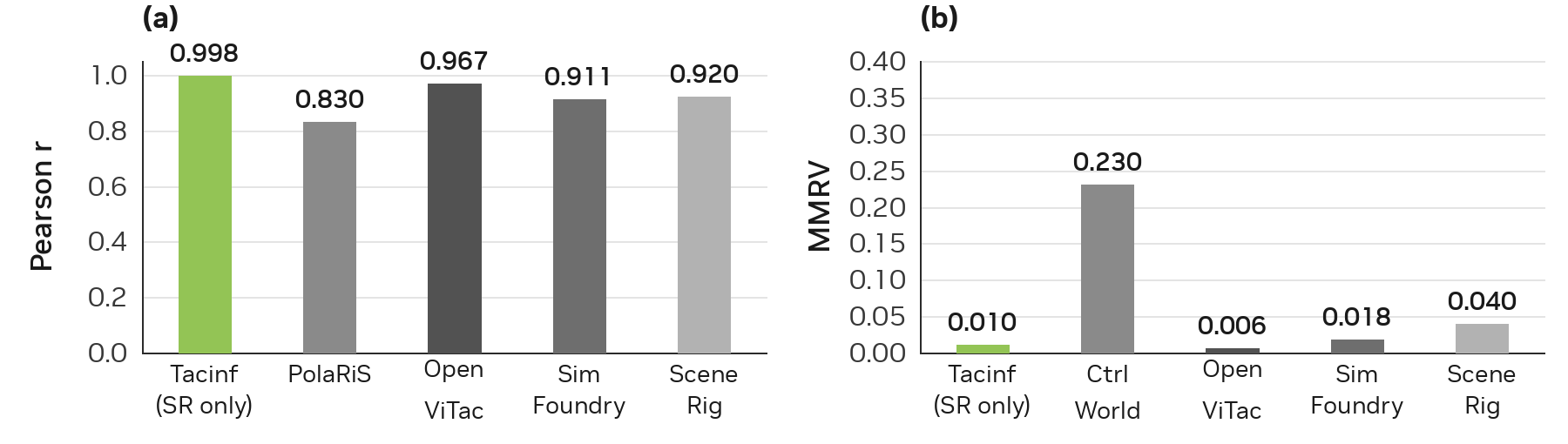}
  \caption{\textbf{Cross-domain agreement of simulated evaluations.} (a) Pearson correlation between simulated and real-robot scores, higher is better; (b) Mean Maximum Rank Violation (MMRV), lower is better. Across eight policy variants, the success-only score gives $r=0.998$ and MMRV$=0.010$. Literature bars: PolaRiS $r=0.83$~\citep{jain2025polarisscalablerealtosimevaluations}, Ctrl-World MMRV$=0.23$~\citep{jain2025polarisscalablerealtosimevaluations}, OpenViTac $r=0.967$/MMRV$=0.006$~\citep{openvitac2026}, SimFoundry $r=0.911$/MMRV$=0.018$~\citep{simfoundry}, and SceneRig $r=0.92$/MMRV$=0.04$~\citep{scenerig2026}. Each literature value is reported on the corresponding benchmark and provides context only. Differences in task suites, policy pools, and scoring protocols preclude direct comparison.}
  \label{fig:agreement_bars}
\end{figure}

Following SIMPLER and PolaRiS~\cite{simpler,jain2025polarisscalablerealtosimevaluations}, we quantify the faithfulness of the simulated evaluation with the Pearson correlation and the Mean Maximum Rank Violation (MMRV) between the Sim2Real and Sim scores, computed across the eight policy variants evaluated in both domains: the seven tactile variants with paired Sim and Sim2Real scores, together with the $\pi_{0.5}$ baseline under the position-controlled gripper system. We use the success-only score, Mean SR. Fig.~\ref{fig:agreement_bars} reports $r=0.998$ and MMRV$=0.010$: the success ordering transfers almost perfectly. Excluding the two baselines with zero success yields $r=0.961$, which still indicates a strong correlation.

\noindent Fig.~\ref{fig:agreement_bars} also places this agreement alongside recent sim-to-real evaluations. PolaRiS~\citep{jain2025polarisscalablerealtosimevaluations} reports $r=0.83$, and its published comparison reports an MMRV of $0.23$ for Ctrl-World~\citep{jain2025polarisscalablerealtosimevaluations}. OpenViTac~\citep{openvitac2026} reports $r=0.967$ and MMRV$=0.006$ across seven methods and eight shared tasks; SimFoundry~\citep{simfoundry} reports $r=0.911$ and MMRV$=0.018$ across seven tasks and five policy families; SceneRig~\citep{scenerig2026} reports $r=0.92$ and MMRV$=0.04$ across ten policy-task pairs. These values provide context; differences in task suites, policy pools, and scoring protocols preclude direct comparison.

\noindent Table~\ref{tab:meas_gap} reports measured AG in both domains for the four policies with the highest success rates, together with the per-task gap $\mathrm{AG}^{\mathrm{meas}}_{\mathrm{Sim}}-\mathrm{AG}^{\mathrm{meas}}_{\mathrm{S2R}}$; a positive gap means that the simulation measures a firmer grasp than the robot. Although the field- and image-based $\pi_{0.5}$ NS policies trail the force-based models in success, their transfer results are smaller in the reference metrics. The measured values also differ from the predicted values; we attribute this difference to the gripper controller parameters.

\subsection{Execution of Force-Position Control}
\label{sec:eval-controller}

\begin{figure}[!htbp]
  \centering
  \begin{subfigure}[t]{0.23\textwidth}
    \centering
    \includegraphics[width=\linewidth]{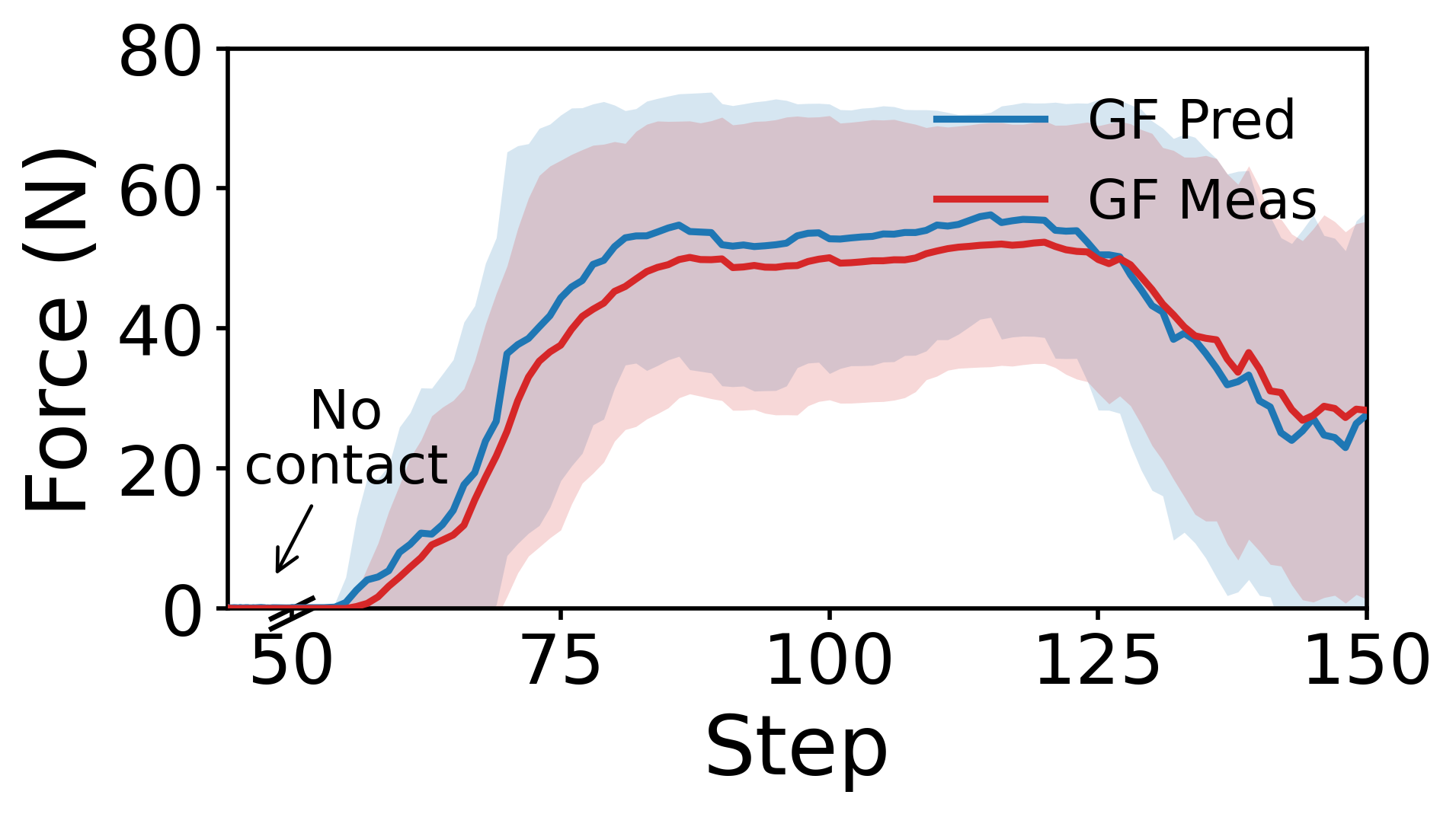}
    \caption{100\% force}
  \end{subfigure}\hfill
  \begin{subfigure}[t]{0.23\textwidth}
    \centering
    \includegraphics[width=\linewidth]{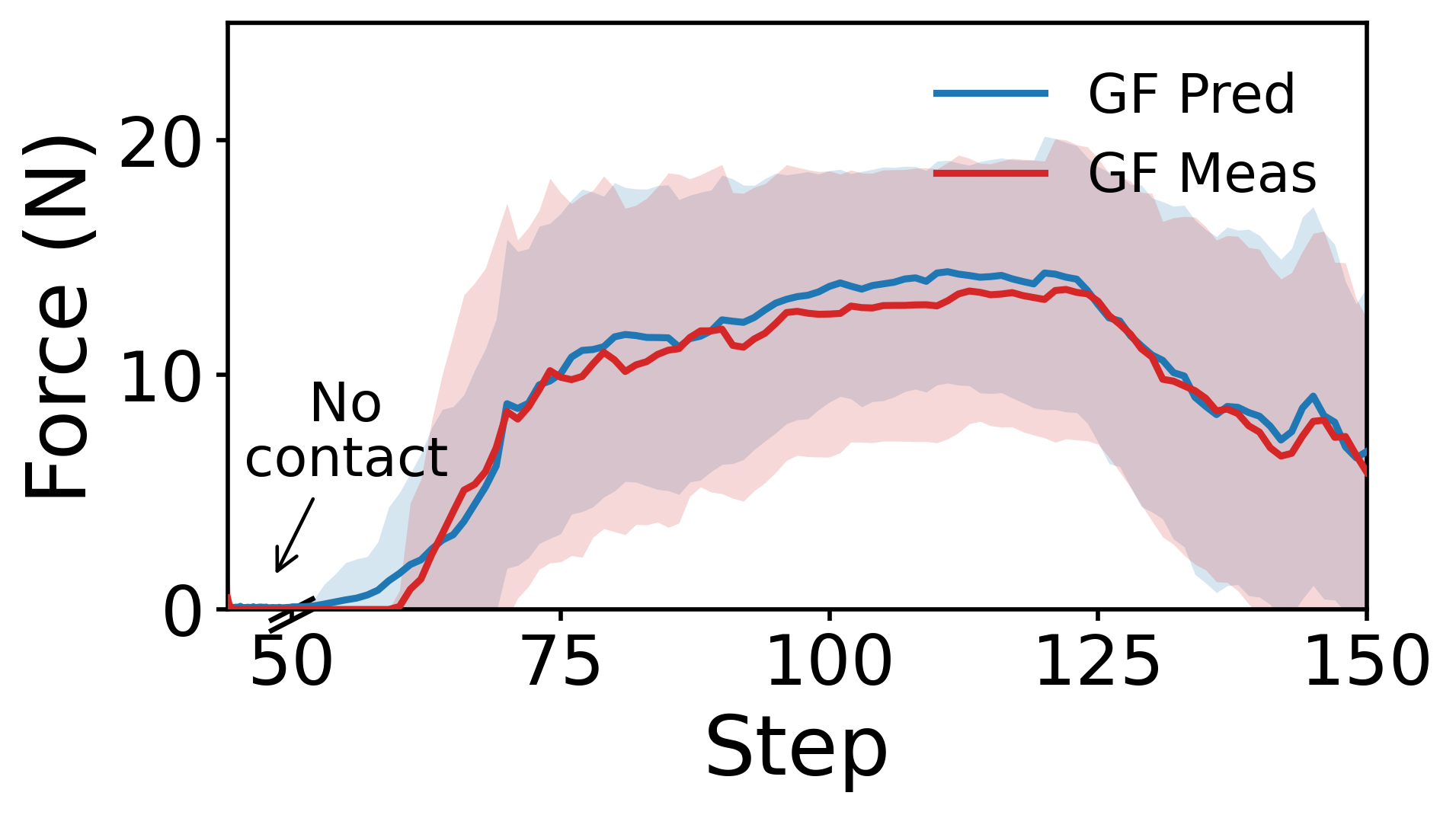}
    \caption{25\% force}
  \end{subfigure}\hfill
  \begin{subfigure}[t]{0.23\textwidth}
    \centering
    \includegraphics[width=\linewidth]{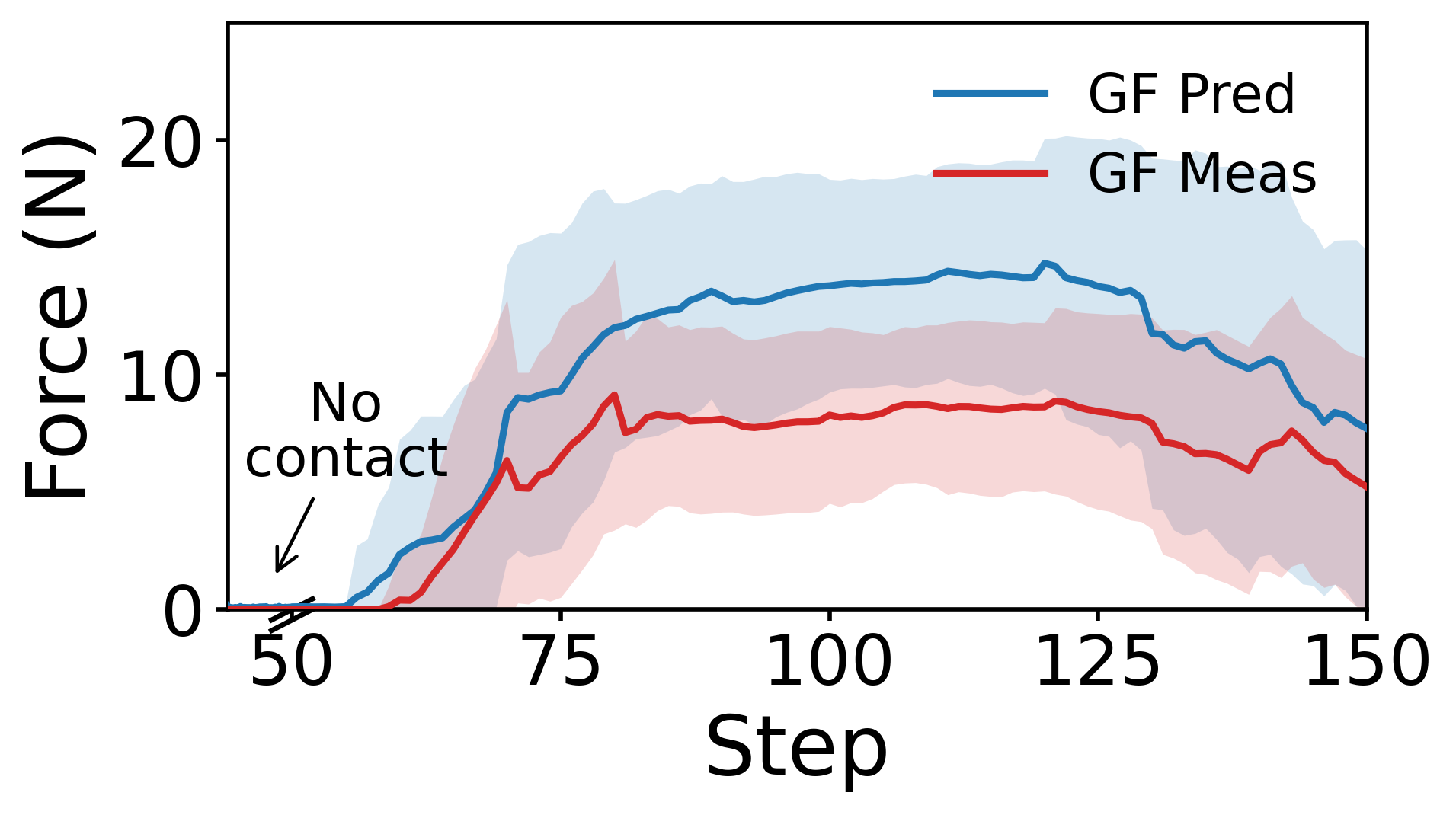}
    \caption{25\%, no feedforward}
  \end{subfigure}\hfill
  \begin{subfigure}[t]{0.23\textwidth}
    \centering
    \includegraphics[width=\linewidth]{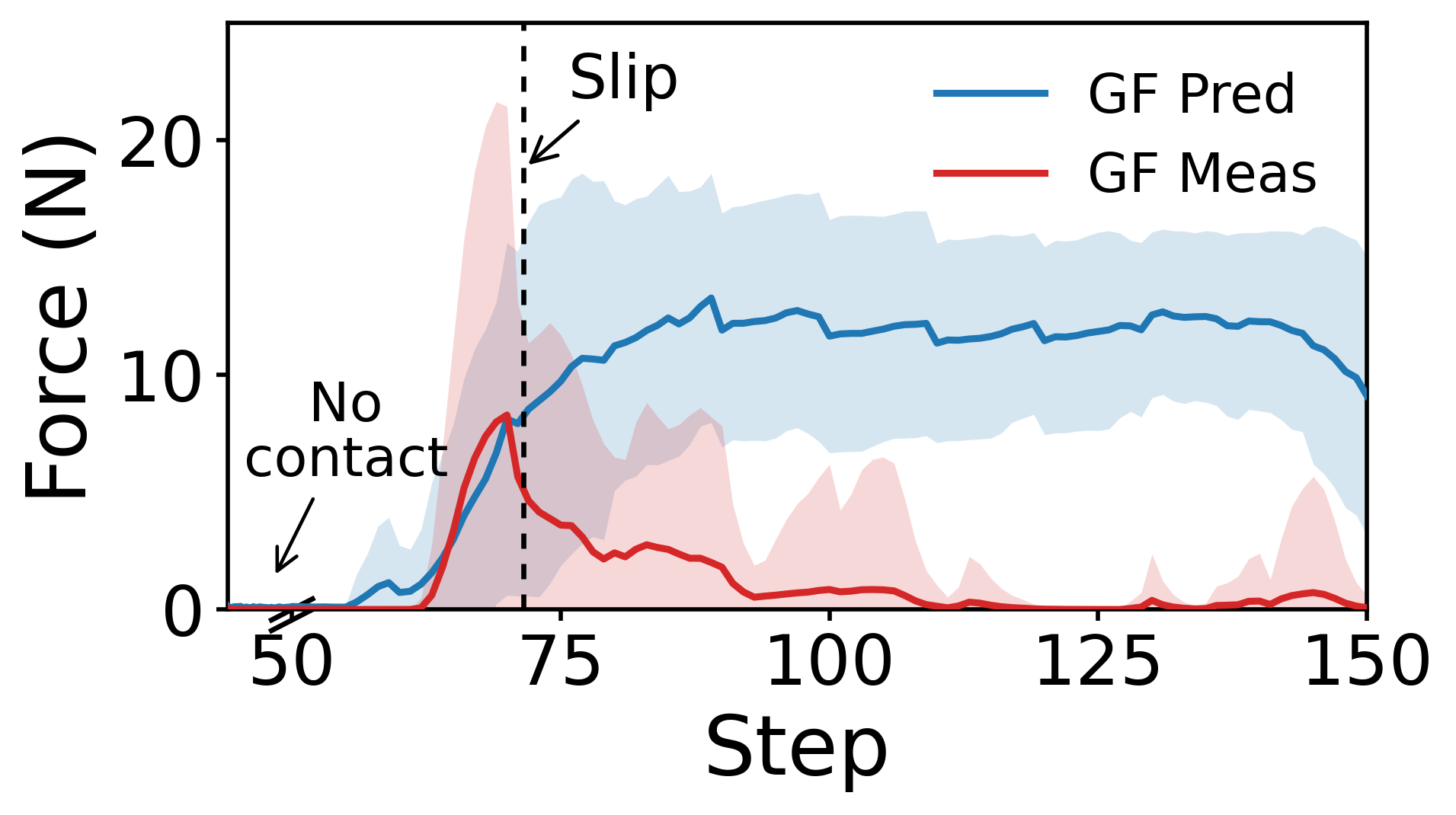}
    \caption{25\%, no admittance}
  \end{subfigure}\\[0.5em]
  \begin{subfigure}[t]{0.23\textwidth}
    \centering
    \includegraphics[width=\linewidth]{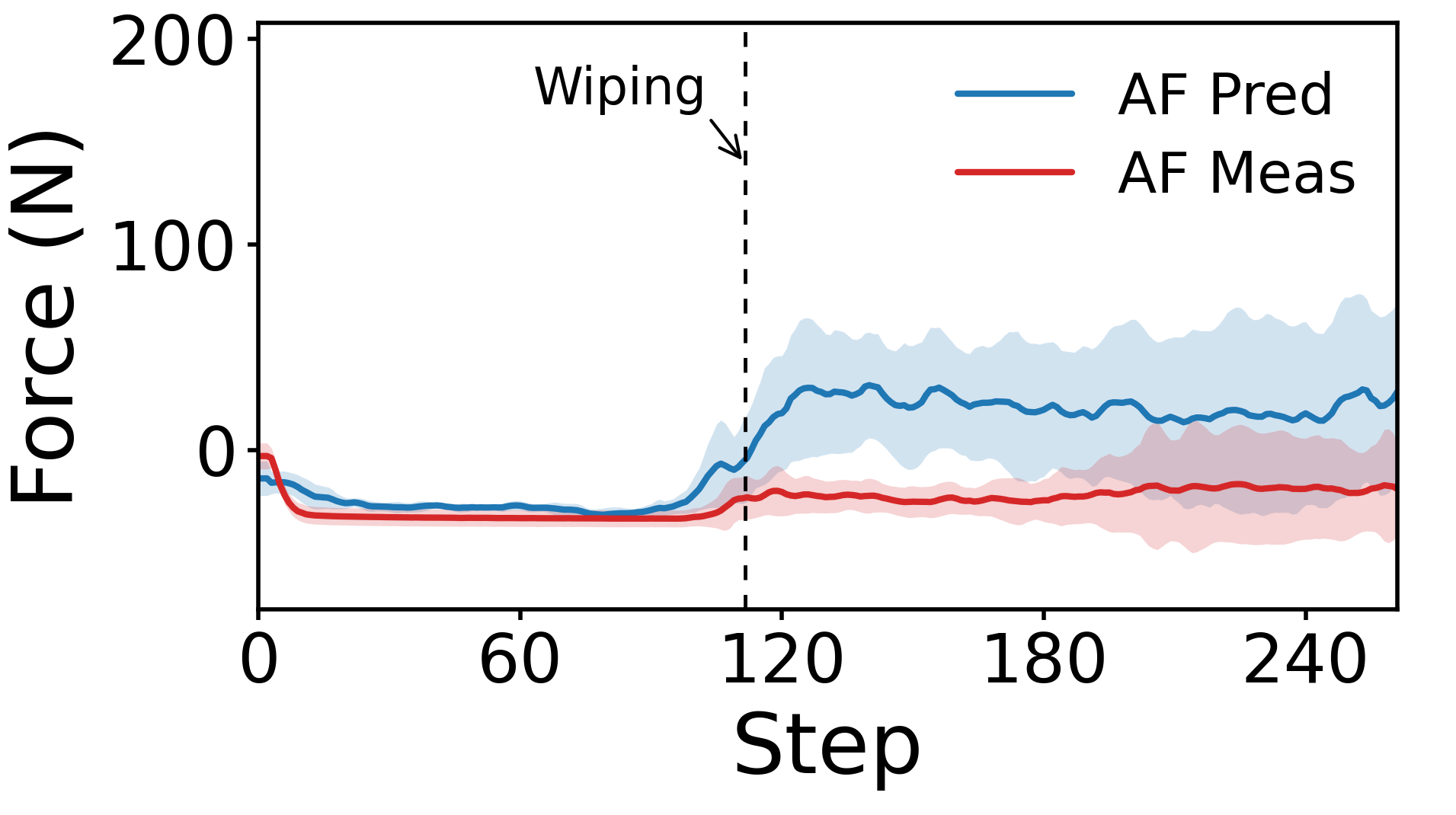}
    \caption{Blackboard $F_z$, off}
  \end{subfigure}\hfill
  \begin{subfigure}[t]{0.23\textwidth}
    \centering
    \includegraphics[width=\linewidth]{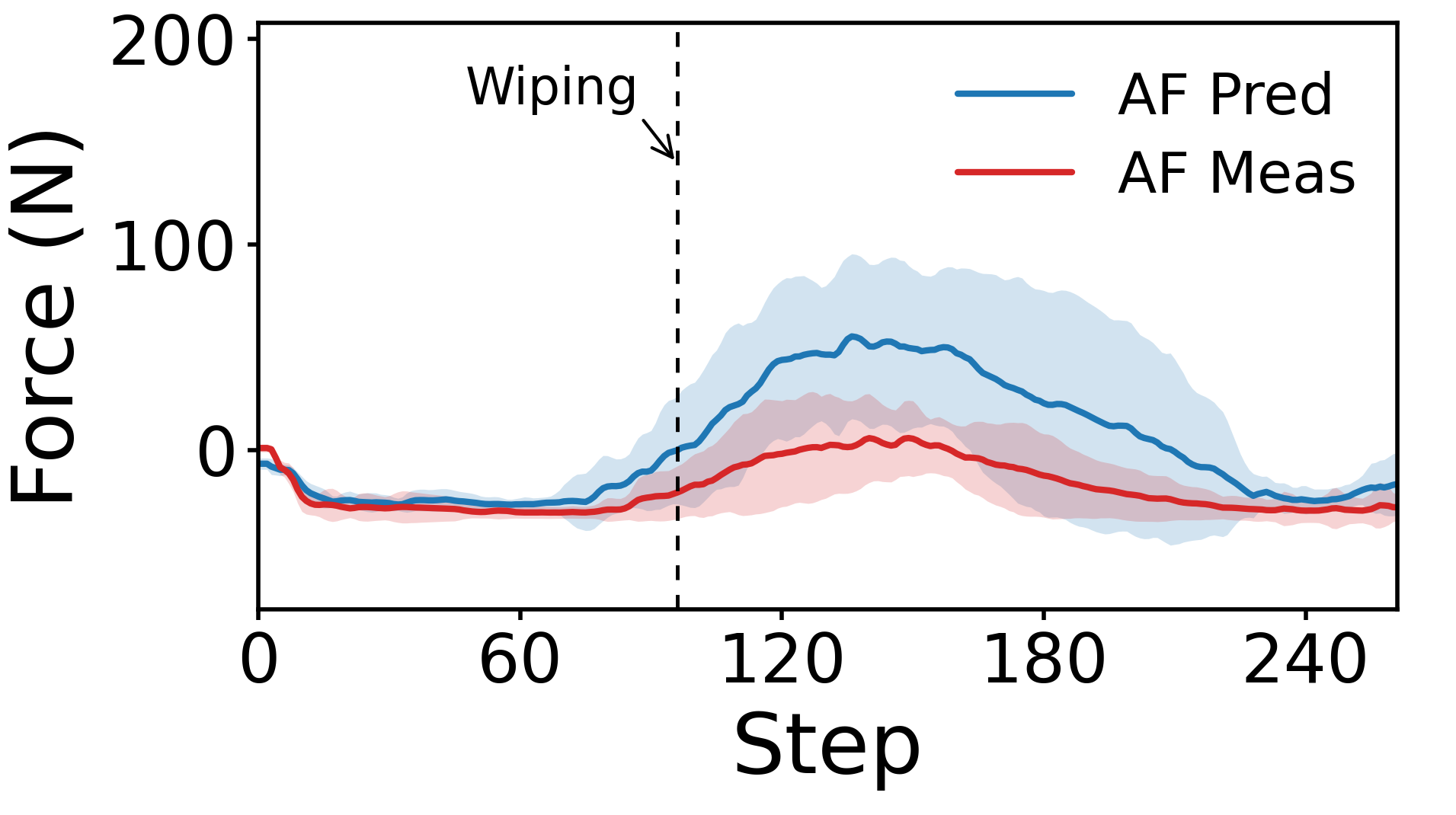}
    \caption{Blackboard $F_z$, on}
  \end{subfigure}\hfill
  \begin{subfigure}[t]{0.23\textwidth}
    \centering
    \includegraphics[width=\linewidth]{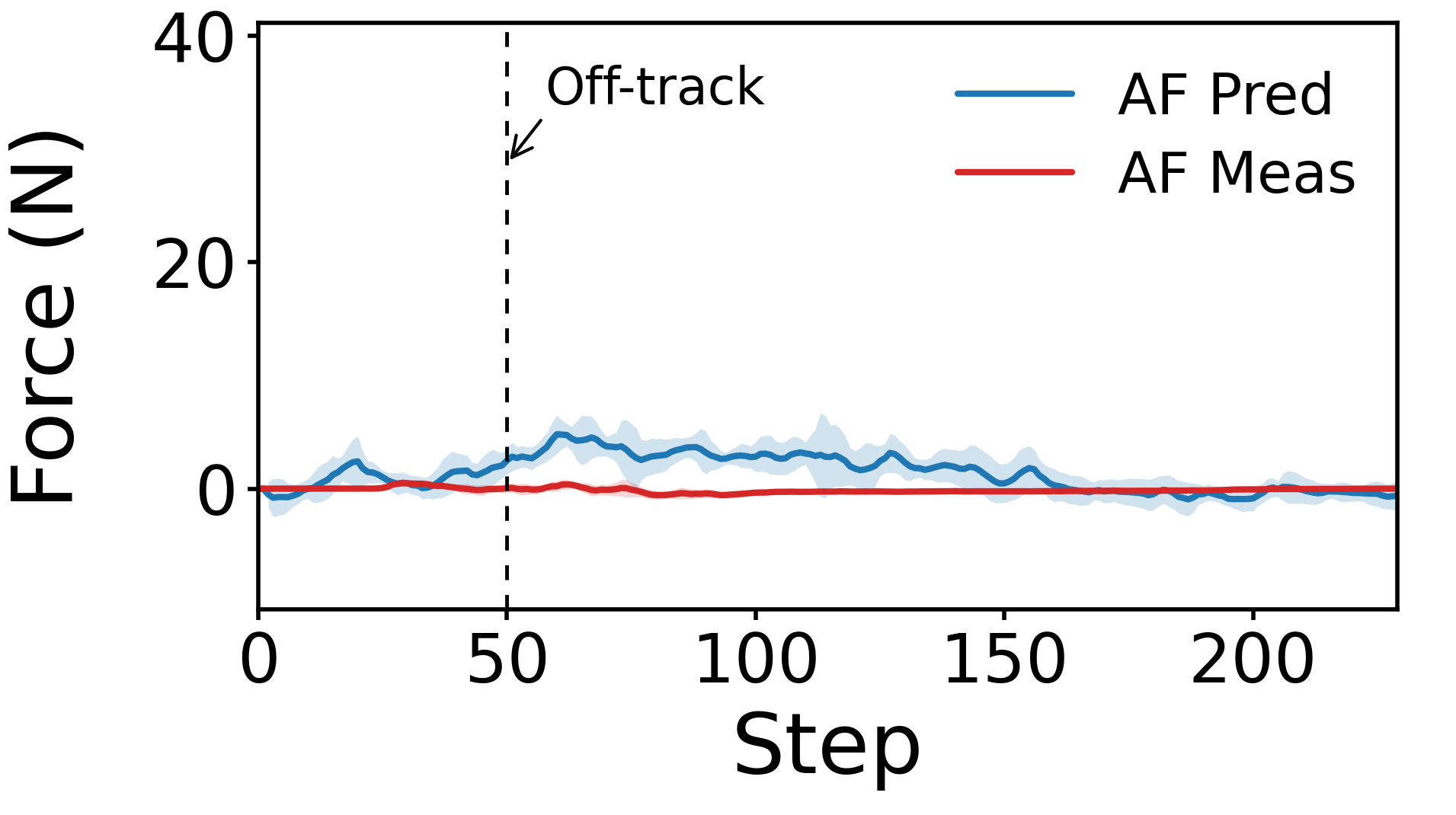}
    \caption{Card $F_x$, off}
  \end{subfigure}\hfill
  \begin{subfigure}[t]{0.23\textwidth}
    \centering
    \includegraphics[width=\linewidth]{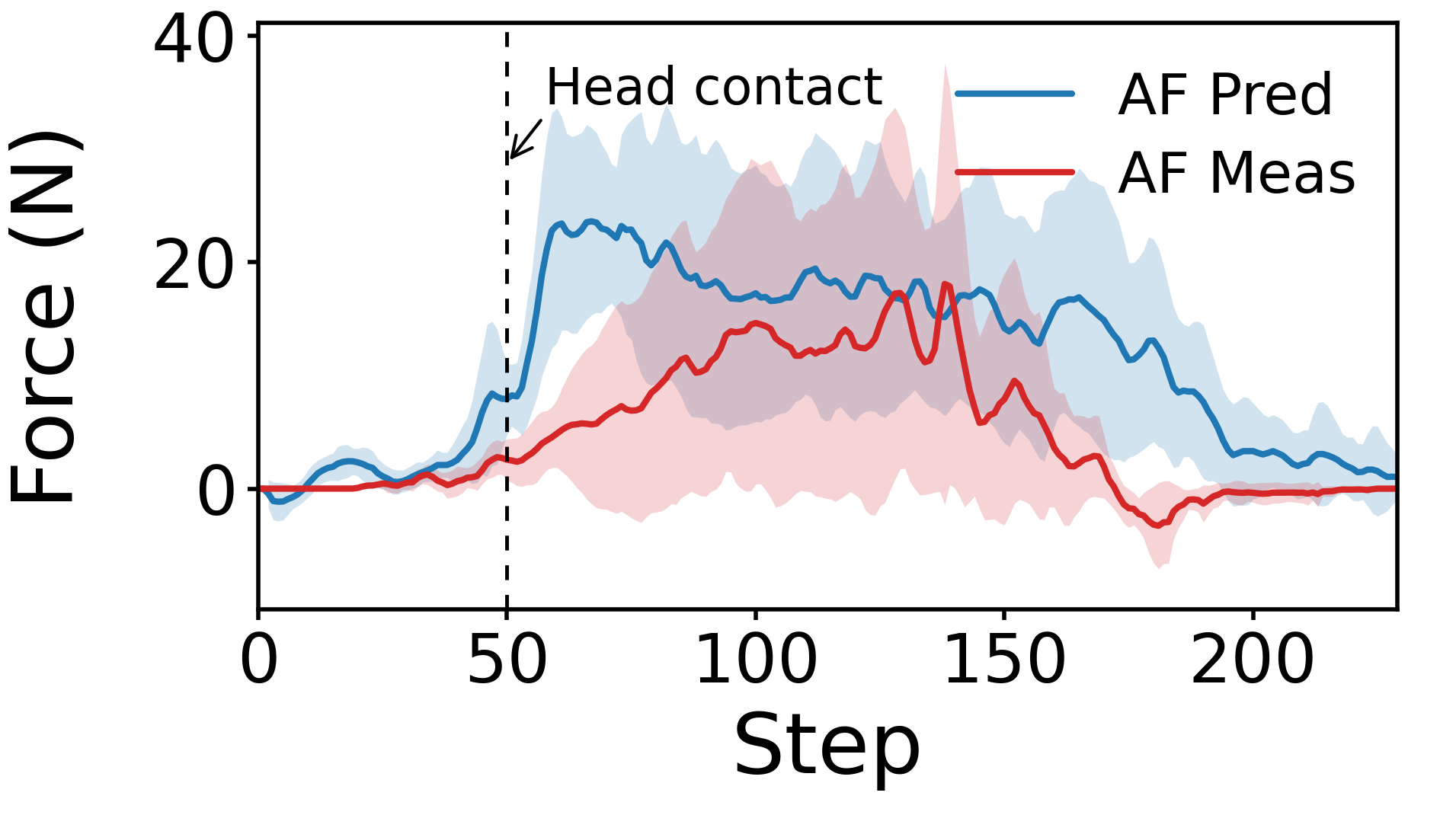}
    \caption{Card $F_x$, on}
  \end{subfigure}\\[0.5em]
  \begin{subfigure}[t]{0.23\textwidth}
    \centering
    \includegraphics[width=\linewidth]{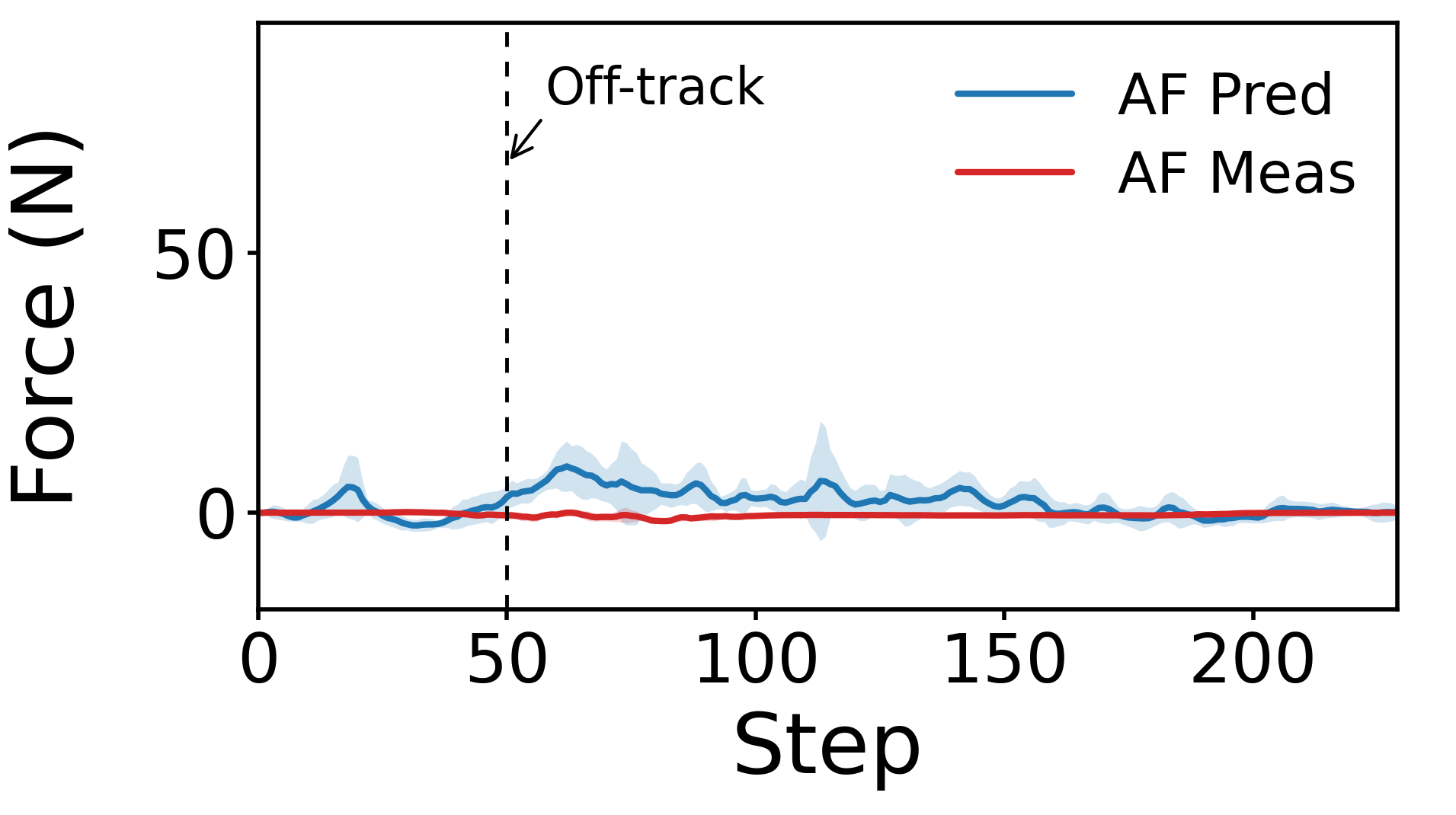}
    \caption{Card $F_z$, off}
  \end{subfigure}\hfill
  \begin{subfigure}[t]{0.23\textwidth}
    \centering
    \includegraphics[width=\linewidth]{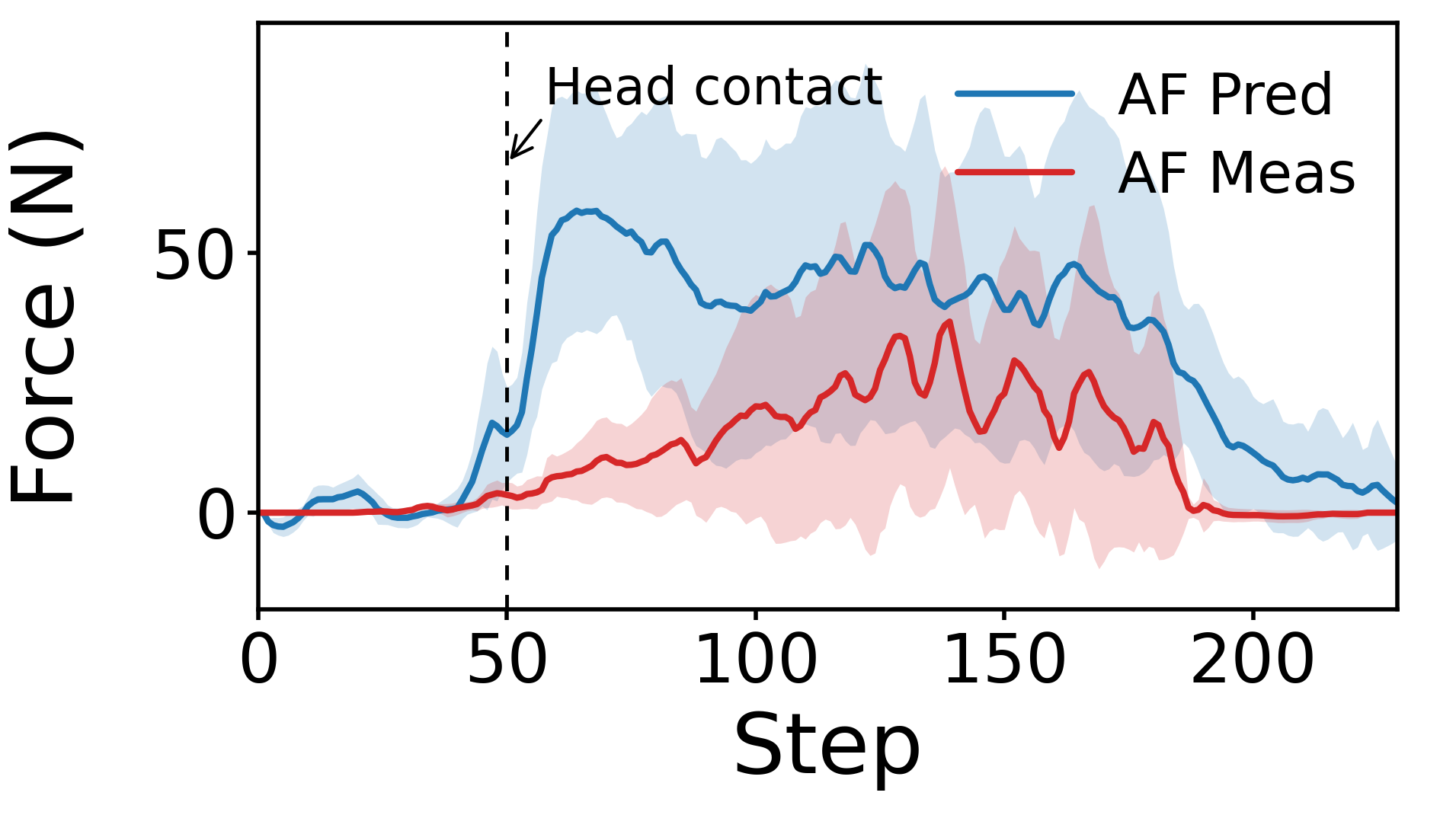}
    \caption{Card $F_z$, on}
  \end{subfigure}\hfill
  \begin{subfigure}[t]{0.23\textwidth}
    \centering
    \includegraphics[width=\linewidth]{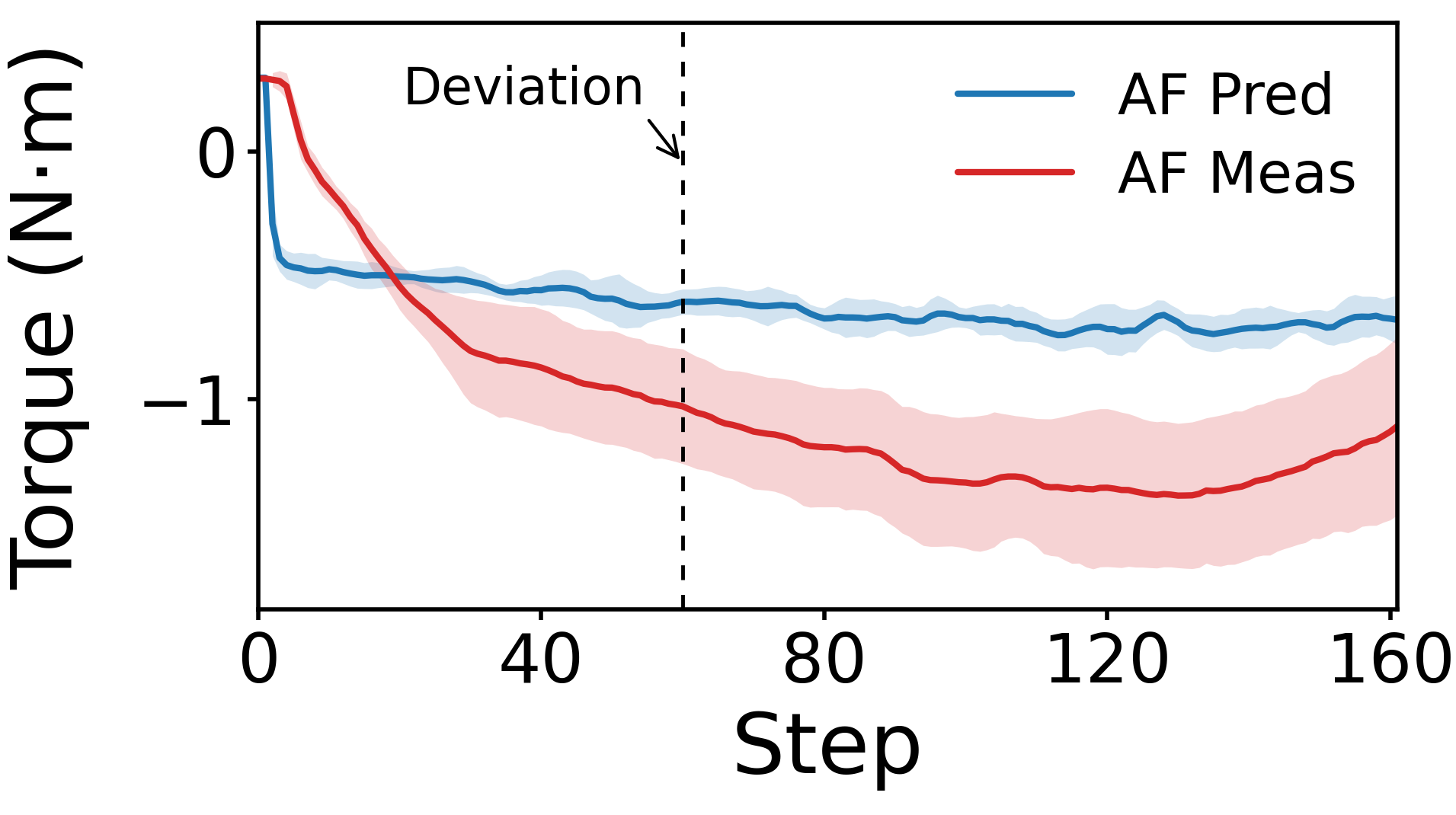}
    \caption{Fan $\tau_z$, off}
  \end{subfigure}\hfill
  \begin{subfigure}[t]{0.23\textwidth}
    \centering
    \includegraphics[width=\linewidth]{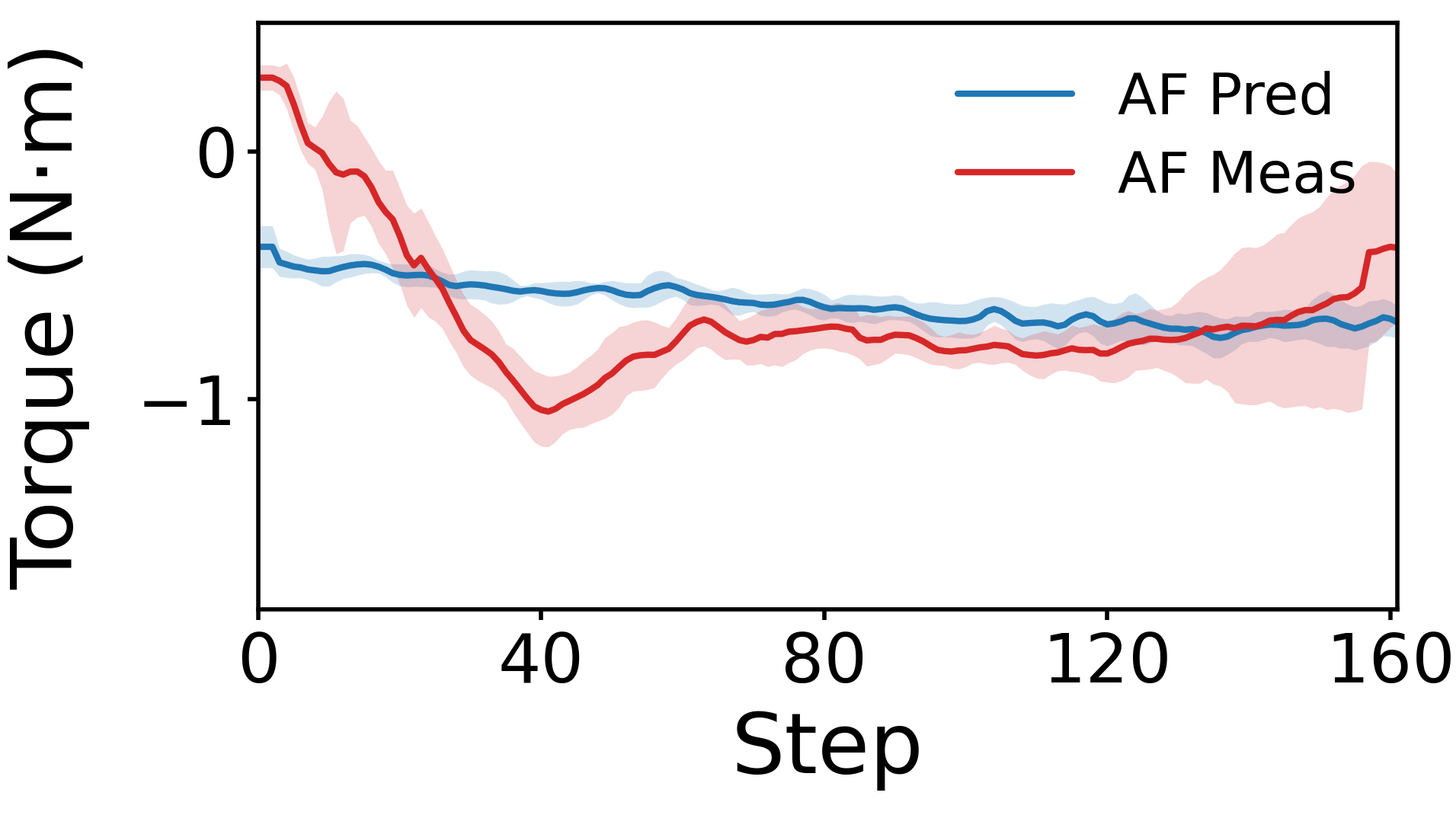}
    \caption{Fan $\tau_z$, on}
  \end{subfigure}
  \caption{\textbf{Gripper force control.} Predicted (blue) and measured (red) forces; shaded bands show run-to-run standard deviation. (a)--(d) Simulated gripper-controller ablations: (a) full force under hybrid control; (b) reduced force under hybrid control; (c) reduced force without the feedforward term; (d) reduced force without the admittance term; GF denotes grip force and Slip marks object loss. (e)--(l) The same controller on the real robot: blackboard wiping ($F_z$), card swiping ($F_x$ and $F_z$), and fan-knob turning ($\tau_z$), each with the loop off and on; AF denotes applied force in the TCP panels. With the loop on, the measured force tracks the predicted command closely; with the loop off, the two diverge, and the card and fan trials drift off the trajectory.}
  \label{fig:force_feedback}
\end{figure}

\begin{figure}[!htbp]
  \centering
  \includegraphics[width=0.54\textwidth]{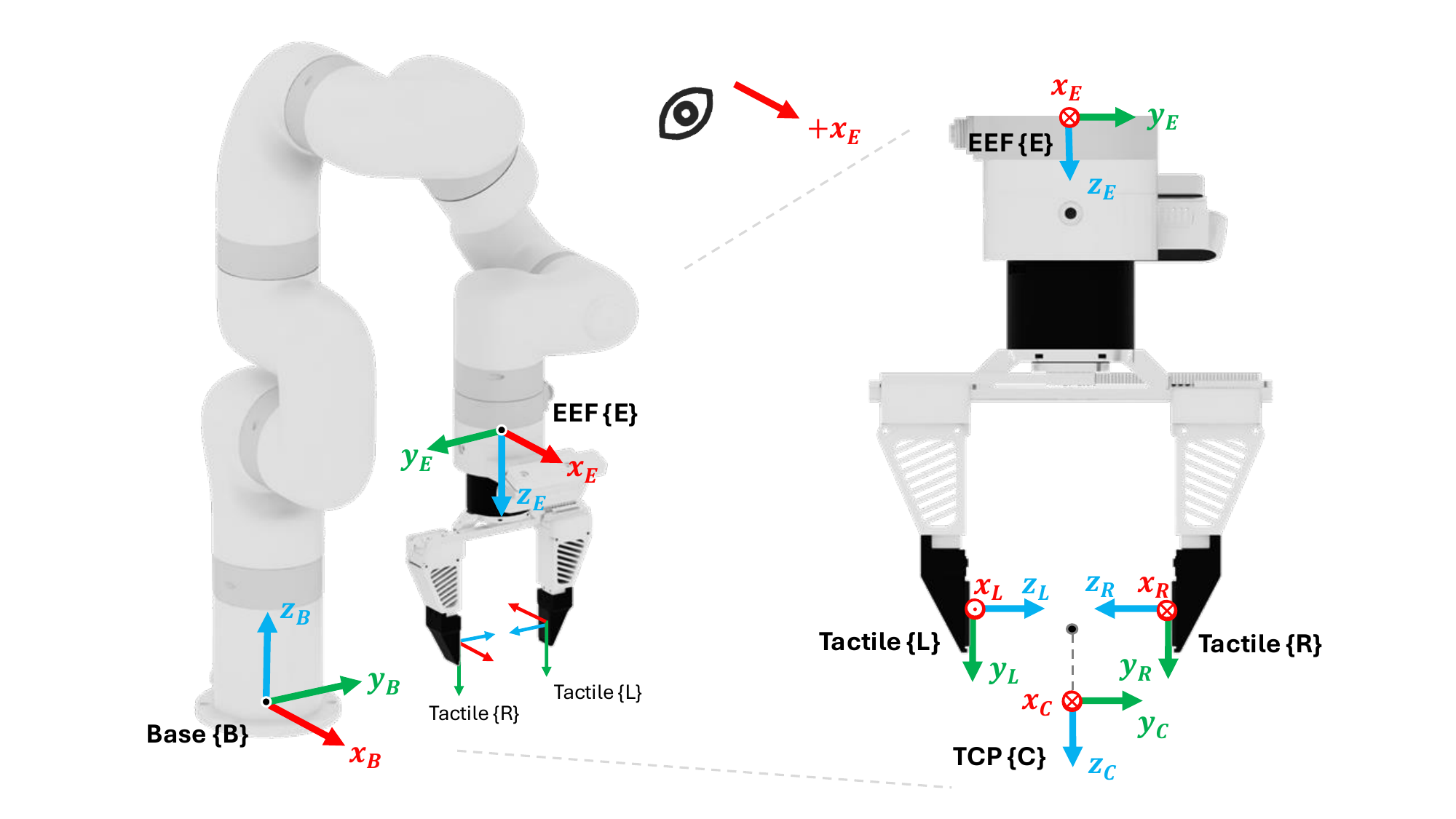}
  \caption{\textbf{Coordinate frames used by the force controller.} The robot base $\{B\}$, end-effector $\{E\}$, TCP $\{C\}$, and left and right tactile-sensor $\{L\}$, $\{R\}$ frames. The figure's TCP label $\{C\}$ corresponds to frame $T$ in the text and is distinct from the Quest controller frame $C$; $\{L\}$ and $\{R\}$ correspond to $S_L$ and $S_R$. TCP forces are expressed in $T$, and the grip force is computed from the sensor-frame contact reactions.}
  \label{fig:control_frames}
\end{figure}

To isolate the contribution of the low-level controller from the high-level policy, we evaluate on a mixture of the Tabero \textbf{Dataset A} and Taco special-task data, with 30 demonstrations per task collected by teleoperation. In Tabero, we use the best-performing Field+FS model and conduct four ablation studies on the gripper controller: (a) full force with hybrid control, (b) reduced force with hybrid control, (c) reduced force without feedforward term, and (d) reduced force without admittance component. Panels~(a)--(d) of Fig.~\ref{fig:force_feedback} demonstrate that both feedforward and admittance terms are essential for accurate force tracking; without closed-loop control, objects slip, causing task failure. Fig.~\ref{fig:control_frames} defines the base, end-effector, TCP, and tactile-sensor frames used to express these forces.

\begin{figure}[!htbp]
  \centering
  \includegraphics[width=\textwidth]{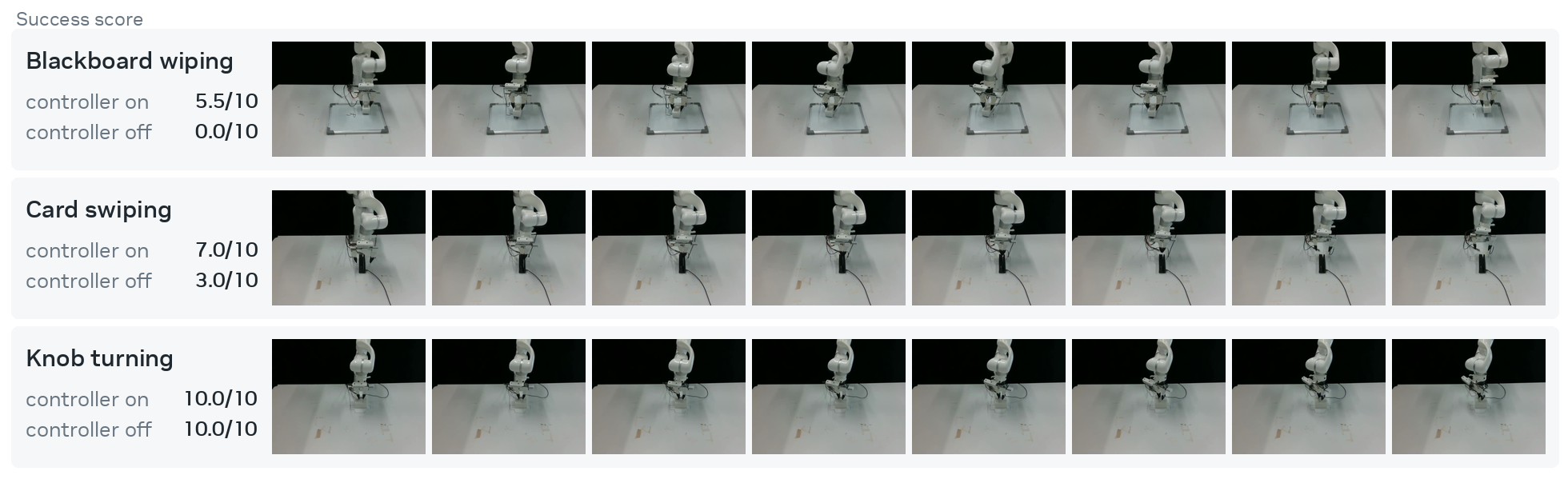}
  \caption{\textbf{Real-robot task expansion.} One row per task shows eight keyframes selected in time order and spaced so that adjacent frames show different stages of the motion. Scores at left are on a 10-point scale, computed over 10 trials for Field+FS ($\pi_{0.5}$, no state) with the loop on and off, using the scoring criteria in Appendix~\ref{app:task-success}.}
  \label{fig:real_task_expansion}
\end{figure}

\noindent We train and test the same marker-field model used in the preceding evaluation, Field+FS ($\pi_{0.5}$, no state), on three new real-robot tasks: blackboard wiping, card swiping, and knob turning. Appendix~\ref{app:task-success} defines their task-level success-score criteria, and Fig.~\ref{fig:real_task_expansion} presents the test results judged by these criteria together with illustrative process frames. Blackboard wiping requires controlling the downward pressure along the $z$-axis, card swiping requires maintaining a stable downward force along $z$ and a normal applied force along $x$, and knob turning requires controlling the torsional torque. Panels~(e)--(l) of Fig.~\ref{fig:force_feedback} show the force-tracking performance for the three tasks under closed-loop-on and closed-loop-off conditions, with 10 trials per condition. With the closed-loop controller on, the measured TCP force stays closer to the predicted command, most clearly where the command changes fastest, and the overall task success score is higher; disabling the controller reduces tracking accuracy and task scores. Blackboard wiping drops to zero, and card swiping falls from 7 to 3 points. Knob turning is less affected because gripper motion is relatively fixed and contact is concentrated on one side, although panels~(k) and~(l) of Fig.~\ref{fig:force_feedback} still show the difference in tracking.

\subsection{Data Synergy}
\label{sec:eval-cotrain}
\begin{figure}[!htbp]
  \centering
  \begin{subfigure}[t]{0.46\textwidth}
    \centering
    \includegraphics[width=\linewidth,height=3.3cm,keepaspectratio]{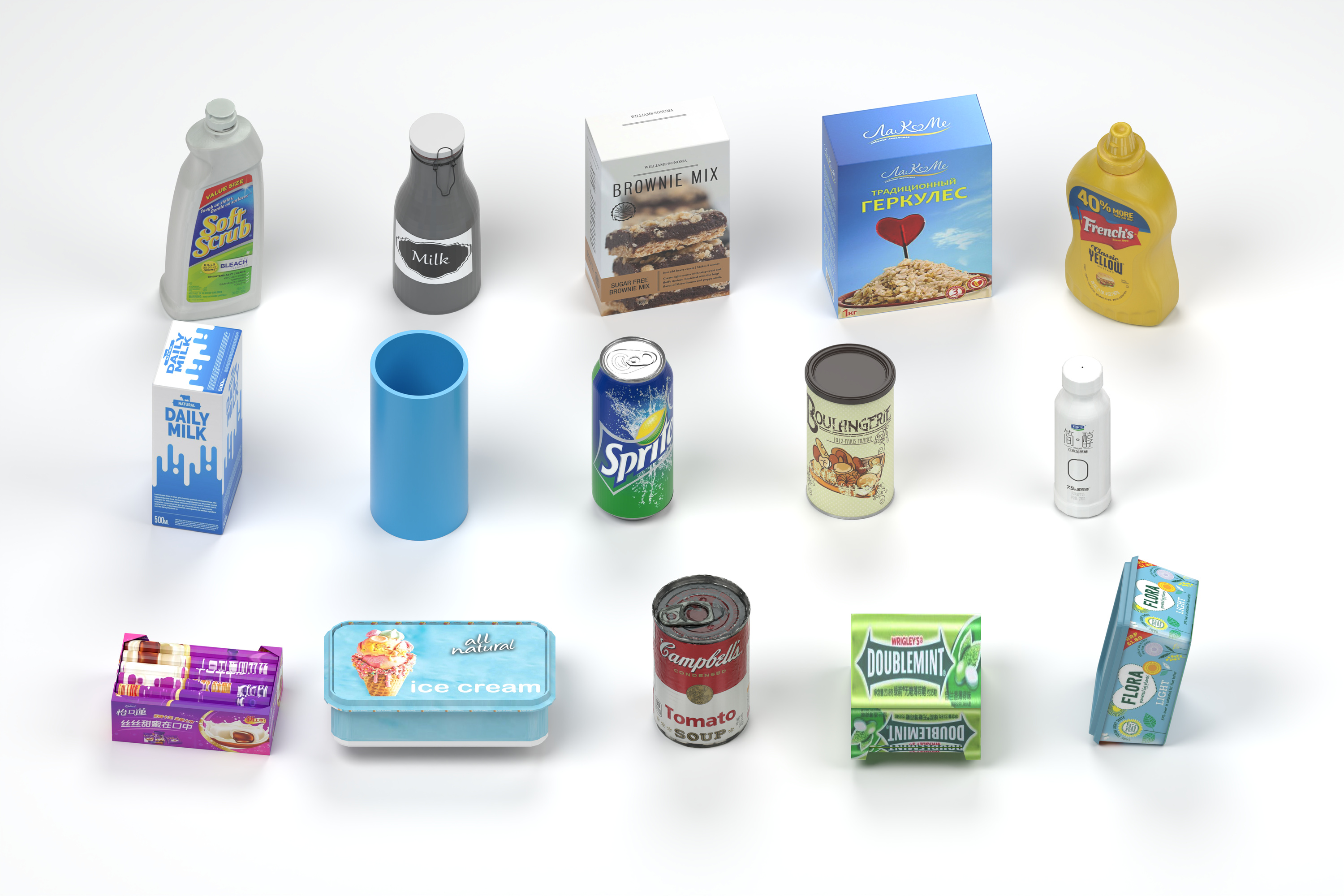}
    \caption{Simulation training objects}
  \end{subfigure}\hfill
  \begin{subfigure}[t]{0.46\textwidth}
    \centering
    \includegraphics[width=\linewidth,height=3.3cm,keepaspectratio]{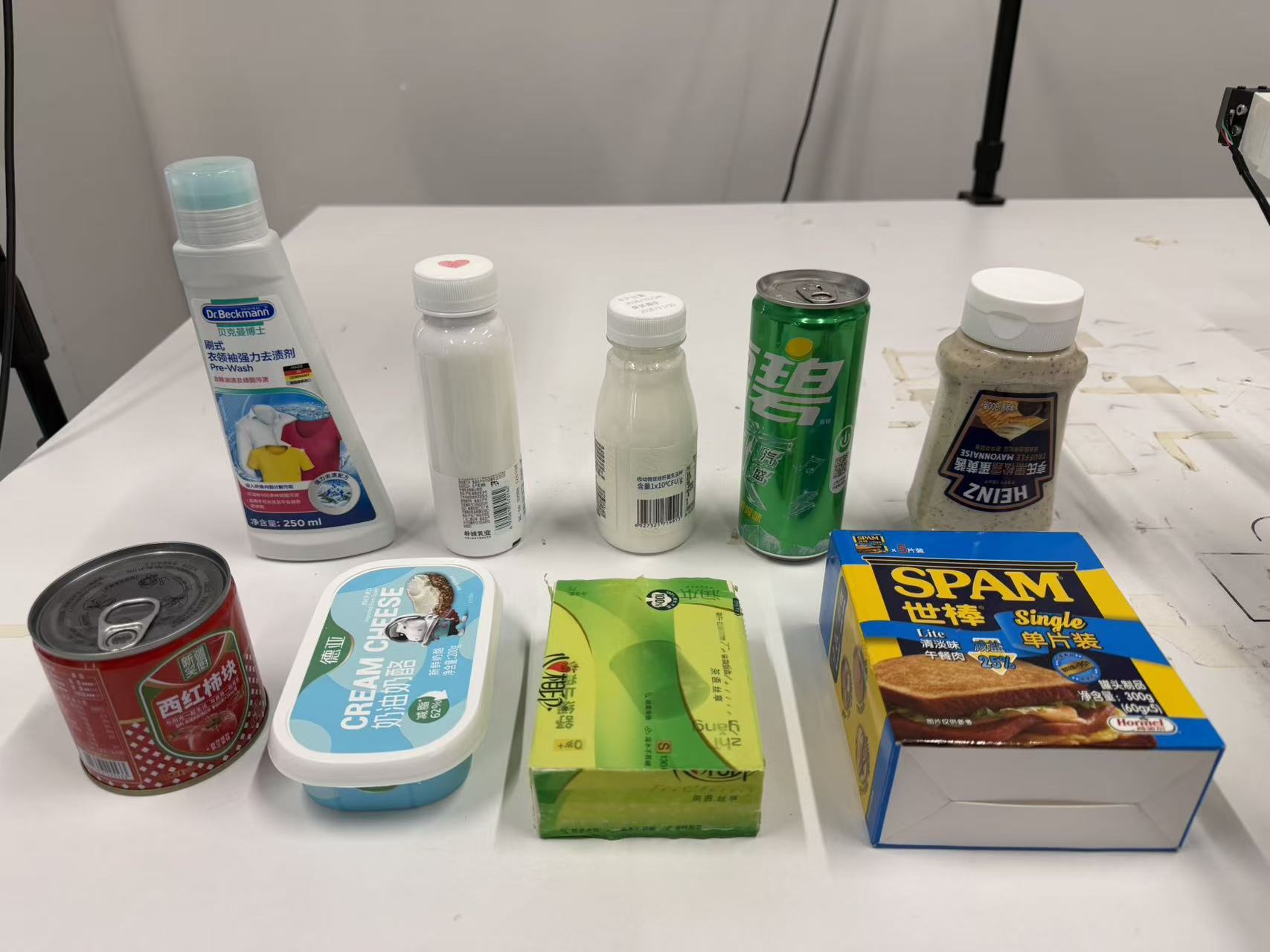}
    \caption{Real-robot test objects}
  \end{subfigure}
  \caption{\textbf{Objects used for mixed-data evaluation.} Simulation training assets (left) and physical test objects (right).}
  \label{fig:cotrain_objects}
\end{figure}

\noindent We evaluate whether the data augmentation and joint post-training workflow in Section~\ref{sec:taco-cotrain-loop} can combine the complementary strengths of sources with different costs and volumes. Teleoperation data cover fixed positions and objects. UMI data add variation in object position and approach path while retaining the same objects. Simulation data introduce variation in object instances or appearance at aligned task positions. Increasing the volume of a single source therefore increases sample density within its existing coverage; it does not directly add positions or object states absent from that source. Teleoperation data take longer to collect than UMI data, and simulation data have the lowest cost. Each data unit consists of 30 demonstrations for each of the six Taco base tasks, and each policy is trained directly on mixtures of different numbers of units using the same coordinate system, observation, action, timestep, and normalization interfaces. In the object-randomization tests, the physical objects differ entirely from the simulation training objects, as illustrated in Fig.~\ref{fig:cotrain_objects}; Appendix~\ref{app:randomization} details the test setup. Table~\ref{tab:cotrain_mix} reports six training mixes (only real, only umi, real+umi (1:1), sim+umi (5:1), real+sim (1:5), and real+sim+umi (1:5:1)) on the six original tasks, evaluated under position randomization and position+object randomization. All entries are firm/gentle success rates and the average grip force predicted by the policy on held-out real-robot scenes.

\noindent Fig.~\ref{fig:cotrain_objects} shows examples of the simulation training objects and real-robot test objects used to evaluate object generalization.

\begin{table}[t]
  \caption{Co-training results for teleoperation, simulation, and UMI data on the six original tasks with background disturbances. Each mix is evaluated under two randomization settings: position randomization, and position+object randomization. Training mixes are given as relative sample volumes; the UMI share is Tacumi demonstrations with object positions randomized within a range. F SR/G SR = success rates under firm/gentle force, reported as fractions; F AG/G AG = predicted average grip force (N) commanded by the policy, before real-robot tracking. A dash marks a condition with no recorded value. Bold and underlined values indicate the best and second-best results, respectively.}
  \label{tab:cotrain_mix}
  \centering
  \small
  \setlength{\tabcolsep}{4pt}
  \begin{tabular}{lcccccccc}
    \toprule
    & \multicolumn{4}{c}{\makecell{Position\\randomized}} & \multicolumn{4}{c}{\makecell{Position + object\\randomized}} \\
    \cmidrule(lr){2-5}\cmidrule(lr){6-9}
    Training Mix & F SR & G SR & F AG (N) & G AG (N) & F SR & G SR & F AG (N) & G AG (N) \\
    \midrule
    only real & 0.300 & 0.333 & 13.52 & 7.81 & 0.200 & 0.200 & 13.71 & 9.17 \\
    only umi & 0.000 & -- & -- & -- & 0.000 & -- & -- & -- \\
    real+umi (1:1) & \textbf{0.833} & \underline{0.700} & 17.96 & 8.44 & 0.100 & 0.100 & 13.36 & 10.60 \\
    sim+umi (5:1) & \underline{0.700} & 0.633 & 21.47 & 9.15 & \underline{0.400} & \underline{0.400}  & 24.66 & 9.34 \\
    real+sim (1:5) & 0.433 & 0.367 & 21.57 & 9.73 & 0.200 & \underline{0.400} & 25.84 & 8.27 \\
    real+sim+umi (1:5:1) & 0.567 & \textbf{0.833} & 22.04 & 10.47 & \textbf{0.600} & \textbf{0.600} & 24.08 & 13.32 \\
    \bottomrule
  \end{tabular}
\end{table}

\paragraph{Key findings.}

\noindent UMI-only training fails in the evaluated low-data setting, with 0\% success under the firm condition in both randomization settings. UMI demonstrations contribute when combined with teleoperation or simulation trajectories, supporting their role as a source of spatial variation in mixed-data training. Under position randomization, every multi-source mixture improved on the real-only baseline. Mean success was highest for real+umi, whose data most closely match the test scenarios, followed by real+sim+umi. Adding UMI to real data raised firm success from 30.0\% to 83.3\% and gentle success from 33.3\% to 70.0\%. Under position+object randomization the ordering changed: real+umi and real-only both showed substantial drops, indicating that the distribution shift makes out-of-distribution tasks difficult, whereas sim+umi reached 40.0\% mean success, real+sim 30.0\%, and real+sim+umi 60.0\%. The three-source mixture improved mean success over real-only by 38.3 and 40.0 percentage points in the two settings, the only multi-source mix with a mean gain above 30 points in both settings. These results reflect the complementary coverage of the sources: UMI demonstrations supply object-position and approach variation, simulation supplies object-appearance and background variation, and real trajectories anchor the sensor and actuator statistics. Data volume also matters, but the setup and results of real+sim show that the capability gain does not rely solely on more data; the characteristics and diversity of the data themselves matter more.

\noindent Every mixture with recorded force statistics predicts a higher mean force under firm than under gentle instructions in both settings. We attribute the larger separation mainly to increased data volume. In particular, adding substantial simulation data brings the force predictions closer to the original teleoperation labels.

\noindent The three data sources form a data pyramid that balances data efficiency and collection effort. Their combination increases mean success on held-out real-robot tasks with randomized positions and out-of-distribution objects from 20.0\% to 60.0\%. In this low-data setting, UMI-only training fails, while UMI data are more effective when combined with other sources. The sim+umi mixture also performs well under randomized conditions without teleoperation data. These results demonstrate complementary contributions from the three sources and improved task performance through co-training with limited data.

\section{Conclusions}
\label{sec:conclusion}

We present Tacinf, a modular infrastructure for tactile foundation models in force-aware manipulation. Its policy interface predicts actions and fingertip-force targets from task language, vision, and touch, with optional language-conditioned force levels for tasks that require explicit force regulation. Our ablations distinguish the roles of force-target supervision, high-level tactile observations, and feedback during execution. In the evaluated simulation setting, force supervision and the force-target interface account for the main performance gains, while some tactile representations further improve task performance. Real-robot controller experiments demonstrate how feedback contributes to force tracking and task execution. Policies adapted solely on simulation data transfer to the real robot without further policy fine-tuning. Across eight paired policy variants and six task conditions (three tasks, each with firm and gentle force levels), simulation closely preserves the real-robot ranking by success rate, with a small gap in the reported force metrics. Co-training on teleoperation, simulation, and UMI data balances data efficiency and collection effort. The three-source mixture increases mean success on held-out real-robot tasks with randomized positions and out-of-distribution objects from 20.0\% to 60.0\%; this gain cannot be attributed solely to increased data volume. In this low-data setting, UMI-only training fails, but UMI data are more effective when combined with other sources. The sim+umi mixture also performs well under randomized conditions without teleoperation data.

\noindent \textbf{Limitations.} Resource constraints limit the scale of the datasets and tasks in this study, and we do not evaluate large-scale training. Future work will expand the data pyramid and extend the system to additional robot embodiments and bimanual manipulation.

\appendix
\newpage
\clearpage

\section{Training and Interface Contract}
\label{App:hyper}
Detailed training hyperparameters and model configurations, including the complete configuration of each variant, will be released with the code. Table~\ref{tab:app_tac_variants} lists the tactile input variants used in the ablations. The Xense values reported in this paper describe the observation interface available to the policy; they do not constitute a mechanically validated characterization of the sensor.

\begin{table}[htb]
  \caption{Tactile input variants used in the ablations. ``Prefix'' and ``suffix'' denote the injection location in the backbone; ``FS'' denotes force supervision.}
  \label{tab:app_tac_variants}
  \centering
  \small
  \begin{tabular}{lcccc}
    \toprule
    Variant & Visual streams & Tactile stream & Adapter & FS \\
    \midrule
    Img+FS & 3 & -- & -- & yes \\
    Img & 3 & -- & -- & no \\
    Force D+FS & 2 & suffix & MLP & yes \\
    Force E & 2 & prefix & MLP & no \\
    Force E+FS & 2 & prefix & MLP & yes \\
    Field+FS & 2 & prefix & TCN & yes \\
    Field & 2 & prefix & TCN & no \\
    FS & 2 & -- & -- & yes \\
    None & 2 & -- & -- & no \\
    \bottomrule
  \end{tabular}
\end{table}

\section{Task Protocol}
\label{app:B}
\subsection{Language-Conditioned Force Protocol}
\label{app:force-language}
For each base instruction, force-related adverbs are inserted either before or after the verb phrase and the insertion position is randomized. The in-domain pairs are ``firmly/tightly'' for the high-force condition and ``gently/softly'' for the low-force condition. Unseen adverbs such as ``lightly'' and ``forcefully'' are reserved for the generalization test and never enter training. Table~\ref{tab:tabero-tasks-1} gives example templates.

\begin{table}[htb]
  \caption{Example instruction templates for force-language conditioning.}
  \label{tab:tabero-tasks-1}
  \centering
  \small
  \begin{tabular}{cl}
    \toprule
    Template & Instruction \\
    \midrule
    Prefix-high & \textit{firmly pick up the alphabet soup and place it in the basket} \\
    Suffix-high & \textit{pick up the alphabet soup and place it in the basket tightly} \\
    Prefix-low & \textit{softly pick up the alphabet soup and place it in the basket} \\
    Suffix-low & \textit{pick up the alphabet soup and place it in the basket gently} \\
    \bottomrule
  \end{tabular}
\end{table}

\noindent The Tabero subset contains the nine base tasks listed in Table~\ref{tabero-raw-9-tasks}. The force adverbs are applied to every task using the protocol above.

\begin{table}[htb]
  \caption{Base tasks in the Tabero subset.}
  \label{tabero-raw-9-tasks}
  \centering
  \small
  \begin{tabular}{rl}
    \toprule
    ID & Task description \\
    \midrule
    0 & pick up the alphabet soup and place it in the basket \\
    1 & pick up the cream cheese and place it in the basket \\
    2 & pick up the salad dressing and place it in the basket \\
    3 & pick up the bbq sauce and place it in the basket \\
    4 & pick up the tomato sauce and place it in the basket \\
    5 & pick up the butter and place it in the basket \\
    6 & pick up the milk and place it in the basket \\
    7 & pick up the chocolate pudding and place it in the basket \\
    8 & pick up the orange juice and place it in the basket \\
    \bottomrule
  \end{tabular}
\end{table}

\subsection{Task-Specific Success Criteria}
\label{app:task-success}

\paragraph{Blackboard Wiping.}
The score is computed from the amount of handwriting remaining after the wipe, compared against the demonstration data. If the residue reaches the level observed in human demonstrations, the score is 1; if minor residue remains (less than one third of the handwriting length), the score is 0.5; if significant residue remains, the score is 0.

\paragraph{Card Swiping.}
A trial is considered successful when the card traverses the predefined path in the intended direction while remaining functionally engaged with the slot and correctly oriented. The card need not remain in contact with the bottom of the slot, but its lower edge should stay within the vertical range observed in human demonstrations. The swipe should proceed continuously, without sustained pauses or substantial reverse motion.

\paragraph{Knob Turning.}
A trial is considered successful when the gripper maintains stable bilateral contact with the knob and continuously drives it with approximately steady rotational motion. Knob rotation should remain coupled with the end-effector motion throughout the manipulation. Fingertip slip or gripper rotation without corresponding knob motion is considered a failure. Rotation produced primarily through unilateral tangential pushing, incidental collision, or sliding against the knob is likewise treated as a failure. Minor variations in gripper position and orientation are acceptable if they remain within the demonstration-derived range.

\subsection{Position and Object Randomization Protocol}
\label{app:randomization}
\begin{figure}[H]
  \centering
  \begin{subfigure}[t]{\textwidth}
    \centering
    \includegraphics[width=\linewidth]{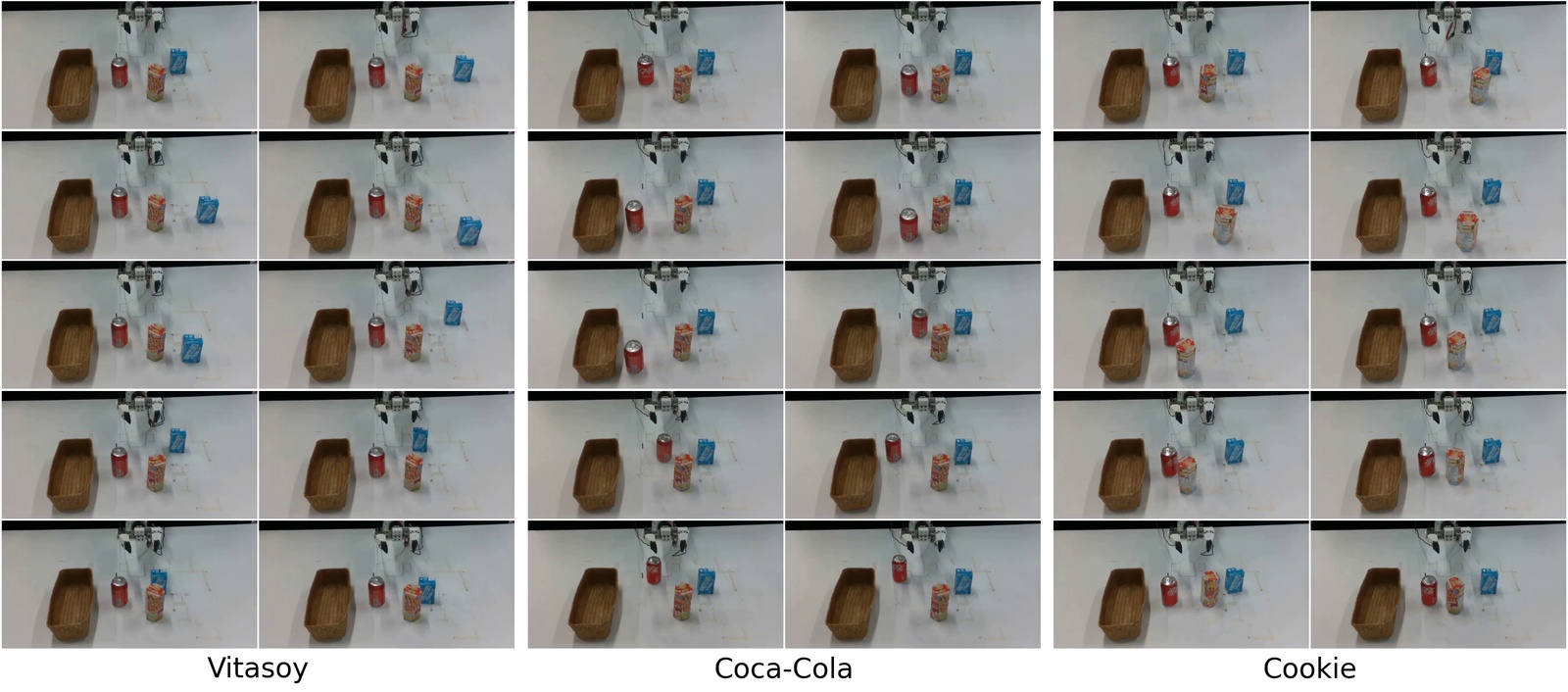}
    \caption{Position randomization}
  \end{subfigure}\\[0.6em]
  \begin{subfigure}[t]{\textwidth}
    \centering
    \includegraphics[width=\linewidth]{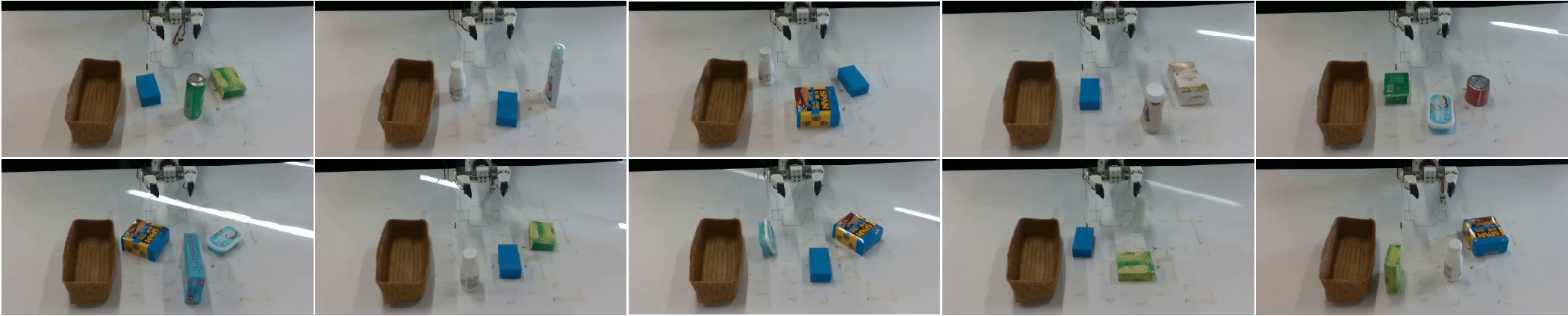}
    \caption{Position and object randomization}
  \end{subfigure}
  \caption{\textbf{Randomized task configurations.} (a) Position variation with the original Vitasoy, Coca-Cola, and cookie objects. (b) Joint position and object variation. The basket remains the placement goal in both settings.}
  \label{fig:cotrain_randomization}
\end{figure}

The real-robot evaluation in Section~\ref{sec:eval-cotrain} comprises six Taco tasks: three with randomized initial object positions and three with both randomized initial positions and object instances (Fig.~\ref{fig:cotrain_randomization}).

\subsection{Additional VR Teleoperation Tasks}
\begin{figure}[H]
  \centering
  \includegraphics[width=\linewidth]{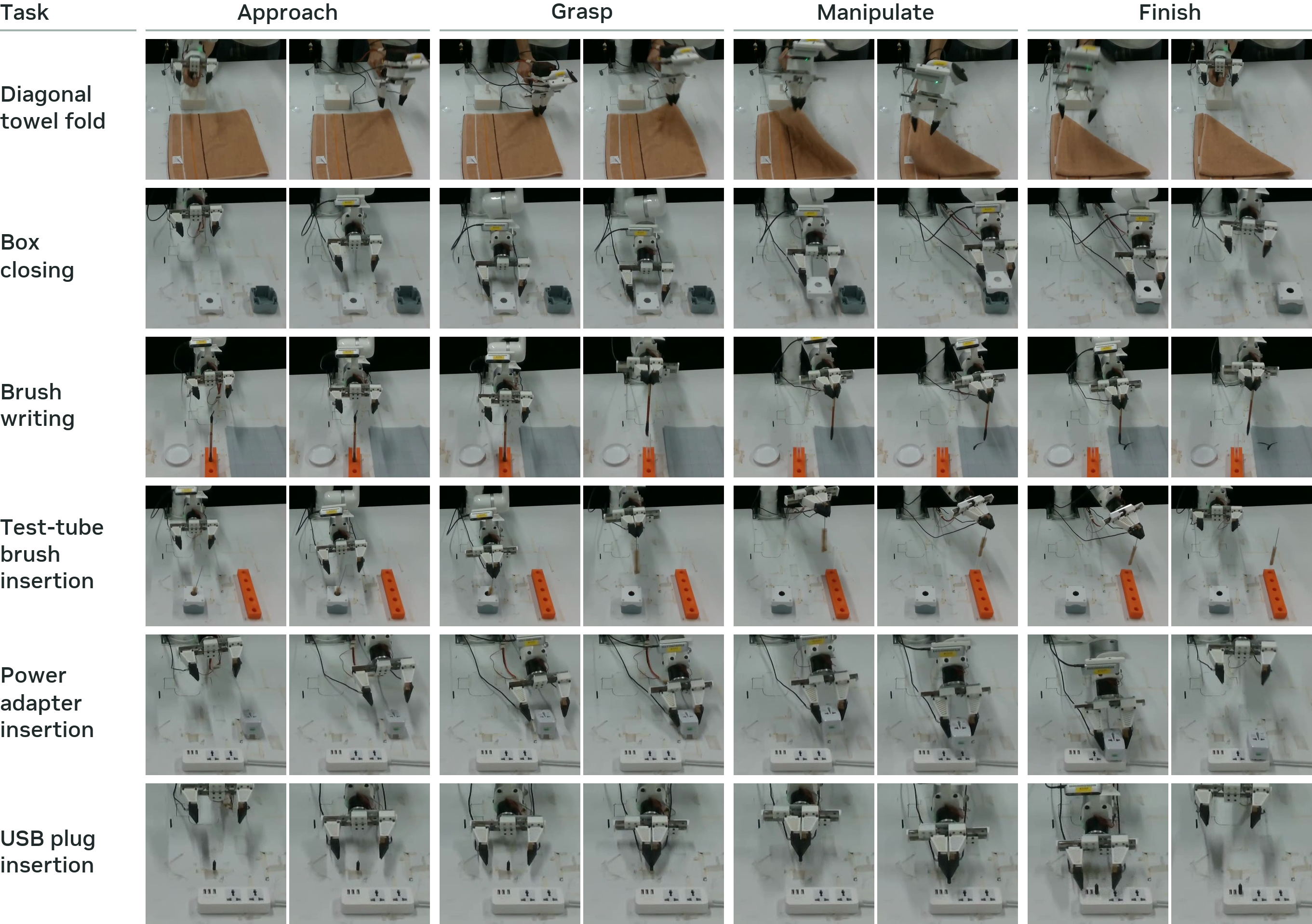}
  \caption{\textbf{VR teleoperation demonstrations.} Each row shows eight frames from one third-person recording, ordered from left to right with two frames per stage.}
  \label{fig:real_world_demonstrations}
\end{figure}

The six additional tabletop tasks we recorded using VR teleoperation were folding a towel diagonally, fitting a lid onto a box, writing on a mat with a brush, inserting a brush into a test tube, plugging a power adapter into a power strip, and inserting a USB plug into the strip's USB port (Fig.~\ref{fig:real_world_demonstrations}).

\section{Supplementary Evaluation}
\label{app:C}
\subsection{Retention Summary}
We define retention as the fraction of replay episodes that complete the task successfully, $R=N_s/N_t$, where $N_s$ is the number of episodes that complete the task and $N_t$ is the number of replayed episodes.

\subsection{Semantic-Force Generalization}
\label{app:semantic-force}
Table~\ref{tab:semantic_force} presents the in-domain and out-of-domain adverb evaluation.

\begin{table}[htb]
  \caption{Semantic-force understanding on Dataset B. Force values are in N.}
  \label{tab:semantic_force}
  \centering
  \small
  \begin{tabular}{lccccc}
    \toprule
    Adverb & SR & AG & AA & MG & MA \\
    \midrule
    \textit{firmly, tightly} & 0.86 & 32.4 & 1.5 & 62.7 & 6.9 \\
    \textit{gently, softly} & 0.52 & 3.7 & 0.4 & 11.4 & 5.1 \\
    \textit{lightly} (OOD) & 0.61 & 14.5 & 0.8 & 40.8 & 7.6 \\
    \textit{forcefully} (OOD) & 0.64 & 18.1 & 0.9 & 49.7 & 6.3 \\
    \bottomrule
  \end{tabular}
\end{table}

This evaluation characterizes the relationship between language and force; it does not calibrate absolute force magnitudes.

\subsection{Img+FS Variants}
\label{app:imgfs_ablation}
Table~\ref{tab:app_imgfs_variants} reports the remaining Img+FS ($\pi_{0.5}$) variants under simulation.

\begin{table}[htb]
  \caption{Additional Img+FS ($\pi_{0.5}$) variants under simulation. The last column is the simulated firm--gentle force separation $\mathrm{F\,AG}-\mathrm{G\,AG}$.}
  \label{tab:app_imgfs_variants}
  \centering
  \small
  \begin{tabular}{lcc}
    \toprule
    Model & Sim SR (\%) & F AG $-$ G AG (N) \\
    \midrule
    Img+FS (state) & 86.35 & 16.21 \\
    Img+FS (no state) & 90.65 & 13.55 \\
    Img+FS (state, $H=1$) & 91.70 & 16.74 \\
    Img+FS (no state, $H=1$) & 95.65 & 15.39 \\
    \bottomrule
  \end{tabular}
\end{table}

\subsection{Sensor Measurement Validation}
\label{app:single_finger}
The force components reported by each Xense fingertip sensor depend on contact location, contact asymmetry, and sensor estimation error. A single sensor's output therefore cannot be treated as precise fingertip-force ground truth. We use the manufacturer's SDK estimates and sum the left and right force magnitudes to obtain a bilateral grip-force proxy on the real robot. The grasp configurations in Fig.~\ref{fig:sensor_measurement} are used to check whether this proxy varies plausibly with contact state.

\noindent Fig.~\ref{fig:sensor_measurement} compares the SDK-estimated resultant force components across five grasp configurations, some of which exhibit asymmetry between the fingertips. AG is computed using the approximation in Eq.~\eqref{eq:grip_force}. We therefore use real-robot force values to compare relative trends and changes in feedback during control.

\begingroup
\setlength{\intextsep}{6pt}
\begin{figure}[H]
  \centering
  \includegraphics[width=0.92\textwidth]{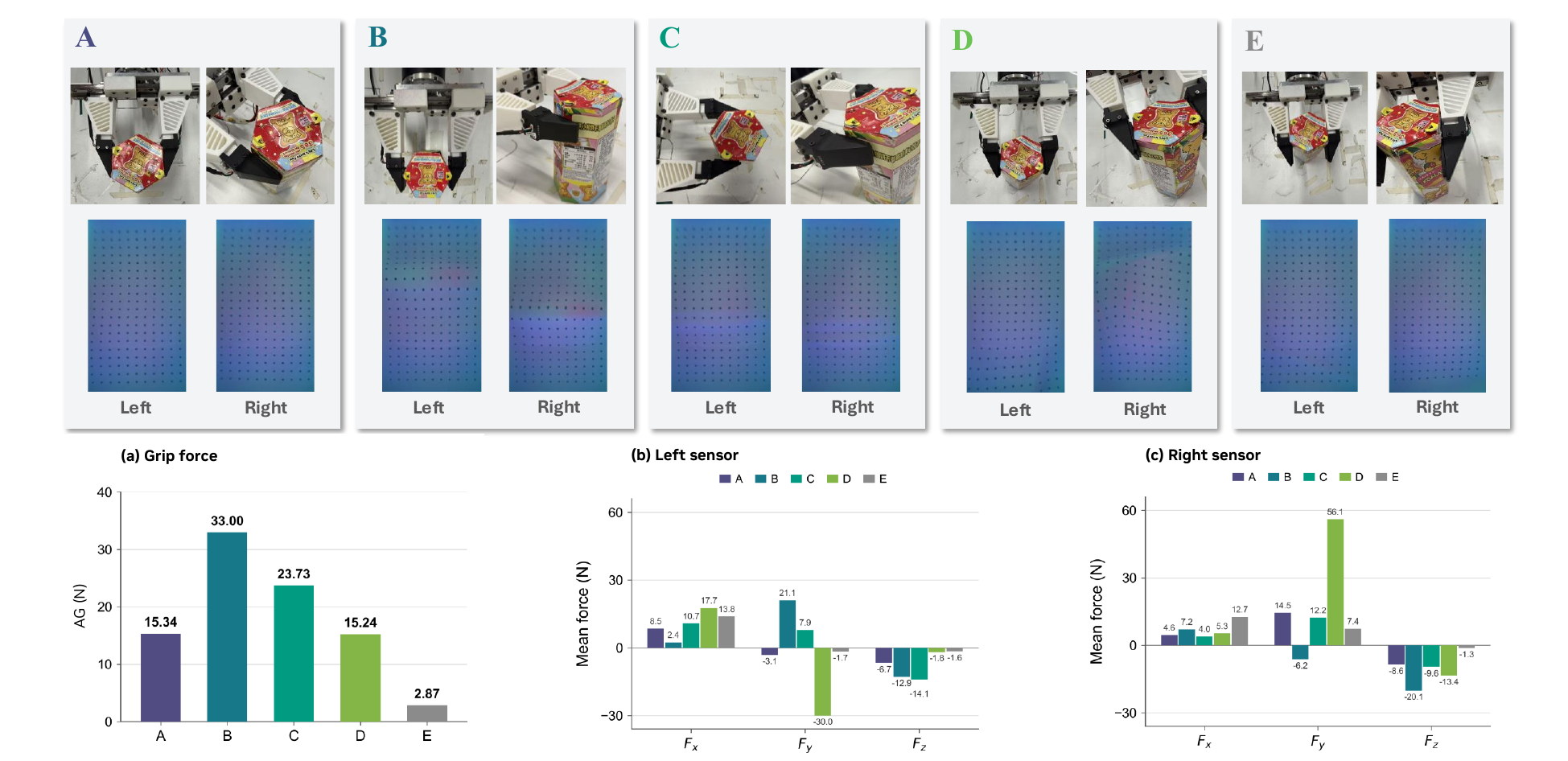}
  \caption{\textbf{Fingertip force measurements.} Top: configurations A--E and paired tactile images. Bottom: (a) grip force (AG), computed according to Eq.~\eqref{eq:grip_force}; (b,c) time-averaged components of the resultant force reported by the left and right tactile sensors.}
  \label{fig:sensor_measurement}
\end{figure}
\endgroup

\Needspace{8\baselineskip}
\subsection{UMI Data Distribution Validation}
\label{app:umi-validation}
\begin{figure}[H]
  \centering
  \includegraphics[width=0.7\textwidth]{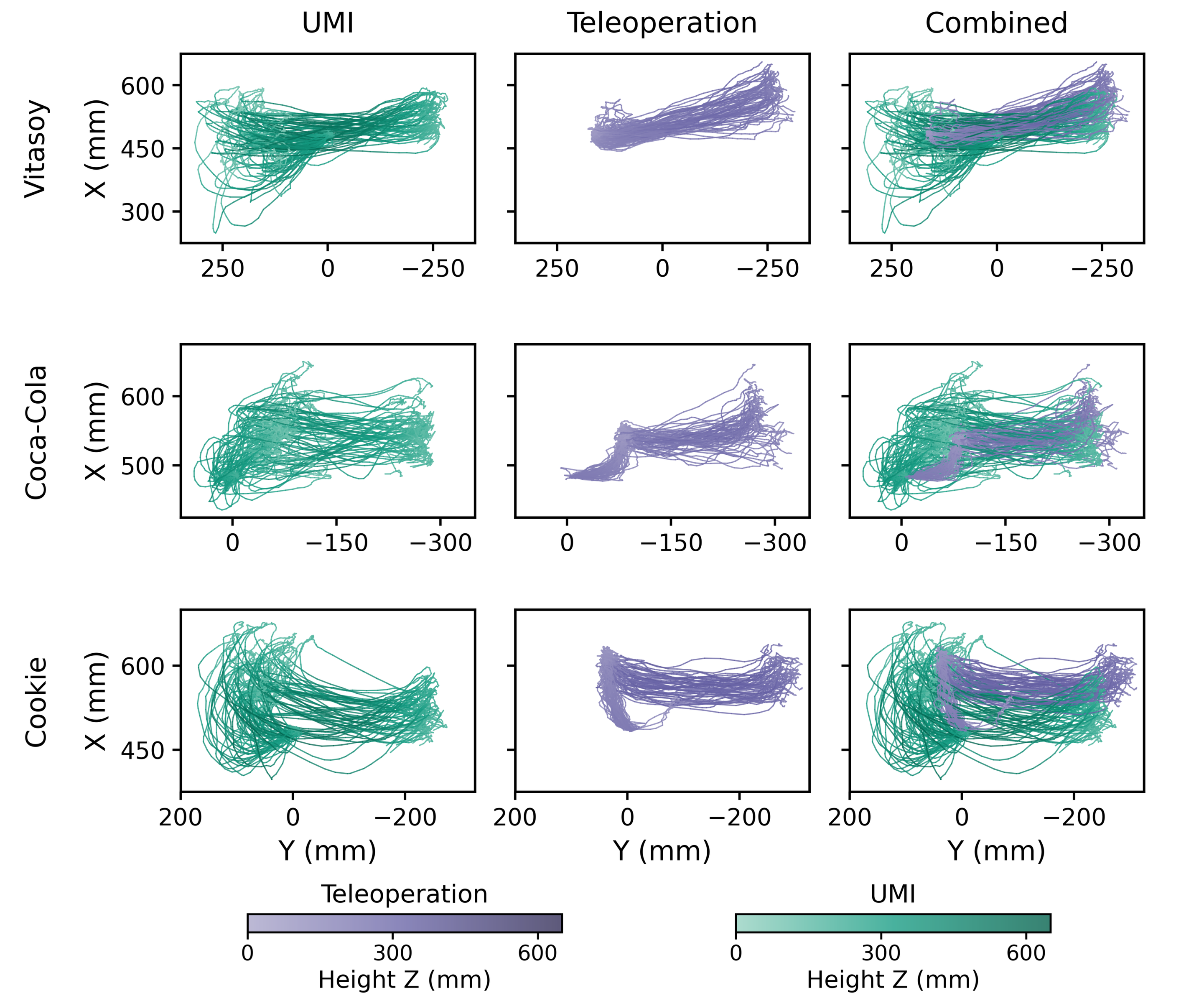}
  \caption{\textbf{Spatial coverage of UMI and teleoperation demonstrations.} Rows: Vitasoy, Coca-Cola, and cookie tasks. Columns: UMI, teleoperation, and their overlay. Curves show trajectory projections onto the $X$--$Y$ plane. Purple denotes teleoperation and green denotes UMI; color shade encodes height $Z$. All coordinates are in millimeters.}
  \label{fig:umi_distribution}
\end{figure}

Fig.~\ref{fig:umi_distribution} compares UMI and teleoperation trajectories for three manipulation tasks. The comparison shows overlapping task paths and additional spatial variation in the UMI demonstrations.

\section{Tactile Real2Sim Calibration}
\label{app:tactile_real2sim}
We calibrate the tactile observation interface for contact onset, patch persistence, force scale, marker response, and channel and temporal conventions.

\subsection{Depth-to-Indentation Mapping}
At each tactile update, the camera produces a depth map $D(u,v)$ in meters. We express the same map in millimeters as $Z(u,v)=1000\,D(u,v)$. Let $d_{\min}=\min_{u,v}D(u,v)$, $d_{\mathrm{ref}}$ be the camera-to-gel reference distance, and $h_g$ the effective gel thickness, all in meters. The distance offset $\delta$ (meters) and shared indentation depth $h_{\text{in}}$ (millimeters) are
\begin{equation}
  \delta=\max\!\left(0,d_{\min}-d_{\mathrm{ref}}\right),\qquad
  h_{\text{in}}=
  \begin{cases}
    (h_g-\delta)\,1000,&\delta\leq h_g,\\
    0,&\delta>h_g.
  \end{cases}
\end{equation}
The estimated indentation is positive when $d_{\min}<d_{\mathrm{ref}}+h_g$ and saturates at $1000h_g$ for $d_{\min}\leq d_{\mathrm{ref}}$. The relative depth field used to compute gel deformation is $Z'(u,v)=Z(u,v)-\min_{u,v}Z(u,v)-h_{\text{in}}$.

\subsection{Taxim and FOTS Interface}
Taxim~\cite{taxim} uses calibrated polynomials to map local gradients of the deformed gel and normalized image positions to RGB intensity changes, which are added to the real no-contact background. Pixels outside the contact mask retain the background. The output uses RGB ordering and no additional shadow term.

\noindent FOTS~\cite{FOTS} uses the same relative depth field $Z'$ and derived contact mask to approximate normal expansion, shear, and planar torsion with Gaussian spatial decay kernels
\begin{equation}
  g_k(x,y)=\exp\!\left[-\lambda_k\big((x-c_x)^2+(y-c_y)^2\big)\right],
\end{equation}
where marker positions $(x,y)$ and kernel centers $(c_x,c_y)$ are in image coordinates, and $\lambda_k$ controls the decay of component $k$. Contact-centroid changes provide the tangential cue, and relative sensor-object yaw provides the rotational cue. Marker displacements are expressed in sensor image coordinates. With zero indentation, the trajectory is cleared and markers return to their reference grid.

\subsection{Calibration and Scope}
The calibration bundle contains the real no-contact background, polynomial coefficients fitted for each gradient bin using real sphere-contact images, and the gel reference map. Camera geometry, pad dimensions, marker scale, spatial decay coefficients, and update settings are specified in the released simulator configuration. The virtual camera pose and depth clipping range align simulated observations with the operating depth range. They do not specify the physical sensor's internal geometry.

\noindent Policy observations follow the 20~Hz sampling convention and configured history length $H$. Force labels for the Taxim-FOTS backend come from Isaac contact sensors and follow Eq.~\eqref{eq:force_frames}; real-robot force measurements use the manufacturer's Xense calibration.

\section{Tacinf-Tacumi Calibration and Synchronization}
\label{app:tacumi_calibration_sync}
This section describes the spatial calibration and temporal convention of the handheld data collection system in Section~\ref{sec:tacinf-tacumi}. Spatial calibration maps the tracked Quest controller pose to the gripper center, while a common host time base relates the visual, tactile, pose, and motor-state streams.

\subsection{Hand-Eye Calibration}
\label{app:tacumi_handeye}
We estimate the rigid transformation between the Quest controller frame and the gripper-center frame using a robot-assisted optical hand-eye calibration procedure. The controller is rigidly mounted on the robot, and 20 paired poses are collected from the Quest tracking system and the robot. The gripper center defines the TCP frame $T$ used for the recorded manipulation trajectory. This calibration follows the geometric formulation used in controller-based handheld acquisition~\cite{li2025vitaminb}.

\begin{figure}[!t]
  \centering
  \includegraphics[width=0.69\textwidth]{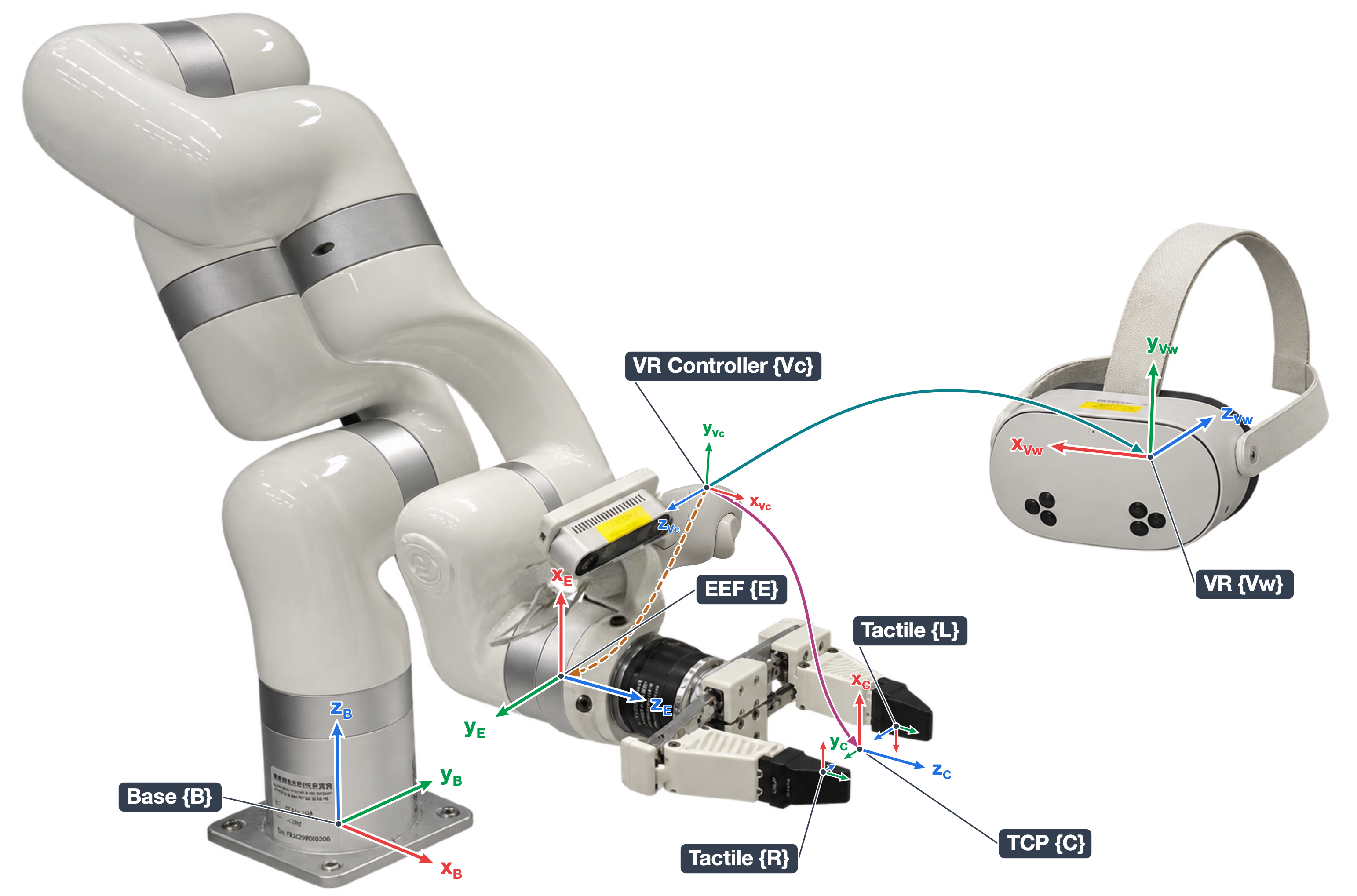}
  \caption{\textbf{Hand-eye calibration setup and coordinate frames of Tacinf-Tacumi.} The Quest controller is rigidly mounted with the gripper on an xArm~7. The magenta curve marks the fixed transform between the gripper-center TCP (labeled $C$) and the VR controller (labeled $\mathrm{Vc}$). In the calibration equations, $T$ denotes the TCP frame, $C$ the controller frame, and $W$ the Quest tracking frame labeled $\mathrm{Vw}$ in the figure. The robot base is labeled $B$.}
  \label{fig:tacumi_handeye_frames}
\end{figure}

\noindent Fig.~\ref{fig:tacumi_handeye_frames} illustrates the calibration setup and the coordinate-frame relationships.

\noindent Let $W$, $B$, $C$, and $T$ denote the Quest tracking reference frame, robot base frame, controller frame, and gripper-center TCP frame, respectively. A transformation $\mathbf{T}_{B}^{A}$ maps coordinates from frame $B$ to frame $A$. For calibration pose $i$, let $\mathbf{Q}_i=\mathbf{T}_{C,i}^{W}$ be the tracked controller pose and $\mathbf{P}_i=\mathbf{T}_{T,i}^{B}$ be the corresponding robot-side pose expressed at the TCP. The unknown fixed extrinsic $\mathbf{X}=\mathbf{T}_{T}^{C}$ between the TCP and controller frames and the fixed reference-frame transformation $\mathbf{Z}=\mathbf{T}_{B}^{W}$ satisfy
\begin{equation}
    \mathbf{Q}_i\mathbf{X}=\mathbf{Z}\mathbf{P}_i,
    \qquad i=1,\ldots,20.
    \label{eq:app_tacumi_handeye_absolute}
\end{equation}
Eliminating $\mathbf{Z}$ between two observations in Eq.~\eqref{eq:app_tacumi_handeye_absolute} yields the relative-motion hand-eye equation
\begin{equation}
    \underbrace{\mathbf{Q}_j^{-1}\mathbf{Q}_i}_{\mathbf{A}_{ji}}\mathbf{X}
    =\mathbf{X}\underbrace{\mathbf{P}_j^{-1}\mathbf{P}_i}_{\mathbf{B}_{ji}},
    \label{eq:app_tacumi_handeye_relative}
\end{equation}
which has the standard form $\mathbf{A}\mathbf{X}=\mathbf{X}\mathbf{B}$. We solve for the fixed transformation $\mathbf{X}$ in Eq.~\eqref{eq:app_tacumi_handeye_relative} from the collected pose pairs. The notation assumes that the two pose streams are expressed using consistent coordinate conventions and length units.

\noindent During handheld collection, the calibrated transformation converts each controller pose to the TCP pose as $\mathbf{T}_{T}^{W}(t)=\mathbf{T}_{C}^{W}(t)\mathbf{X}$, matching the TCP transformation in Section~\ref{sec:tacinf-tacumi}. This mapping accounts for both the translational offset and rotational alignment between the controller and the grasping tool. The calibration remains applicable while their relative mounting geometry is unchanged; changing that geometry requires updating the extrinsic transformation.

\subsection{Multimodal Temporal Synchronization}
\label{app:tacumi_sync}
The collection system acquires Quest controller poses, visual images, tactile observations, and gripper motor states at different native rates, as summarized in Table~\ref{tab:app_tacumi_rates}. Timestamps for all four streams are generated on the same host computer, providing a common temporal reference for recording the demonstrations.

\begin{table}[htb]
  \caption{Native acquisition rates of Tacinf-Tacumi. All streams use timestamps generated on the same host computer. These are data acquisition rates; policy execution uses a separate frequency.}
  \label{tab:app_tacumi_rates}
  \centering
  \small
  \begin{tabular}{lc}
    \toprule
    Data stream & Acquisition rate (Hz) \\
    \midrule
    Quest controller pose & 72 \\
    Visual camera & 30 \\
    Tactile sensor & 100 \\
    Gripper motor state & 200 \\
    \bottomrule
  \end{tabular}
\end{table}

\noindent For modality $m$, the recorded stream is represented as $\mathcal{D}_m=\{(t_{m,k},\mathbf{z}_{m,k})\}_k$, where $t_{m,k}$ is the host-generated timestamp and $\mathbf{z}_{m,k}$ denotes the corresponding measurement. Because the sampling rates differ, sample indices do not directly identify simultaneous observations across modalities. The shared timestamp reference provides the basis for associating gripper motion and motor state with visual and tactile observations.

\noindent  A common host time base establishes timestamp comparability but does not by itself establish simultaneous physical acquisition. Sensor acquisition, transport, and buffering can introduce modality-dependent delays; prior handheld systems explicitly distinguish timestamp alignment from per-modality latency compensation~\cite{li2025vitaminb}. Accordingly, the common-clock convention described here does not imply a measured synchronization-error bound.

\clearpage
\bibliographystyle{plainnat}
\bibliography{tacinf}

\end{document}